\documentclass[runningheads]{llncs}

\usepackage[T1]{fontenc}
\usepackage{graphicx}
\usepackage{amsmath,amssymb}
\usepackage[hidelinks]{hyperref}
\usepackage{color}

\usepackage{subcaption}
\usepackage{wrapfig}
\usepackage{array}
\usepackage{booktabs}
\usepackage{enumitem}
\usepackage{threeparttable}
\usepackage{marvosym}
\usepackage{placeins}

\newif\ifappendix
\appendixtrue      % set to \appendixtrue to include the appendix

\newif\ifanonymous
\anonymousfalse     % set to \anonymousfalse for camera-ready
\ifappendix\else
  \usepackage{xr}
  \makeatletter
  \let\XR@orig@XRO@input\XRO@input
  \def\XRO@input#1{%
    \begingroup
    \let\bibcite\@gobbletwo
    \XR@orig@XRO@input{#1}%
    \endgroup}
  \makeatother
\fi

\newcommand{\appref}[1]{%
  \ifappendix
    \S\ref{#1}%
  \else
    \href{https://doi.org/10.6084/m9.figshare.33032678}{\S\ref*{#1}}~\cite{heiss-figshare26a}%
  \fi
}

\newcommand{\dataset}{\mathcal{D}}           % observed dataset
\newcommand{\featurespace}{\mathcal{X}}      % feature space
\newcommand{\real}{\mathbb{R}}               % real numbers

\newcommand{\featureRV}[1]{X_{#1}}          % feature random variable, e.g. \featureRV{j}
\newcommand{\targetRV}{Y}                    % target random variable

\newcommand{\distribution}[1]{\mathbb{P}_{#1}}   % probability measure, e.g. \distribution{X}

\newcommand{\featureVec}[2][]{%
  \mathbf{x}%
  \ifx\relax#2\relax\else_{#2}\fi%
  \ifx\relax#1\relax\else^{(#1)}\fi%
}
\newcommand{\featureVal}[2][]{%
  x%
  \ifx\relax#2\relax\else_{#2}\fi%
  \ifx\relax#1\relax\else^{(#1)}\fi%
}

\newcommand{\ours}{\textsc{Sails}}           % short name (small caps)
\newcommand{\ourslong}{\textbf{S}urrogate-based \textbf{A}nalysis of \textbf{I}nteractions via \textbf{L}ocal effect \textbf{S}mooths}  % full name

\newcommand{\truefunc}{f}                    % true underlying function
\newcommand{\model}{\hat{f}}                 % trained ML model

\newcommand{\surr}[1]{\hat{s}^{(#1)}}          % fitted GAM surrogate for interval #1: \surr{k}
\newcommand{\smooth}[2]{\hat{s}_{#1}^{(#2)}}   % smooth term, feature #1, interval #2: \smooth{l}{k}

\newcommand{\ale}{\text{ALE}}                % accumulated local effects
\newcommand{\expectation}{\mathbb{E}}        % expectation operator

\newcommand{\targetVal}[1]{y^{(#1)}}             % target value, \targetVal{i} → y^{(i)}
\newcommand{\targetspace}{\mathbb{R}}             % target space (regression)
\newcommand{\noise}{\varepsilon}                  % noise term

\begin{document}

\title{\textsc{Sails}: Surrogate-based Analysis of Interactions via Local Effect Smooths}

\titlerunning{\textsc{Sails}: Surrogate-based Analysis of Interactions via Local Effect Smooths}

\ifanonymous
  \author{Anonymous Author(s)}
  \authorrunning{Anonymous}
  \institute{Affiliation(s)\\
  Address(es)\\
  \email{e-mail}}
\else
  \author{Timo Heiß\inst{1,2}\orcidID{0009-0002-0392-4308}  % chktex 8
  \and
  Julia Herbinger\inst{3,4}\orcidID{0000-0003-0430-8523}  % chktex 8
  \and
  Bernd Bischl\inst{1,2}\orcidID{0000-0001-6002-6980}  % chktex 8
  \and
  Giuseppe Casalicchio\inst{1,2,4}\orcidID{0000-0001-5324-5966}  % chktex 8
  }
  \authorrunning{T.\ Heiß et al.}
  \institute{Department of Statistics, LMU Munich, Munich, Germany\\
  \and
  Munich Center for Machine Learning (MCML), Munich, Germany\\
  \and Leibniz Institute for Prevention Research and Epidemiology, Bremen, Germany\\
  \and Center for Advanced Studies (CAS), LMU Munich, Munich, Germany
  \email{{\Large\Letter}\{timo.heiss, giuseppe.casalicchio\}@stat.uni-muenchen.de}
  }
\fi

\maketitle

\begin{abstract}
Feature interactions drive much of the predictive power of machine learning models, yet existing explanation methods only detect and quantify interactions without revealing their functional form, or visualize only restricted interaction types.
We propose \ourslong\ (\ours), a model-agnostic framework that analyzes pairwise interactions through generalized additive model (GAM) surrogates fitted to the local effects of a black-box model.
For each interval of a feature of interest, the surrogate smooth terms isolate the interaction components on derivative level, enabling
(i)~interaction detection through a heuristic derived from significance tests on smooth terms,
(ii)~interaction form categorization into linear, product-separable, and non-product-separable types,
and (iii)~tailored, interpretable visualizations for each interaction type.
We empirically validate the framework through controlled simulations and a real-world task, showing its effectiveness for pairwise interactions, with limitations under strong feature correlations and higher-order interactions.
\ours{} fills a notable gap in the eXplainable AI (XAI) toolbox, going beyond detecting interactions alone to characterizing their functional form.

% \keywords{
%   explainable AI
%   \and feature interactions
%   \and local effects
% }
\end{abstract}

\section{Introduction}\label{sec:intro}

Machine learning (ML) models achieve high predictive performance across many domains~\cite{pugliese-dsm21a}, but their complexity makes them difficult to interpret.
This becomes particularly problematic in critical domains such as healthcare, legal, or finance, where decisions must be transparent and accountable.
The complex non-linearity in ML models arises from two sources~\cite{zhang-ipm23a}: non-linear feature transformations and feature interactions, i.e., non-additive effects where the influence of one feature on the prediction depends on another~\cite{sorokina-icml08a}.
Current eXplainable AI (XAI) methods primarily focus on the former, often overlooking feature interactions:
feature effect methods such as 1D partial dependence (PD)~\cite{friedman-annals01a} and accumulated local effects (ALE) plots~\cite{apley-jrsss20a} explain non-linear transformations of individual features, but obscure the presence of interactions~\cite{goldstein-jcgs15a}.

\begin{figure}[tb]
    \centering
    \begin{minipage}[t]{\textwidth}
        \begin{subfigure}[t]{0.3\textwidth}
            \centering
            \includegraphics[width=\textwidth]{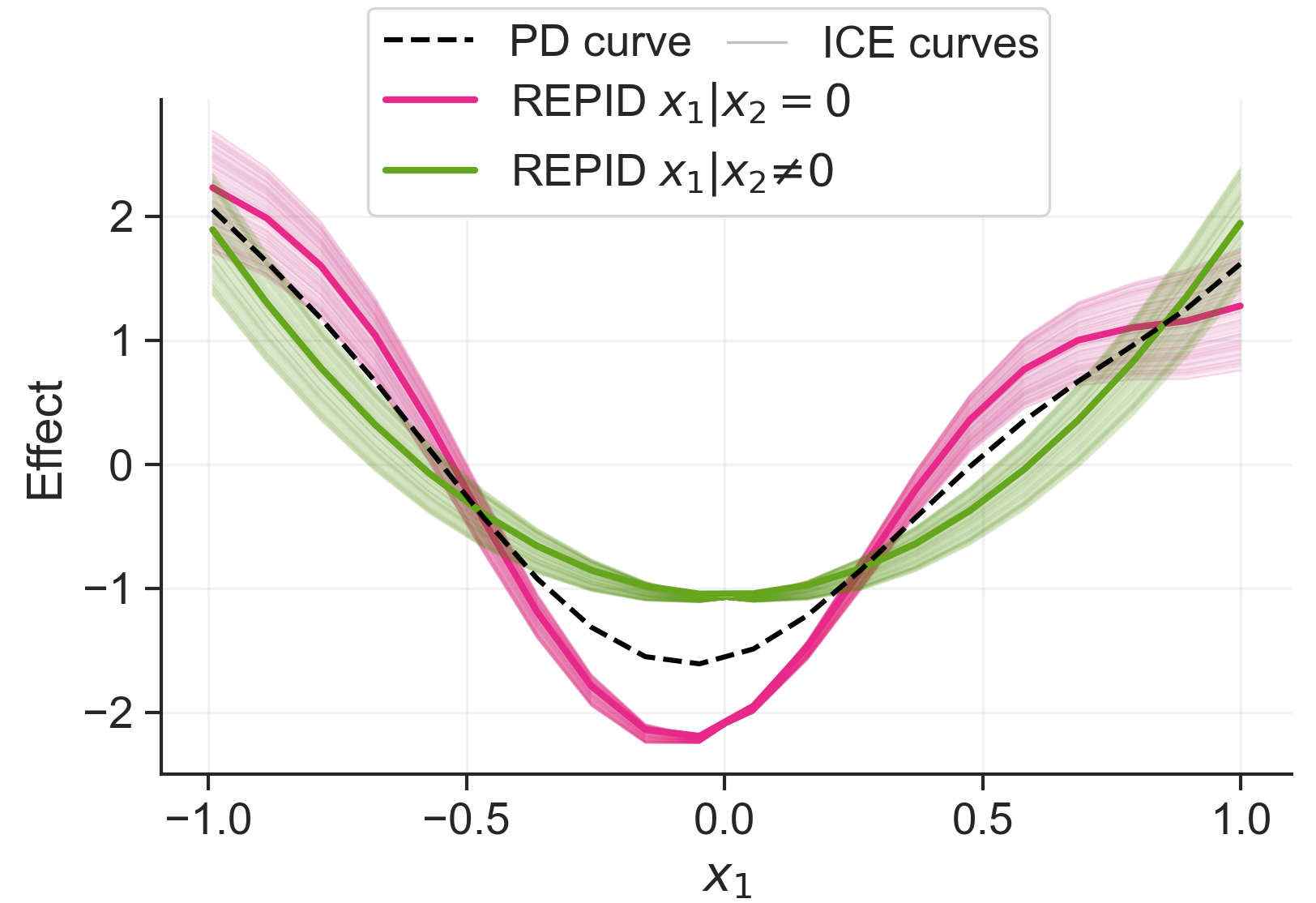}
            \caption{REPID-PD of $\hat{f}_{a}$}\label{fig:model-a-repid}
        \end{subfigure}
        \hfill
        \begin{subfigure}[t]{0.3\textwidth}
            \centering
            \includegraphics[width=\textwidth]{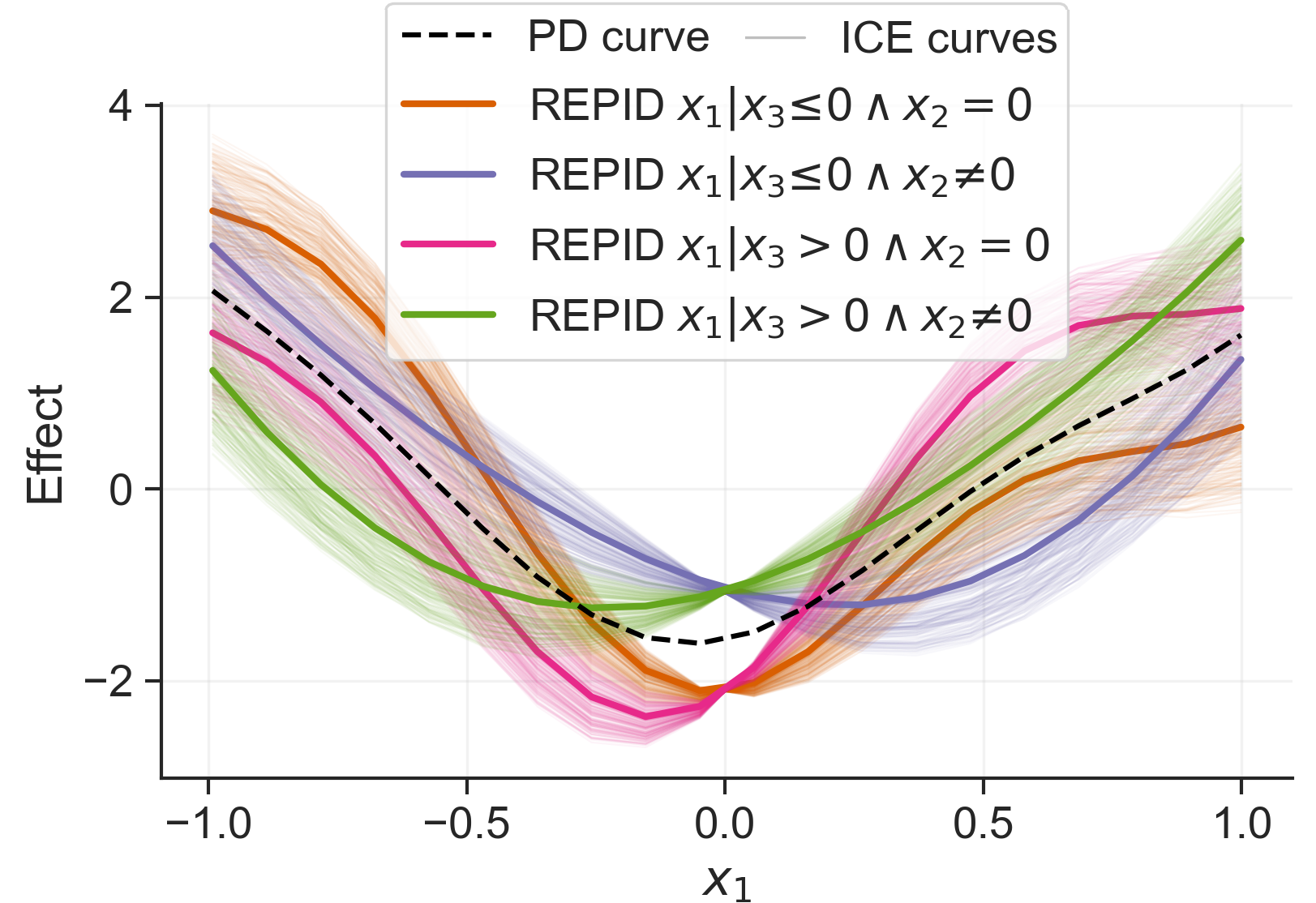}
            \caption{REPID-PD of $\hat{f}_{b}$}\label{fig:model-b-repid}
        \end{subfigure}
        \hfill
        \begin{subfigure}[t]{0.275\textwidth}
            \centering
            \includegraphics[width=\textwidth]{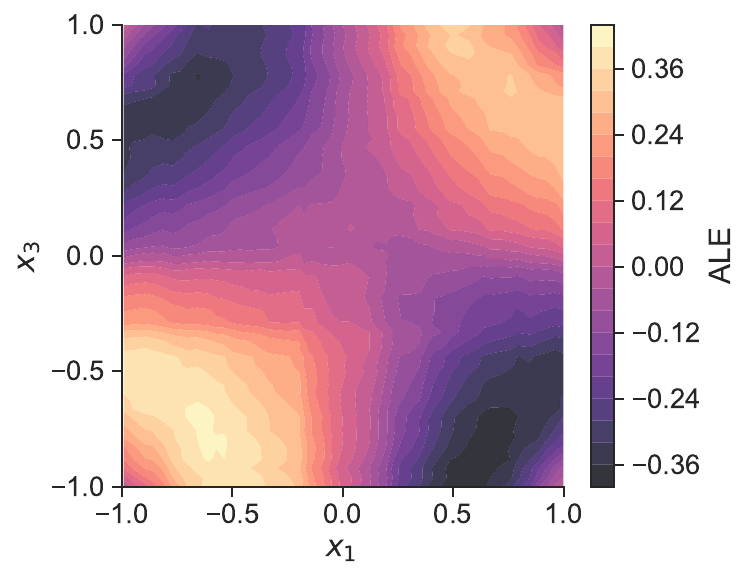}
            \caption{$X_1X_3$ 2D-ALE of $\hat{f}_{b}$}\label{fig:example-2d-ale}
        \end{subfigure}
    \end{minipage}
    \caption{Feature effect plots for the motivating example.
      Dashed lines are PD curves; thin lines are ICE~\cite{goldstein-jcgs15a} curves; bold lines are REPID~\cite{herbinger-pmlr22a} regional averages.}\label{fig:motivational-example}
\end{figure}

\textit{Motivating Example.}
Consider a synthetic dataset with features $X_1, X_3 \sim U(-1, 1)$, $X_2 \sim B(1, 0.5)$, where $X_1$ represents the \texttt{hour of the day}, $X_2$ a \texttt{weekend indicator}, and $X_3$ \texttt{temperature}.
Suppose a black-box model has learned to predict \texttt{power consumption} based on these features as $\hat{f}_{a}(\mathbf{x}) = 3x_1^2 + x_2 - \cos(1.5\pi x_1 + 0.5)(1-x_2) + r(\mathbf{x})$, where $r$ denotes other small effects.
The PD curve (dashed line) in Fig.~\ref{fig:model-a-repid} shows the marginal effect of $X_2$, hiding interactions.
Regional effects (REPID~\cite{herbinger-pmlr22a}) reveal how the effect of $X_1$ differs between weekdays and weekends.
Now consider the model
$\hat{f}_{b}(\mathbf{x}) = \hat{f}_{a}(\mathbf{x}) - \exp(x_3) + \sin(\pi x_1 x_3)$,
which adds a continuous interaction between $X_1$ and $X_3$.
Due to the discrete nature of its regions, REPID in Fig.~\ref{fig:model-b-repid} now fails to adequately capture the heterogeneity in the ICE curves caused by the continuous interaction with $X_3$.
Second-order ALE plots (2D-ALE; Fig.~\ref{fig:example-2d-ale}) visualize the interaction effect of $X_1$ and $X_3$, but the 3D display is difficult to interpret.

\textit{Related Work.}
Most XAI methods for feature interactions focus on detection or quantification~\cite{greenwell-arxiv18a,sorokina-icml08a}, most prominently the H-statistic~\cite{friedman-annals08a} and Shapley interaction indices~\cite{fumagalli-neurips23a,grabisch-ijgt99a,lundberg-nmi20a}.
Only a few methods aim to visualize interactions.
\texttt{shapiq} network plots~\cite{muschalik-neurips24a} show which features interact and how strongly, but do not reveal interaction forms.
Regional effects such as REPID~\cite{herbinger-pmlr22a} and GADGET~\cite{herbinger-jmlr24a} partition the feature space into regions where the feature effect is less affected by interactions, using a tree-based algorithm to reduce heterogeneity in local feature effects (see Fig.~\ref{fig:model-a-repid}).
% The splits reveal interacting features, and interaction strength is quantified by the reduction in heterogeneity across regions.
However, this does not visualize the interaction form itself and faces limitations with continuous interactions.
FINCH~\cite{kleinau-arxiv25a} visualizes how features interact for a single observation, but is only local.
2D feature effects like 2D-ALE plots~\cite{apley-jrsss20a} capture second-order interaction effects for any interaction type. However, they produce difficult-to-interpret (3D) contour plots and require prior knowledge of which feature pairs interact before visualizing them.
The closest existing integrated approach for capturing interaction forms globally is presented in~\cite{zhang-ipm23a}, which detects interactions via HDMR~\cite{zhang-kbs22a}, determines pairwise product-separability, infers separable interaction sets, and visualizes their functional form. This is restricted to product-separable interactions, assumes independent features, and is sensitive to the chosen interaction-strength threshold.

\textit{Contributions.}
The XAI literature thus lacks an integrated approach for detecting and visualizing all types of interaction forms.
As a remedy, we introduce \ours{} (\ourslong), a model-agnostic framework analyzing pairwise interactions of a feature of interest (FOI) via GAM surrogates fitted to the local effects of a black-box model, offering three integrated capabilities:
(i)~interaction detection through a heuristic derived from significance tests on smooth terms,
(ii)~interaction form categorization into linear, product-separable, and non-product-separable,
and (iii)~interaction visualizations for each type.
We empirically validate \ours{} in controlled simulations and apply it to a real-world task, showing its effectiveness for pairwise interactions, with limitations for strong correlations and higher-order interactions.

\textit{Delimitation.}
Like GADGET~\cite{herbinger-jmlr24a}, \ours{} exploits the heterogeneity of local effects. While GADGET uses it to partition the feature space into homogeneous regions, we aim to characterize the interaction form.
Unlike existing surrogate model approaches (global~\cite{herbinger-ecai24a} or local~\cite{ribeiro-acm16a}) that fit interpretable models to approximate black-box predictions,
\ours{} does not aim at model-level approximation but fits local effects to directly characterize feature interactions.

\section{Notation and Background}\label{sec:background}

% \paragraph{Notation.}
Let $\featurespace \subseteq \real^p$ be the feature and $\targetspace$ the target space for a regression task.
Random feature vectors are denoted $\featureRV{} = (\featureRV{1}, \ldots, \featureRV{p})$ with realizations $\featureVec{} \in \featurespace$.
A dataset $\dataset = {\{(\featureVec[i]{}, \targetVal{i})\}}_{i=1}^n$ consists of $n$ i.i.d.\ draws from an unknown distribution $\mathbb{P}_{XY}$.
We denote feature values of the $i$-th observation by $\featureVec[i]{}=(\featureVal[i]{1}, \dots, \featureVal[i]{p})^\top$ and the corresponding target value by $\targetVal{i}$.
We refer to the $j$-th feature by $\featureVec{j} = (\featureVal[1]{j}, \dots, \featureVal[n]{j})^\top$ or $\featureRV{j}$; complement sets are prefixed by $-$ (e.g., $-j$).
We assume $\targetRV{}=\truefunc(\featureRV{})+\noise$, i.e., a true function $\truefunc: \featurespace \to \targetspace$ and random noise term $\noise$.
An ML model $\model: \featurespace \to \targetspace$ is learned on $\dataset$ via empirical risk minimization (ERM).
We use $\partial_j$ to denote partial derivatives w.r.t.\ $x_j$ (other subscripts analogously).

\textit{Functional Decomposition.}
The functional ANOVA (FANOVA)~\cite{hooker-acm04a,hooker-jcgs07a} decomposes any square-integrable function $f$ as sum of lower-order components:
\[
    f(\featureVec{}) = g_0 + \sum_j g_j(x_j) + \sum_{j < l} g_{jl}(x_j, x_l) + \cdots + g_{1,2,\ldots,p}(x_1, x_2, \ldots, x_p),
\]
where $g_0$ is a constant, $g_j$ a main effect, $g_{jl}$ a pure two-way interaction effect, etc.
A unique decomposition is enforced by the vanishing conditions $\forall W \neq \emptyset, \ \forall j \in W$ $\int g_W \, d\distribution{\featureRV{j}} = 0$ under independence or $\int g_W(\mathbf{x}_W)\, w(\mathbf{x})\, dx_j\, d\mathbf{x}_{-W} = 0$ with weight function $w$ (e.g., joint density) under correlation~\cite{hooker-jcgs07a}.
Two features $\featureRV{j}$ and $\featureRV{l}$ interact iff $g_W \not\equiv 0$ for some $W \supseteq \{j, l\}$, equivalently if $\expectation[(\partial^2_{jl} f)^2] > 0$~\cite{friedman-annals08a}.  % chktex 3

\textit{Product-separability.}
A function $f$ is \emph{(pairwise) product-separable} in the features $X_j$ and $X_l$ if it can be written as $f(\featureVec{}) = \phi_{-j}(\featureVec{-j})\cdot\phi_{-l}(\featureVec{-l})$ for some functions $\phi_{-j}, \phi_{-l}: \real^{p-1} \to \real$.
A necessary and sufficient condition is that the ratio $f(\ldots, x_j, \ldots, x_l, \ldots) / f(\ldots, x_j', \ldots, x_l, \ldots)$ is independent of $x_l$ (and vice versa)~\cite{zhang-ipm23a}, providing a practical detection criterion for separability.

\textit{ALE Local Effects.}
ALE~\cite{apley-jrsss20a} is a feature effect method that respects feature correlations by integrating the conditional expectation of the local derivative along an FOI.  % chktex 13
With local effects $h(x_j, \featureVec{-j}) := \partial_j \model(x_j, \featureVec{-j})$, the (un-centered) 1D-ALE is given by
$
  \ale_j(x_j) = \int_{x_{\min,j}}^{x_j} \expectation_{\featureRV{-j} \mid \featureRV{j} = z}\bigl[h(z, \featureRV{-j})\bigr]\, dz.
$
In practice, $h$ is approximated by finite differences.
Partitioning the support of $\featureRV{j}$ into $K$ intervals $I_k = (z_{k-1,j}, z_{k,j}]$, the per-observation finite difference in interval $I_k$ is
$\hat{\tilde{h}}_k^{(i)} = \model(z_{k,j}, \featureVec[i]{-j}) - \model(z_{k-1,j}, \featureVec[i]{-j})$, and    % chktex 9
$
  \widehat{\ale}_j(x_j) = \sum_{k=1}^{k_j(x_j)} \frac{1}{n_k} \sum_{i:\, \featureVal[i]{j} \in I_k} \hat{\tilde{h}}_k^{(i)}
$
the (un-centered) ALE estimate, where $n_k$ is the number of observations in $I_k$ and $k_j(x_j)$ the interval index of $x_j$. The local effects are themselves studied, e.g., in RHALE~\cite{gkolemis-ecai23a} to visualize heterogeneity or in GADGET~\cite{herbinger-jmlr24a} to reduce it.
%This is closely related to the concept of marginal effects~\cite{scholbeck-dmkd24a}.

\textit{Local Decomposability.}
Similar to centered ICE curves and Shapley values, ALE local effects satisfy local decomposability, i.e., they solely depend on the main effect and interaction components that involve the FOI~\cite{herbinger-jmlr24a}:  % chktex 13
\begin{equation}\label{eq:local-decomp}
    h(x_j, \featureVec[i]{-j}) = \partial_j g_j(x_j)
      + \sum_{k=2}^{p} \sum_{\substack{W \subseteq \{1,\dots,p\},\\|W|=k,\, j \in W}}
        \partial_j g_W(x_j, \featureVec[i]{W \setminus \{j\}}).
\end{equation}
%
% This is a key property of local effects that our framework exploits.  % chktex 17

\section{\ours: Analyzing Interactions via Local Effects}\label{sec:method}

\begin{figure}[tb]
    \centering
    \includegraphics[width=\linewidth]{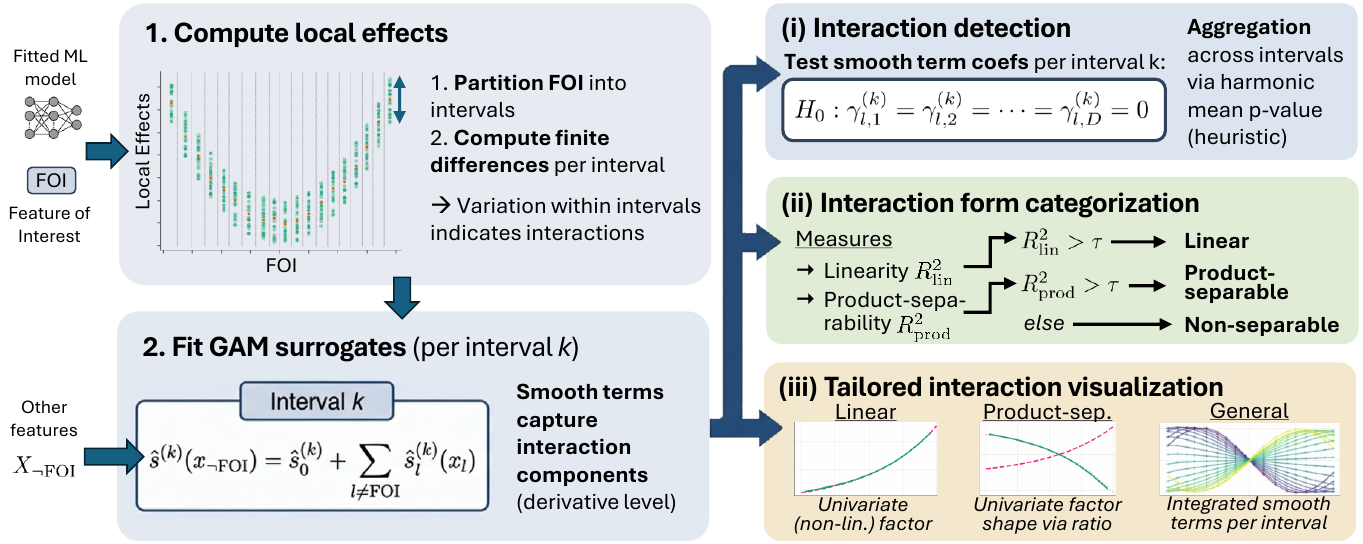}
    \caption{Overview of the \ours{} framework.}\label{fig:overview}
\end{figure}

\ours{} consists of two steps (1-2) that enable three integrated capabilities (i-iii) (see Fig.~\ref{fig:overview}).
It requires (1) estimating local effects in intervals of the FOI (\S\ref{sec:idea}),
and (2) fitting one GAM surrogate per interval on the local effects,
whose smooth terms isolate pairwise interactions with the FOI (\S\ref{sec:gam-surrogates}).
Based on them, \ours{} first (i) detects interactions via a heuristic derived from hypothesis tests (\S\ref{sec:detection}),
(ii) categorizes detected interactions as linear, product-separable, or non-separable (\S\ref{sec:categorization}),
and (iii) visualizes them accordingly (\S\ref{sec:visualization}).

\subsection{General Idea}\label{sec:idea}

The core concept of \ours{} is to exploit the heterogeneity of ALE local effects to capture and analyze interactions.
This is based on their local decomposability and the consequence that any variation of local effects at a fixed value of the FOI $X_j$ must arise from interactions involving $X_j$.
Consider the local effects of $\hat{f}(\mathbf{x}) = x_1^3 + x_2 + 0.5 x_1 x_2$ for illustration (cf.\ scatterplot in step 1 of Fig.~\ref{fig:overview}).
Since $\partial_{1} \hat{f} = 3x_1^2 + 0.5 x_2$, fixing $x_1$ leaves $0.5 x_2$ as the only source of variation, i.e., it is caused by the interaction term $0.5 x_1 x_2$.
The FANOVA decomposition of local effects in Eq.~\eqref{eq:local-decomp} shows that this generally holds: fixing $X_j$ leaves the interactions with it as the only variation source.
To characterize this heterogeneity, we fit interpretable surrogate models to the local effects at fixed FOI values.
In practice, this is realized through interval-based estimation of the local effects (as in~\cite{apley-jrsss20a}): we define a grid partitioning the FOI in $K$ intervals
$\{(z_{k-1,j}, z_{k,j}]\}_{k=1}^K$.  % chktex 9 chktex 3
% We recommend quantile-based intervals to ensure similar observation counts per interval and adequate representation of the distribution of $X_j$.
We estimate the local effects for all observations $\featureVec[i]{}$ via finite differences normalized by interval width (to account for varying interval size), for $k$ such that $\featureVal[i]{j} \in (z_{k-1,j}, z_{k,j}]$: % chktex 9
$
    \hat h^{(i)}_k = \left(\hat f(z_{k,j}, \mathbf x_{-j}^{(i)})-\hat f(z_{k-1,j},\mathbf x_{-j}^{(i)})\right)/\left(z_{k,j}-z_{k-1,j}\right).
$

\subsection{GAM Surrogates}\label{sec:gam-surrogates}

We use generalized additive models (GAMs) as surrogate models. They are interpretable and enable hypothesis tests.
\ours{} fits one GAM per interval $k$, with the remaining features as inputs and (normalized) local effects $\hat{h}_k^{(i)}$ as target:
\[
    \surr{k}(\featureVec[i]{-j}) = \smooth{0}{k} + \sum_{l \neq j} \smooth{l}{k}(x_l^{(i)}),
\]
with smooth terms $\smooth{l}{k}$ (e.g., penalized B-splines). The ERM problems are then:
\begin{equation*}
    {\arg\min}_{\surr{k}} \mathcal{R}_\text{emp}^{(k)}(\surr{k}) = {\arg\min}_{\smooth{0}{k},\smooth{l}{k}} \sum_{i: \ x_j^{(i)} \in I_k} {\bigg(\hat{h}^{(i)}_k - \smooth{0}{k} - \sum_{l\neq j} \smooth{l}{k}(x_l^{(i)})\bigg)}^2.
\end{equation*}
The GAM structure implies only up to two-way interactions in $\hat f$. This assumption is often reasonable, as functional decompositions up to order two provide, in many cases, adequate descriptions of high-dimensional functions~\cite{li-jmc12a}.
Nonetheless, we evaluate the behavior of \ours{} under higher-order interactions in \S\ref{sec:eval}.
In addition, we propose a goodness-of-fit check (e.g., in-sample generalized $R^2$) to ensure that the smooth terms capture local effect variation adequately or, conversely, flag potential higher-order interactions.

\textit{Theoretical Analysis.}
To analyze what the smooth terms capture, consider~local effects $h$ with $x_j$ fixed to
$\bar{z}_{k,j}$ in the population L2-risk over $\featureRV{-j} \mid \featureRV{j}=\bar{z}_{k,j}$:\footnote{%
  To make the connection between theoretical formulation and empirical estimation more explicit, one might think of $\bar{z}_{k,j}$ as the midpoint of the interval $I_k$.
}
\begin{align*}
    &\mathcal{R}^{(k)}(\surr{k})
    = \expectation_{\featureRV{-j} \mid \featureRV{j}=\bar{z}_{k,j}}\!\bigg[{\bigg(h(\bar{z}_{k,j}, \featureRV{-j}) - \smooth{0}{k} - \sum_{l\neq j} \smooth{l}{k}(\featureRV{l})\bigg)}^{2}\bigg] \\
    &= \expectation_{\featureRV{-j} \mid \featureRV{j}=\bar{z}_{k,j}}\!\bigg[{\bigg(\partial_j g_j(\bar{z}_{k,j}) + \sum_{l\neq j} \partial_j g_{jl}(\bar{z}_{k,j}, \featureRV{l}) + r - \smooth{0}{k} - \sum_{l\neq j} \smooth{l}{k}(\featureRV{l})\bigg)}^{2}\bigg],
\end{align*}
where $r$ is a remainder term of higher-order interactions.
Since GAM terms are centered by definition, we conjecture based on the results of~\cite{stone-annals85a,stone-annals86a} and under the assumption $r=0$ that the intercept and smooth terms capture:\footnote{Stone~\cite{stone-annals85a,stone-annals86a} shows that centered additive model components without penalization converge to the true underlying components under additivity and regularity conditions, given basis dimensions grow with $n$. We conjecture approximate relationships here, given our finite samples, penalization, and fixed basis dimensions.}
\begin{equation}\label{eq:smooth-terms}
    \smooth{0}{k} \approx \partial_j g_j(\bar{z}_{k,j}) + c_0^{(k)},
    \qquad
    \smooth{l}{k}(x_l) \approx \partial_j g_{jl}(\bar{z}_{k,j}, x_l) - c_l^{(k)},
\end{equation}
with $c_l^{(k)} = \expectation_{\featureRV{l} \mid \featureRV{j} = \bar{z}_{k,j}}[\partial_j g_{jl}(\bar{z}_{k,j}, \featureRV{l})]$
and $c_0^{(k)} = \sum_{l \neq j} c_l^{(k)}$.
Intuitively, $\smooth{l}{k}$ captures how the FOI's interaction with $X_l$ in interval $k$ contributes to the local effects, with $c_l^{(k)}$ as centering to match GAM constraints.
The intercept $\smooth{0}{k}$ absorbs the main effect of $X_j$ plus centering constants.
We empirically validate the conjecture via capabilities (i)-(iii) in diverse scenarios in \S\ref{sec:eval}.

\subsection{Interaction Detection}\label{sec:detection}

To identify which features $X_l$ interact with the FOI $X_j$, \ours{} tests each smooth term $\smooth{l}{k}$ against the null hypothesis that its basis coefficients are zero, i.e., $H_0: \gamma_{l,1}^{(k)} = \gamma_{l,2}^{(k)} = \cdots = \gamma_{l,D}^{(k)} = 0$ with $D$ basis coefficients $\gamma_{l,d}^{(k)}$.
Under i.i.d.\ Gaussian errors, this can be done via an F-test~\cite{fahrmeir-book13a}, yielding an interval-specific p-value $p_l^{(k)}$.
Since the $\hat{h}_k^{(i)}$ are computed directly from $\hat{f}$ with no additive noise term, the i.i.d.\ error assumption does not formally hold.
Together with spline penalization, this motivates treating the p-values as a heuristic.
To obtain a global value, and since tests within different intervals for the same feature are likely dependent
(a strong interaction in an interval suggests significance in adjacent ones), \ours{} aggregates p-values across intervals via the harmonic mean p-value,
which is robust under dependence and provides approximate control of the family-wise error rate~\cite{wilson-nasusa19a}.
To account for multiple comparisons across features, a multiple-testing correction (here: Benjamini-Hochberg false discovery rate (FDR) correction~\cite{benjamini-jrss95a}) is applied.
Since we use the harmonic-mean directly and apply a standard F-test despite penalization and non-i.i.d.\ errors, the result should be seen as a heuristic for global interaction significance rather than an exact p-value.
Features with a value below a threshold $\alpha$ are flagged as interacting.
By analogy with conventional significance levels, we default to $\alpha = 0.05$.

\subsection{Interaction Categorization}\label{sec:categorization}

Knowing whether an interaction is linear, product-separable, or neither guides visualization and aids interpretation.
We introduce two $R^2$-based heuristic measures that operate directly on the smooth terms of the surrogate GAMs.
\begin{figure}[tb]
    \centering
    \begin{subfigure}[t]{\textwidth}
        \centering
        \includegraphics[width=\textwidth]{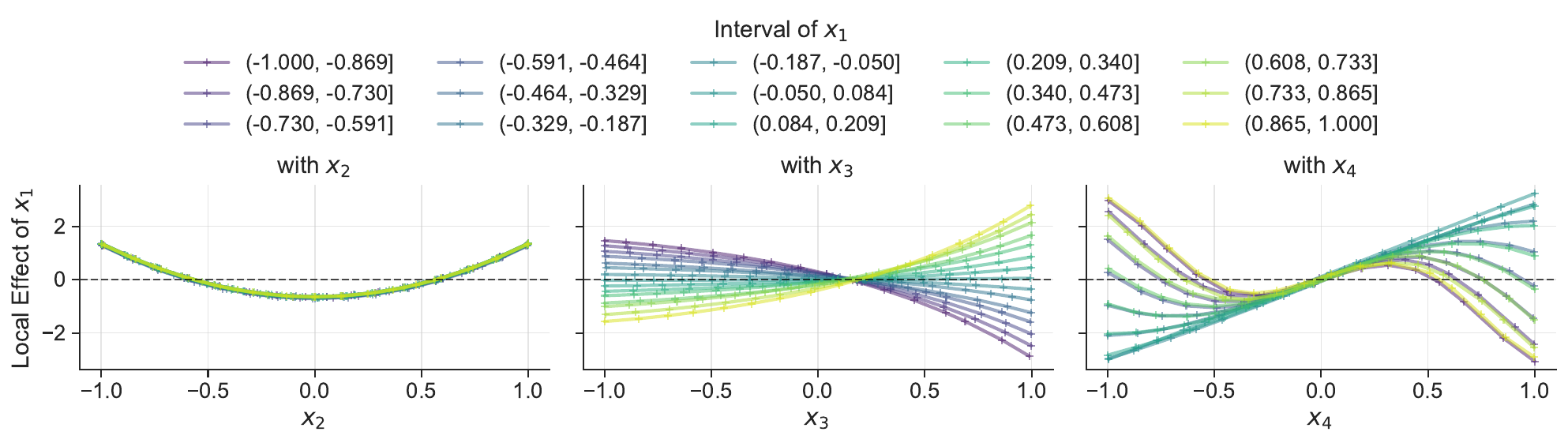}
    \end{subfigure}
    \hfill
    \begin{subfigure}[t]{\textwidth}
        \centering
        \includegraphics[width=\textwidth]{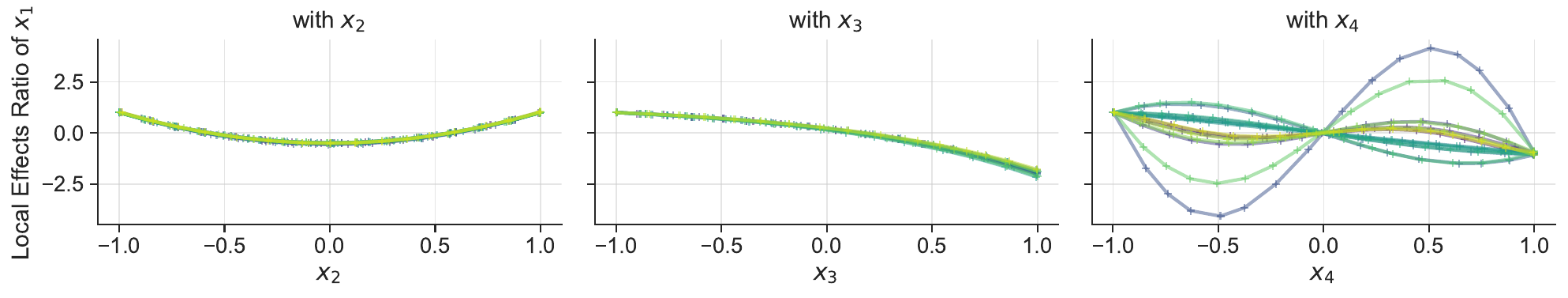}
    \end{subfigure}
    \caption{Smooth terms $\smooth{l}{k}$ (1st row) \& ratios $r_l^{(k)}$ (2nd) for $\hat{f}(\mathbf{x}) = 3x_1 + 2x_1 x_2^2 + x_1^2 \exp(x_3) + \sin(\pi x_1 x_4)$ (FOI: $X_1$);
      features are indep. $U(-1,1)$.
      Linear interaction ($X_2$): all $\smooth{2}{k}$ are identical and so are the ratios.
      Product-separable interaction ($X_3$): all $\smooth{3}{k}$ have the same shape but scaled differently, ratios are identical.
      Non-product-separable interaction ($X_4$): shapes of both $\smooth{4}{k}$ and $r_4^{(k)}$ vary.
      }\label{fig:intuition}
\end{figure}
\begin{definition}[Linear two-way interaction]\label{def:linear}
  A two-way interaction $g_{jl}$ from the FANOVA decomposition is linear w.r.t.\ a FOI $X_j$ if it can be expressed as
  $g_{jl}(x_j, x_l) = (x_j - \expectation[\featureRV{j}]) \cdot \phi_l(x_l)$
  for a univariate function $\phi_l$ with $\expectation[\phi_l(\featureRV{l})] = 0$.
\end{definition}
\begin{proposition}[Linearity via derivatives]\label{prop:linearity}
A two-way interaction $g_{jl}$ is linear w.r.t.\ $X_j$ if and only if $\partial_j g_{jl}(x_j, x_l)$ is independent of $x_j$.
  (Proof in \appref{app:proofs})
\end{proposition}

\textit{Linearity Measure.}
Following from Proposition~\ref{prop:linearity}, for a linear interaction, all smooth terms $\smooth{l}{k}$ are identical up to a centering constant across intervals (see Fig.~\ref{fig:intuition}).
\ours{} exploits this by evaluating each term at interval-specific grid points $\{\tilde{x}_l^{(k,g)}\}_{g=1}^{G_l}$ (e.g., quantiles of $X_l$ within $I_k$),
and fitting a single penalized spline to the pooled data $\bigcup_{k=1}^K \{(\tilde{x}_l^{(k,g)},\, \smooth{l}{k}(\tilde{x}_l^{(k,g)}))\}_{g=1}^{G_l}$.
Its coefficient of determination $R^2_{\text{lin}}$ measures how well a single curve (independent of $x_j$/$k$) explains all evaluations.
High values indicate that the interaction is linear in $X_j$.
However, under correlations, the centering $c_l^{(k)}$ varies across intervals, leading to shifts in the smooth terms, potentially causing the proposed mechanism to fail.
\begin{definition}[Product-separable two-way interaction]\label{def:prod-sep}
  A two-way interaction $g_{jl}$ from the FANOVA decomposition is product-separable if it can be expressed as
  $g_{jl}(x_j, x_l) = \phi_j(x_j) \cdot \phi_l(x_l)$
  for univariate functions $\phi_j$ and $\phi_l$;
  linearity is a special case with $\phi_j(x_j) = x_j - \expectation[\featureRV{j}]$.
\end{definition}
\begin{proposition}[Product-separability via ratios]\label{prop:ratio}
  A two-way interaction $g_{jl}$ is product-separable if and only if the ratios $\partial_j g_{jl}(x_j, x_l)/\partial_j g_{jl}(x_j, \tilde{x}_l^{\text{ref}})$ with a fixed reference value $\tilde{x}_l^{\text{ref}}$ are independent of $x_j$. (Proof in \appref{app:proofs})
\end{proposition}

\textit{Product-Separability Measure.}
For a product-separable interaction, smooth terms $\smooth{l}{k}$ share the same shape, differing only by a scaling factor.
Hence, following Proposition~\ref{prop:ratio}, their ratios at a fixed $\tilde{x}_l^{\text{ref}}$ are independent of $X_j$ (see Fig.~\ref{fig:intuition}).
\ours{} exploits this by reusing the evaluations $\smooth{l}{k}$ at interval-specific grid points (see linearity measure), computing ratios
$
  r_l^{(k)}(\tilde{x}_l^{(k,g)}) = \smooth{l}{k}(\tilde{x}_l^{(k,g)})/\smooth{l}{k}(\tilde{x}_l^{\text{ref}})
$,
and fitting a single penalized spline to $\bigcup_{k=1}^K \{(\tilde{x}_l^{(k,g)},\, r_l^{(k)}(\tilde{x}_l^{(k,g)}))\}_{g=1}^{G_l}$.
Its coefficient of determination $R^2_{\text{prod}}$ shows how well a single curve (independent of $x_j$ or $k$) explains the ratios. High values indicate product-separability.
Under correlations, however, the centering constants $c_l^{(k)}$ become interval-dependent,
introducing additive shifts in the smooth terms that disrupt ratio invariance even for truly product-separable interactions (see \appref{app:proofs} for details).

\subsection{Interaction Visualization}\label{sec:visualization}

We propose three visualization strategies, selected based on the categorization measures and a threshold $\tau$ (e.g., $\tau = 0.9$ for 90\% explained variance).

\textit{Linear Interactions.}
If $R^2_{\text{lin}} \geq \tau$, \ours{} can visualize the interaction via the univariate smoother fitted in \S\ref{sec:categorization},
estimating $\phi_l(x_l)$ of the linear interaction.

\textit{Product-Separable Interactions.}
If $R^2_{\text{prod}} \geq \tau$ (but $R^2_{\text{lin}} < \tau$), \ours{} can visualize the interaction via the univariate smoother fitted to the ratios in \S\ref{sec:categorization},
estimating the functional shape of $\phi_l(x_l)$ up to a scaling factor.

\textit{General Visualization (Any Interaction Type).}
For non-product-separable interactions, or as general alternative,
\ours{} recovers the interaction $g_{jl}$ through integration.
Since $\smooth{l}{k}(x_l) \approx \partial_j g_{jl}(\bar{z}_{k,j}, x_l) - c_l^{(k)}$ (see Eq.~\eqref{eq:smooth-terms}), integration w.r.t.\ $x_j$ yields an estimate of $g_{jl}$ plus univariate functions of $x_l$ and $x_j$, which double centering removes.
In practice, we evaluate each $\smooth{l}{k}$ on a grid $\{\tilde{x}_l^{(g)}\}_{g=1}^{G_l}$ (e.g., quantiles of $X_l$),
ignoring grid points outside the support of a smooth term to avoid extrapolation.
The integral is approximated via the trapezoidal rule:
\[
  \hat{\tilde{g}}_{jl}^{(k')}(\tilde{x}_l^{(g)}) = \sum_{k=2}^{k'} \frac{\bar{z}_{k,j} - \bar{z}_{k-1,j}}{2}
  \bigl(\smooth{l}{k-1}(\tilde{x}_l^{(g)}) + \smooth{l}{k}(\tilde{x}_l^{(g)})\bigr),
\]
with starting value $\hat{\tilde{g}}_{jl}^{(1)} = 0$.
The result is doubly centered w.r.t.\ the marginals of $X_j$ and $X_l$ and their joint distribution
(analogously to~\cite{apley-jrsss20a}) to enforce the FANOVA centering.
This yields one curve $\hat{g}_{jl}^{(k')}$ per interval, each showing how the interaction effect varies with $X_l$.
This approach can represent any two-way interaction type, requiring no prior knowledge of the interaction form.  % chktex 17

\begin{table}[tb]
    \centering
    \scriptsize
    \caption{Simulation settings. $X_7$ to $X_9$ are noise features throughout.}\label{tab:settings}
    \begin{tabular}{lp{6.5cm}>{\raggedright\arraybackslash}p{3cm}}
        \toprule
        \textbf{Setting} & \textbf{Function $f$} & \textbf{Correlation} \\
        \midrule
        \textbf{I} & $f_{\textbf{I}}(\featureVec[]{})=3x_1 + x_2 + \cdots + x_6 + x_1 x_2 + x_1\exp(x_3) + x_1^2 x_4 + x_1^2\log(|x_5|+1) + \sin(\pi x_1 x_6)$ & $\rho_{jl}=0 \ \forall j\neq l$ \\
        \midrule
        \textbf{II} & $f_{\textbf{II}}(\featureVec[]{})=f_{\textbf{I}}(\featureVec[]{})$ & $\rho_{12}\approx 0.3$,~$\rho_{13}\approx 0.55$, $\rho_{1\{4,7\}}\approx 0.85$, $\rho_{1\{5,6,8,9\}}\approx 0$ \\
        \midrule
        \textbf{III} & $f_{\textbf{III}}(\featureVec[]{})=f_{\textbf{I}}(\featureVec[]{}) + x_1 x_2 x_3$ & $\rho_{jl}=0 \ \forall j\neq l$ \\
        \midrule
        \textbf{IV} & $f_{\textbf{IV}}(\featureVec[]{}) = f_{\textbf{III}}(\featureVec[]{}) + \cos(\pi x_1 x_2 x_4) + x_1^2 x_5 x_6^2$ & $\rho_{jl}=0 \ \forall j\neq l$ \\
        \bottomrule
    \end{tabular}
\end{table}

\section{Empirical Evaluation}\label{sec:eval}

We empirically validate our framework in controlled simulations.
We consider four settings with nine features $X_1, \ldots, X_9 \sim U(-1, 1)$ (marginals),
where $X_7$ to $X_9$ are noise features. The ground truth functions $f$ and correlation structures $\rho$ are specified in Tab.~\ref{tab:settings}. The target is generated as $Y = f(X) + \varepsilon$ with $\varepsilon \sim \mathcal{N}(0, \mathrm{Var}[f]/5)$.
We draw 1000 samples and apply \ours{} to the local effects of $f$ (``oracle predictor'') evaluated on these samples, as this allows for a comparison against known interaction structures.\footnote{%
  Additionally, we use four ML models (GAM with interactions, interaction-restricted XGBoost, full XGBoost, and SVM-RBF), for which we report results in \appref{app:supp-results}.
}
We perform 30 repetitions per setting with $X_1$ as FOI.
Further details and parametrizations are provided in \appref{app:experiment-details}.
We report the goodness-of-fit of the surrogate models of \ours{} in \appref{app:supp-results}, which indicate sufficient capacity to fit the local effect variation in-sample.\footnote{We publish a reproducible Python implementation of our experiments with results at \url{https://github.com/slds-lmu/paper_2026_interaction_analysis_workshop_code}.}

\subsection{Interaction Detection Results}\label{sec:eval-sig}

\begin{figure}[tb]
    \centering
    \begin{subfigure}[t]{0.31\textwidth}
        \centering
        \includegraphics[width=\textwidth]{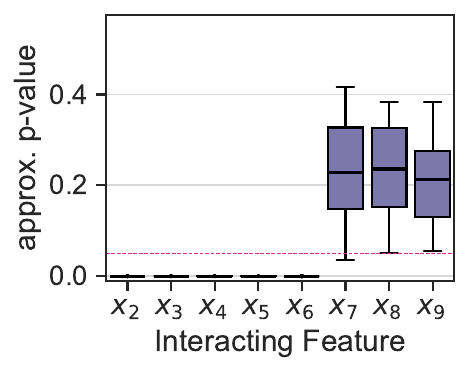}
        \caption{Setting \textbf{I}.}
    \end{subfigure}
    \hfill
    \begin{subfigure}[t]{0.31\textwidth}
        \centering
        \includegraphics[width=\textwidth]{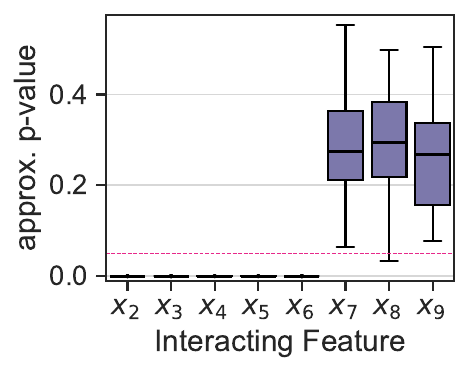}
        \caption{Setting \textbf{II}.}
    \end{subfigure}
    \hfill
    \begin{subfigure}[t]{0.31\textwidth}
        \centering
        \includegraphics[width=\textwidth]{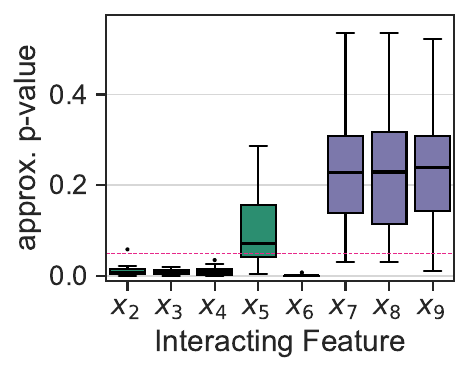}
        \caption{Setting \textbf{IV}.}
    \end{subfigure}
    \caption{Interaction detection for the oracle across 30 repetitions.
      Green boxes: truly interacting features ($X_2$ to $X_6$);
      purple boxes: non-interacting features ($X_7$ to $X_9$);
      dashed line: $\alpha = 0.05$.
      Full results in \appref{app:further-results-detection}.}\label{fig:significance}
\end{figure}

We compute the \ours{} interaction detection heuristic from \S\ref{sec:detection} and compare the results to a threshold of $\alpha = 0.05$ in Fig.~\ref{fig:significance}.
Under two-way interactions and independent features (Setting \textbf{I}), the method correctly assigns low values to all interacting features
and high values to noise features, with very few false positives.
Under feature correlations (\textbf{II}), interaction detection remains robust, consistent with ALE's conditional perspective.
Notably, $X_7$ (noise feature strongly correlated with the FOI) is correctly identified as non-interacting.
Higher-order interactions (\textbf{IV}) cause a slight increase in values for weaker interactions ($X_5$) and inflate false positive rates marginally,
consistent with surrogate misspecification.
Overall, the error rates remain acceptably low across all settings.

\subsection{Interaction Categorization Results}\label{sec:eval-cat}

\begin{figure}[tb]
    \centering
    \begin{subfigure}[t]{0.24\textwidth}
        \centering
        \includegraphics[width=\textwidth]{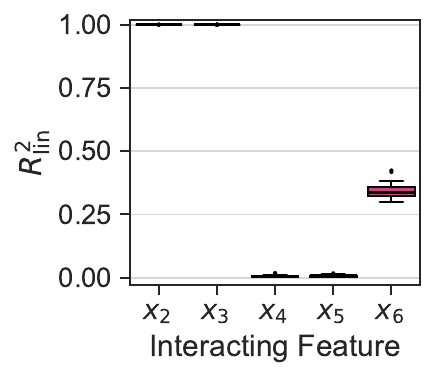}
        \caption{Setting \textbf{I}: $R^2_{\text{lin}}$.}
    \end{subfigure}
    \hfill
    \begin{subfigure}[t]{0.24\textwidth}
        \centering
        \includegraphics[width=\textwidth]{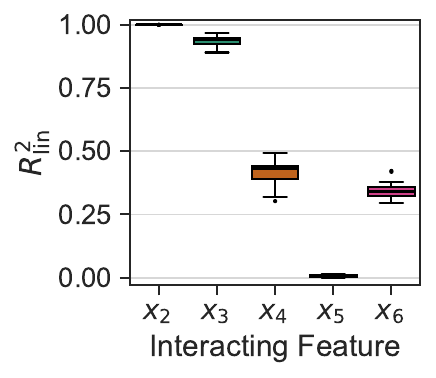}
        \caption{Setting \textbf{II}: $R^2_{\text{lin}}$.}
    \end{subfigure}
    \begin{subfigure}[t]{0.24\textwidth}
        \centering
        \includegraphics[width=\textwidth]{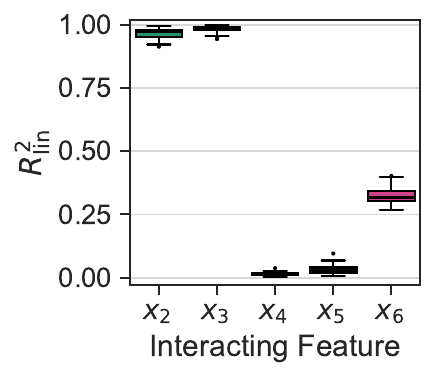}
        \caption{Setting \textbf{III}: $R^2_{\text{lin}}$.}
    \end{subfigure}
    \hfill
    \begin{subfigure}[t]{0.24\textwidth}
        \centering
        \includegraphics[width=\textwidth]{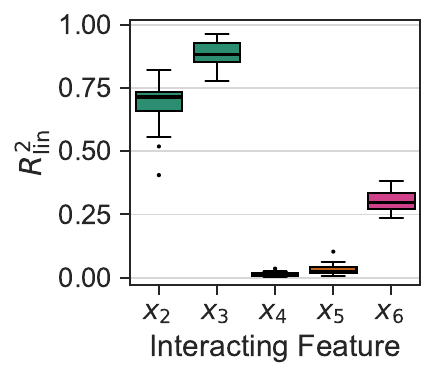}
        \caption{Setting \textbf{IV}: $R^2_{\text{lin}}$.}
    \end{subfigure}

    \vskip 0.4\baselineskip
    \begin{subfigure}[t]{0.24\textwidth}
        \centering
        \includegraphics[width=0.9\textwidth]{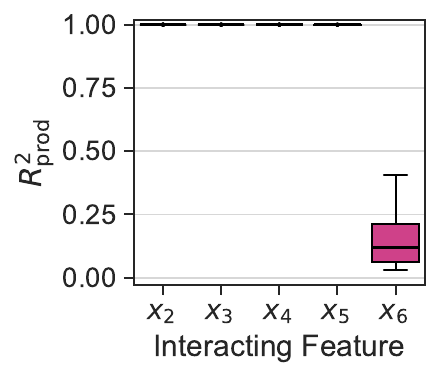}
        \caption{Setting \textbf{I}: $R^2_{\text{prod}}$.}
    \end{subfigure}
    \hfill
    \begin{subfigure}[t]{0.24\textwidth}
        \centering
        \includegraphics[width=0.9\textwidth]{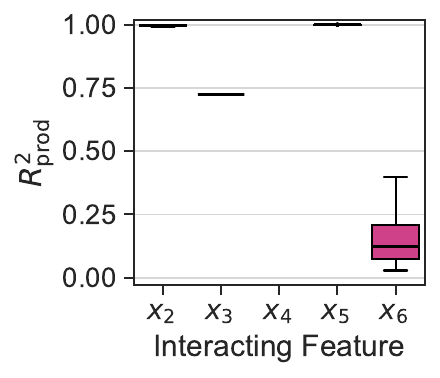}
        \caption{Setting \textbf{II}: $R^2_{\text{prod}}$.}
    \end{subfigure}
    \begin{subfigure}[t]{0.24\textwidth}
        \centering
        \includegraphics[width=0.9\textwidth]{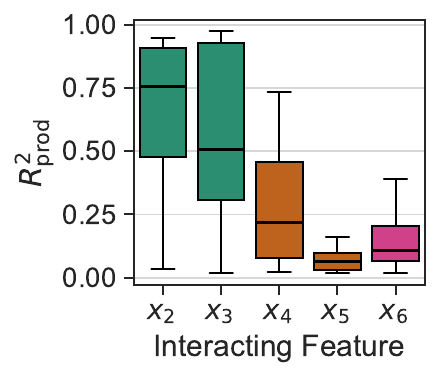}
        \caption{Set.~\textbf{III}:~$R^2_{\text{prod}}$.}
    \end{subfigure}
    \hfill
    \begin{subfigure}[t]{0.24\textwidth}
        \centering
        \includegraphics[width=0.9\textwidth]{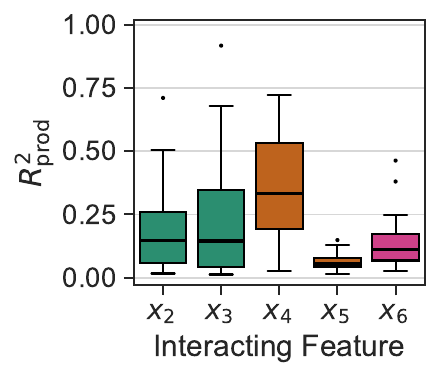}
        \caption{Set.~\textbf{IV}:~$R^2_{\text{prod}}$.}
    \end{subfigure}
    \caption{Categorization measures for the oracle across 30 repetitions.
      Colors indicate true interaction type:
      green = linear, orange = non-linear product-separable, pink = non-product-separable. Full results in \appref{app:further-results-categorization}.}\label{fig:categorization}
\end{figure}

We compute the \ours{} linearity and product-separability measures from \S\ref{sec:categorization} and report the results in Fig.~\ref{fig:categorization}.
Under independence and two-way interactions (Setting \textbf{I}), both measures achieve perfect discrimination. Linear interactions ($X_2, X_3$) receive high $R^2_{\text{lin}}$  (and non-linear ones low $R^2_{\text{lin}}$),
product-separable interactions ($X_2$ to $X_5$) receive high $R^2_{\text{prod}}$,
and non-product-separable interactions receive low $R^2_{\text{prod}}$.
Under correlations (\textbf{II}), $R^2_{\text{lin}}$ degrades only modestly, discrimination between linear and non-linear is still perfect.
$R^2_{\text{prod}}$ is more sensitive to correlations, and for high correlations ($X_4$), no reference point exists that lies within the support of all smooth terms, preventing computation of $R^2_{\text{prod}}$.
Under linear higher-order interactions (\textbf{III}), $R^2_{\text{lin}}$ does not degrade visibly: the higher-order interaction is linear in $X_1$, so the smooth terms still behave linearly in the FOI.
Under non-linear higher-order interactions (\textbf{IV}), $R^2_{\text{lin}}$ degrades, yet remains higher than for truly non-linear two-way interactions.
For $R^2_{\text{prod}}$, any higher-order interaction causes considerable degradation, even for features not involved in a higher-order term. Misspecified surrogates no longer capture interaction shapes consistently across intervals, causing the ratio to break.
Notably,~both measures are conservative, rarely claiming linearity/product-separability when absent.

\subsection{Interaction Visualization Results}\label{sec:eval-vis}

\begin{figure}[tb]
    \centering
    \begin{subfigure}[t]{0.4\textwidth}
        \centering
        \includegraphics[width=\textwidth]{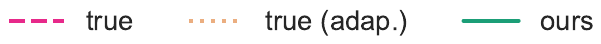}
    \end{subfigure}
   \vskip 0.0\baselineskip
    \begin{subfigure}[t]{0.32\textwidth}
        \centering
        \includegraphics[width=\textwidth]{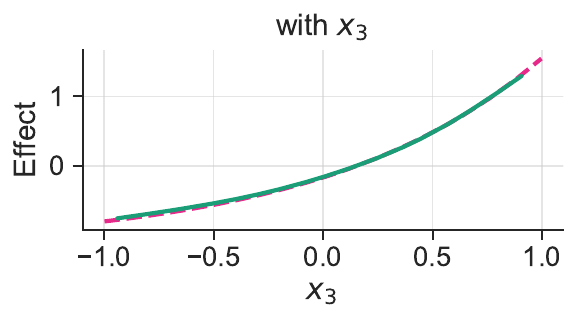}
        \caption{Setting \textbf{I}: linear.}\label{fig:vis-lin-i}
    \end{subfigure}
    \hfill
    \begin{subfigure}[t]{0.32\textwidth}
        \centering
        \includegraphics[width=\textwidth]{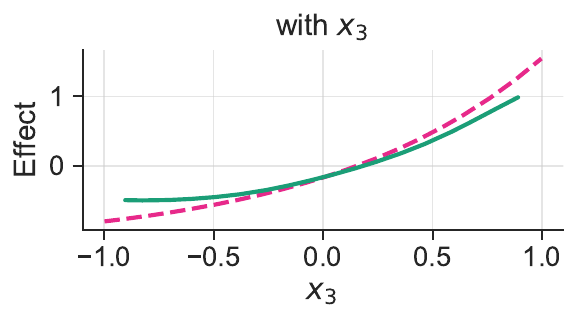}
        \caption{Setting \textbf{II}: linear.}\label{fig:vis-lin-ii}
    \end{subfigure}
    \hfill
    \begin{subfigure}[t]{0.32\textwidth}
        \centering
        \includegraphics[width=\textwidth]{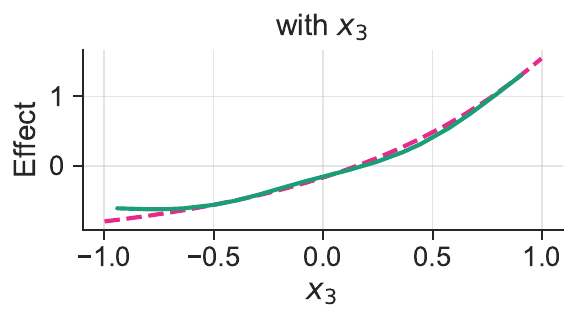}
        \caption{Setting \textbf{III}: linear.}\label{fig:vis-lin-iii}
    \end{subfigure}
    \begin{subfigure}[t]{0.595\textwidth}
        \centering
        \includegraphics[width=\textwidth]{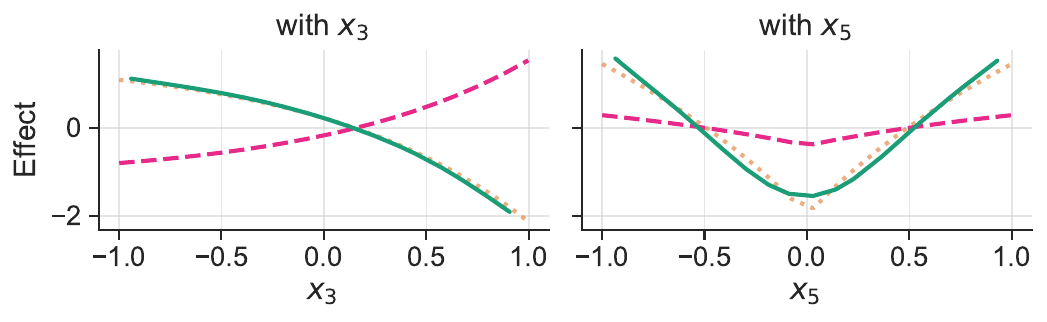}
        \caption{Setting \textbf{I}: ratio-based.}\label{fig:vis-ps-i}
    \end{subfigure}
    \hfill
    \begin{subfigure}[t]{0.325\textwidth}
        \centering
        \includegraphics[width=\textwidth]{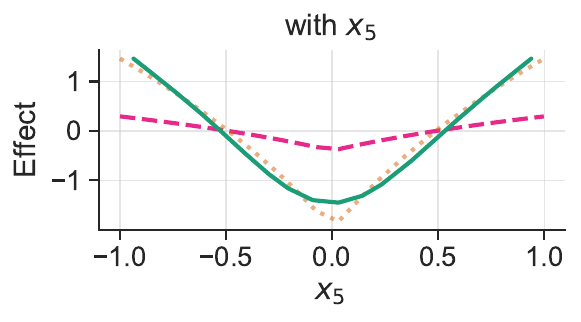}
        \caption{Setting \textbf{II}: ratio-based.}\label{fig:vis-ps-ii}
    \end{subfigure}
    \caption{Linear and product-separable interaction visualizations for the oracle (last repetition).
    True is $\phi_l(x_l)$ (cf. Def.~\ref{def:linear}~\&~\ref{def:prod-sep}) as denoted in Tab.~\ref{tab:settings}.
    Generally, $\phi_l$ is not uniquely defined.
    For ratio-based visualizations, the result depends on $\tilde{x}_l^\text{ref}$. True (adap.) is rescaled accordingly as  $\phi_l/\phi_l(\tilde{x}_l^\text{ref})$.
    Full results in \appref{app:further-results-lin-prod-visualization}.}
\end{figure}

\begin{figure}[tb]
    \centering
    \begin{subfigure}[t]{\textwidth}
        \centering
        \includegraphics[width=0.98\textwidth]{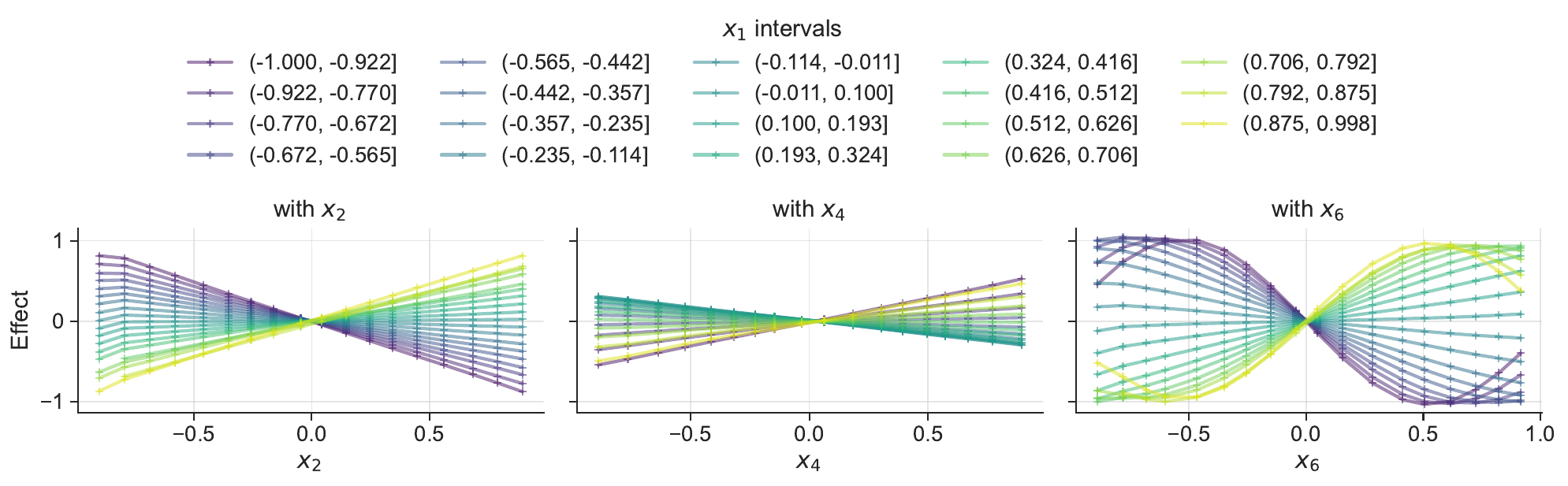}
        \caption{Setting \textbf{I} (two-way, independent).}\label{fig:vis-integrated-i}
    \end{subfigure}
    \begin{subfigure}[t]{\textwidth}
        \centering
        \includegraphics[width=0.98\textwidth]{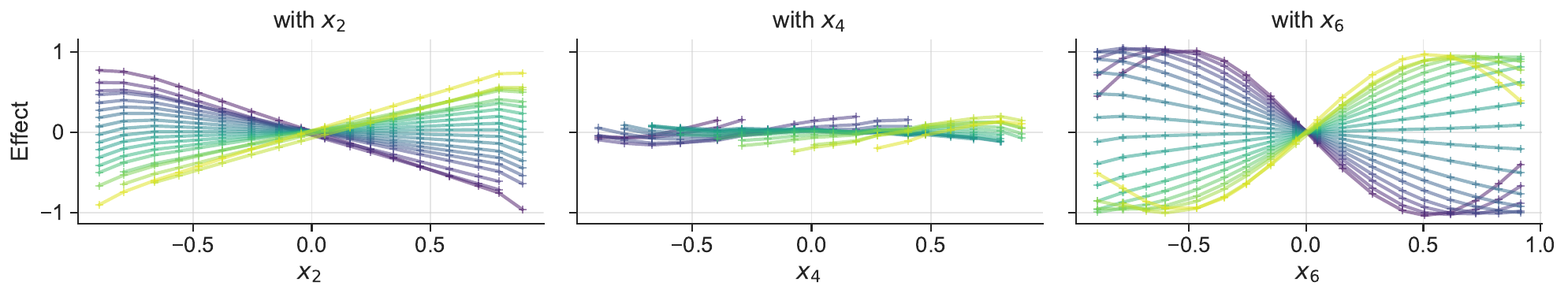}
        \caption{Setting \textbf{II} (two-way, correlated).}\label{fig:vis-integrated-ii}
    \end{subfigure}
    \begin{subfigure}[t]{\textwidth}
        \centering
        \includegraphics[width=0.98\textwidth]{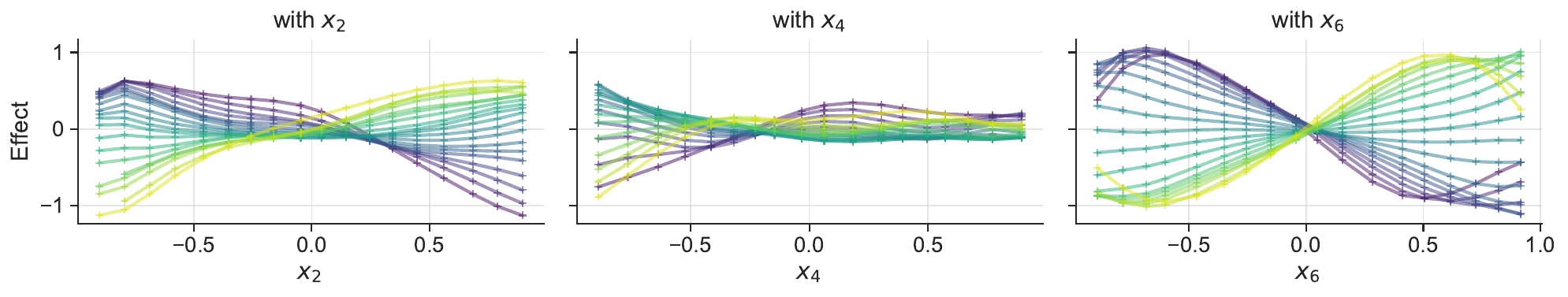}
        \caption{Setting \textbf{IV} (complex higher-order interactions).}\label{fig:vis-integrated-iv}
    \end{subfigure}
    \caption{General interaction visualization (integrated smooths) for the oracle (last repetition).
      Each curve corresponds to an interval of $X_1$.
      Full results in \appref{app:further-results-general-visualization}.}\label{fig:vis-integrated}
\end{figure}

To evaluate \ours{} interaction visualizations, we compare (1) linear interaction visualizations for linear interactions ($R^2_\text{lin} \geq 0.9$), (2) product-separable interaction visualizations for product-separable interactions ($R^2_\text{prod} \geq 0.9$), and (3) general visualizations for any detected interactions ($\alpha=0.05$) to the respective ground truths.
We focus on the representative last repetition per setting.

\textit{Linear Visualization.} For detected linear interactions, under two-way interactions and independence (Setting \textbf{I}), near perfect visualization is possible (Fig.~\ref{fig:vis-lin-i}). Under correlations (\textbf{II}), the visualization becomes less accurate due to the varying centering issue (Fig.~\ref{fig:vis-lin-ii}).
When linear higher-order interactions are present (\textbf{III}), the visualization is close to the ground truth but less smooth (Fig.~\ref{fig:vis-lin-iii}), potentially an artifact of surrogate misspecification.
With non-linear higher-order interactions (\textbf{IV}), no interactions are detected as linear.

\textit{Product-Separable Visualization.} For detected product-separable interactions, ratio-based visualization accurately recovers the (rescaled) true form under independence and two-way interactions (\textbf{I}). However, the surrogate GAMs have issues at the non-smooth $X_5=0$ (Fig.~\ref{fig:vis-ps-i}).
Under dependence (\textbf{II}), no product-separability is found for the correlated $X_3$, the uncorrelated $X_5$ remains unchanged (Fig.~\ref{fig:vis-ps-ii}).
In \textbf{III} \& \textbf{IV}, no product-separable interactions are found.
Thus, linear/product-separable visualizations inherit limitations of $R^2_\text{lin}$/$R^2_\text{prod}$.

\textit{General Visualization.} Under independence and two-way interactions (\textbf{I}), the general visualizations (Fig.~\ref{fig:vis-integrated-i}) closely match the ground truth effects provided in \appref{app:vis-ground-truth}.
$X_1 \times X_2$ is linear in both features (linear interaction),
$X_1 \times X_4$ shows linear curves that scale quadratically with $X_1$ (product-separable),
and $X_1 \times X_6$ captures the sine-shaped pattern accurately (non-product-separable).
Correlations (\textbf{II}) have minimal impact for weakly correlated features ($X_2$)
but reduce the support of interval-specific curves for highly correlated ones ($X_4$).
While intentional to avoid extrapolation, this also makes it more difficult to capture the curves accurately (Fig.~\ref{fig:vis-integrated-ii}).
Higher-order interactions (\textbf{IV}) introduce artifacts from surrogate misspecification,
and visualizations become mixtures of the two-way and higher-order effects (Fig.~\ref{fig:vis-integrated-iv}),
which may still be informative.

\section{Application: Power Consumption Prediction}\label{sec:application}

We apply \ours{} to a real-world power consumption dataset\footnote{The dataset is available at \url{https://archive.ics.uci.edu/dataset/849/power+consumption+of+tetouan+city} (last accessed: 05/22/2026)} from Tetouan, Morocco~\cite{salam-irsec18a}.
The dataset contains 52,417 observations at ten-minute intervals for the year 2017,
with features \texttt{Temperature}, \texttt{general\_diffuse\_flows} (total solar radiation), \texttt{diffuse\_flows} (scattered radiation), \texttt{Humidity}, \texttt{WindSpeed},
and a timestamp.
Power consumption is recorded for three zones.
We additionally extract the \texttt{Hour} of the day and the \texttt{UnixTime} as features.
We use hourly Zone~1 power consumption as target, aggregating the ten-minute values to hourly sums (of the target) and averages (of the features) beforehand.
The first $\approx$10 months are used as training (85\%) and the last $\approx$2 months as test data (15\%).

On the training set, we tune an XGBoost for 200 trials with a Tree-structured Parzen Estimator (TPE)~\cite{bergstra-neurips11a} and search space from~\cite{probst-jmlr19a} to minimize the MSE under 3-fold cross-validation.
We retrain the best model on the full training set, which achieves $R^2 = 0.94$ on training and $R^2 = 0.92$ on test set, indicating good generalization.
We apply \ours{} to that model on training data (recommended by~\cite{heiss-arxiv26a} for feature effects) with \texttt{general\_diffuse\_flows} as FOI.  % chktex 13
An average in-sample $R^2$ of $0.96$ across the GAM surrogates indicates that they capture most of the local effect variation. The remainder is potentially due to higher-order interactions or difficulties with XGBoost's piecewise-constant nature.

\begin{table}[tb]
    \centering
    \scriptsize
    \caption{Interaction detection and categorization results of \ours{} for XGBoost on the Tetouan city dataset with \texttt{general\_diffuse\_flows} as FOI.}
    \label{tab:application}
    \begin{tabular}{llcc}
        \toprule
        \textbf{Interacting feature} & \textbf{approx.\ p-value} & \textbf{Linearity} $R^2_{\text{lin}}$ & \textbf{Prod.\ sep.} $R^2_{\text{prod}}$ \\
        \midrule
        \texttt{Hour}           & $1.78 \times 10^{-250}$ & 0.596 & 0.271 \\
        \texttt{Temperature}    & $4.37 \times 10^{-119}$ & 0.191 & 0.229 \\
        \texttt{UnixTime}       & $1.32 \times 10^{-63}$  & 0.174 & 0.111 \\
        \texttt{diffuse\_flows} & $6.19 \times 10^{-32}$  & 0.023 & 0.660 \\
        \texttt{Humidity}       & $1.63 \times 10^{-25}$  & 0.212 & 0.273 \\
        \texttt{WindSpeed}      & $5.55 \times 10^{-10}$  & 0.116 & 0.889 \\
        \bottomrule
    \end{tabular}
\end{table}
\begin{figure}[tb]
    \centering
    \includegraphics[width=0.98\linewidth]{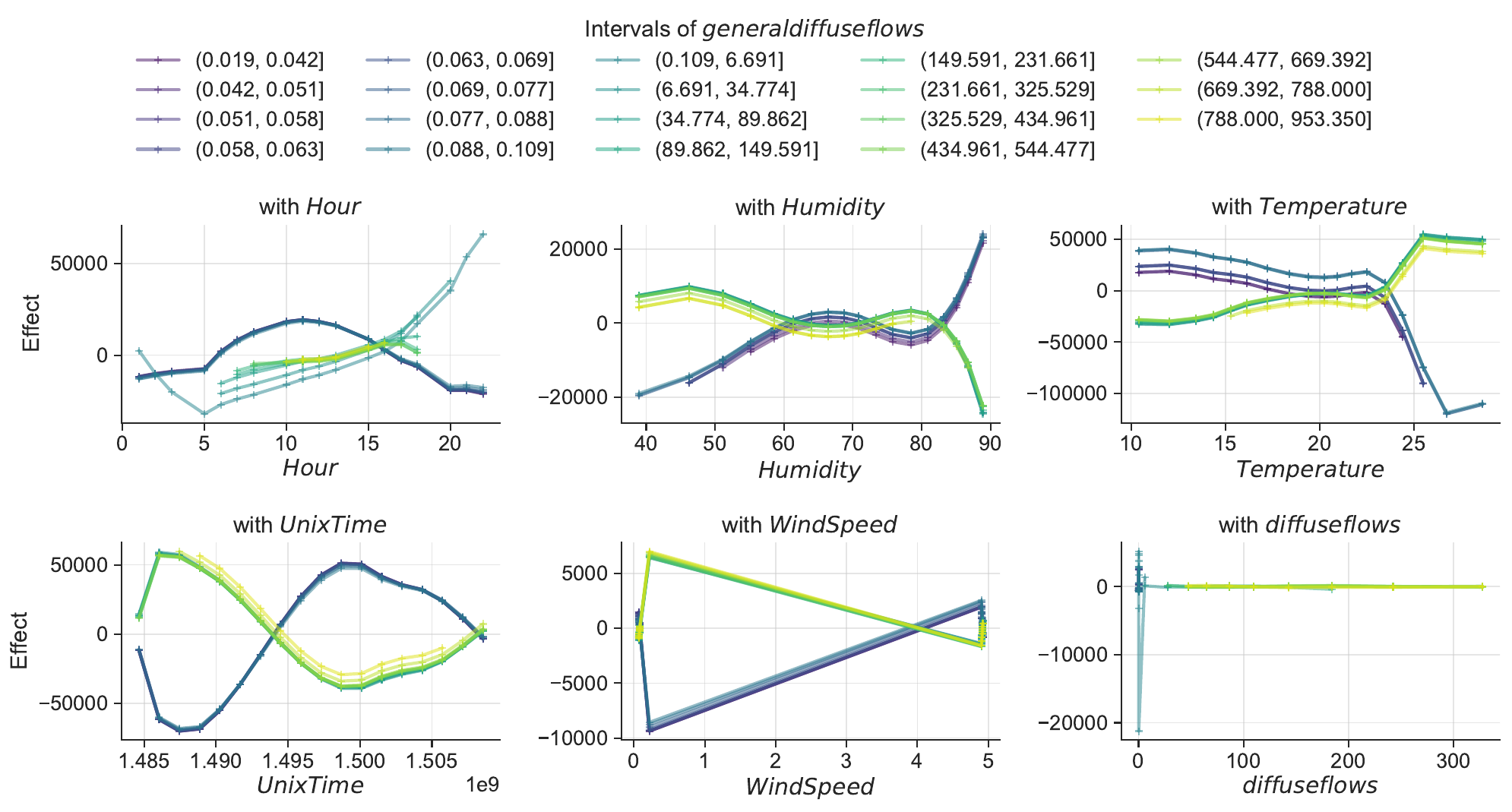}
    \caption{Interaction visualization for \texttt{general\_diffuse\_flows} as FOI of XGBoost on the Tetouan city dataset.
      Each curve corresponds to one interval of \texttt{general\_diffuse\_flows},
      colored from low (purple) to high (yellow) radiation.} \label{fig:application}
\end{figure}

Step (i) in the \ours{} pipeline is to identify interacting features.
Tab.~\ref{tab:application} reports the derived heuristic, indicating that all features interact with the FOI (values considerably below $\alpha = 0.05$).
We therefore (ii) compute $R^2_\text{lin}$ and $R^2_\text{prod}$ for all features.
Both are below $\tau=0.9$ for all features (see Tab.~\ref{tab:application}), motivating the general interaction visualization. However, \texttt{WindSpeed} with $R^2_\text{prod}=0.889$ is close to the threshold, suggesting a near product-separable interaction form.

For (iii) visualization, we apply the general approach in Fig.~\ref{fig:application}, revealing interpretable patterns in the model's interactions.
For \texttt{Hour}, high radiation has almost no interaction effect, while intermediate radiation is associated with increasing power consumption in the evening (dusk) and low radiation with increased daytime consumption (overcast; both conditions require artificial lighting).
High-radiation curves are only observed during daytime, illustrating how \ours{} respects correlations.
For \texttt{Temperature}, the model captures a synergistic pattern:
low radiation combined with low temperature shows increased consumption (simultaneous lighting and heating demand),
as does high radiation with high temperature (cooling demand).
The interaction with \texttt{UnixTime} reflects seasonal effects:
in summer, low radiation (likely still high ambient temperatures) shows increased consumption (cooling demand);
in winter, high radiation (clear, cold days) has a similar effect (heating needs).
\texttt{WindSpeed} curves indeed each share a similar shape, consistent with the high product-separability score. Overall, the interactions recovered by \ours{} align with plausible demand patterns.

\section{Conclusion}\label{sec:conclusion}

We presented \ours, a model-agnostic framework that goes beyond interaction detection and quantification to characterize the functional form of pairwise interactions.
\ours{} fits interpretable GAM surrogates per interval of a FOI to ALE local effects,
enabling (i)~interaction detection via an aggregated F-test heuristic,
(ii)~categorization into linear, product-separable, and non-product-separable types via $R^2$-based measures,
and (iii)~tailored visualizations, including a general integration-based approach applicable to any two-way interaction.

Empirically, detection is reliable under both independence and correlations, with only marginal degradation under higher-order interactions.
Categorization discriminates interaction types perfectly under two-way independence. While $R^2_\text{lin}$ is relatively robust, $R^2_\text{prod}$ degrades under correlations and higher-order interactions.
All three visualization strategies closely recover interaction forms under independence and degrade gracefully under moderate correlations.
The general integration-based visualization remains an informative approximation even under higher-order interactions.
In a real-world power consumption application, \ours{} revealed interpretable interactions with plausible patterns. A surrogate $R^2$ of $0.96$ suggests that \ours{} captures most of the local effect variation.

\textit{Limitations \& Future Work.}
First, \ours{} currently targets regression tasks, while an extension to classification via scoring or probability outputs is left for future work.
Second, the key theoretical conjecture (that smooth terms approximate FANOVA partial derivatives) lacks rigorous guarantees and remains an open problem.
Third, the categorization measures are sensitive to correlations, which is an open challenge.
Fourth, extending the framework beyond two-way interactions, e.g., via $\text{GA}^2\text{M}$ surrogates, is an important direction.
Finally, \ours{} may further be extended to other local effect methods (e.g., ICE, Shapley values)
and unified with the closely related GADGET for regional effects.
Using model-based tree surrogates may bridge the gap between \ours{} and GADGET.

\ifanonymous\else
\begin{credits}
\subsubsection{\discintname}
The authors have no competing interests to declare.
\end{credits}
\fi

\bibliographystyle{splncs04}
\bibliography{references}

\ifappendix
\clearpage
\appendix
\section{Appendix}
\addcontentsline{toc}{section}{Appendix}

{\small\parskip=2pt
\noindent\hbox to \textwidth{\textbf{\ref{app:proofs}\hspace{0.5em}\hyperref[app:proofs]{Theoretical Evidence}\dotfill\pageref{app:proofs}}}
\noindent\hbox to \textwidth{\hspace{2.5em}\hyperref[app:proof-linearity]{Proof of Proposition~\ref{prop:linearity}}\dotfill\pageref{app:proof-linearity}}
\noindent\hbox to \textwidth{\hspace{2.5em}\hyperref[app:proof-ratio]{Proof of Proposition~\ref{prop:ratio}}\dotfill\pageref{app:proof-ratio}}
\noindent\hbox to \textwidth{\hspace{2.5em}\hyperref[prop:prodsep-corr]{Product-Separability Measure Under Correlations}\dotfill\pageref{prop:prodsep-corr}}
\noindent\hbox to \textwidth{\hspace{2.5em}\hyperref[app:fanova-vanishing]{FANOVA Vanishing Conditions for Product-Separable Interactions}\dotfill\pageref{app:fanova-vanishing}}
\noindent\hbox to \textwidth{\textbf{\ref{app:experiment-details}\hspace{0.5em}\hyperref[app:experiment-details]{Experiment Details}\dotfill\pageref{app:experiment-details}}}
\noindent\hbox to \textwidth{\hspace{2.5em}\hyperref[app:data-generation]{Data Generation Details}\dotfill\pageref{app:data-generation}}
\noindent\hbox to \textwidth{\hspace{2.5em}\hyperref[app:ml-models]{ML Models \& Tuning}\dotfill\pageref{app:ml-models}}
\noindent\hbox to \textwidth{\hspace{2.5em}\hyperref[app:parametrization]{\ours{} Parametrization}\dotfill\pageref{app:parametrization}}
\noindent\hbox to \textwidth{\textbf{\ref{app:supp-results}\hspace{0.5em}\hyperref[app:supp-results]{Supplementary Experiment Results}\dotfill\pageref{app:supp-results}}}
\noindent\hbox to \textwidth{\hspace{2.5em}\hyperref[app:model-perf]{ML Model Performance}\dotfill\pageref{app:model-perf}}
\noindent\hbox to \textwidth{\hspace{2.5em}\hyperref[app:surrogate-gof]{Surrogate Goodness-of-Fit}\dotfill\pageref{app:surrogate-gof}}
\noindent\hbox to \textwidth{\textbf{\ref{app:vis-ground-truth}\hspace{0.5em}\hyperref[app:vis-ground-truth]{Ground Truth Two-Way Interaction Effects}\dotfill\pageref{app:vis-ground-truth}}}
\noindent\hbox to \textwidth{\textbf{\ref{app:further-results-detection}\hspace{0.5em}\hyperref[app:further-results-detection]{Further Simulation Results: Interaction Detection}\dotfill\pageref{app:further-results-detection}}}
\noindent\hbox to \textwidth{\textbf{\ref{app:further-results-categorization}\hspace{0.5em}\hyperref[app:further-results-categorization]{Further Simulation Results: Interaction Categorization}\dotfill\pageref{app:further-results-categorization}}}
\noindent\hbox to \textwidth{\textbf{\ref{app:further-results-lin-prod-visualization}\hspace{0.5em}\hyperref[app:further-results-lin-prod-visualization]{Further Simulation Results: Linear \& Product-Sep. Visualization}\dotfill\pageref{app:further-results-lin-prod-visualization}}}
\noindent\hbox to \textwidth{\textbf{\ref{app:further-results-general-visualization}\hspace{0.5em}\hyperref[app:further-results-general-visualization]{Further Simulation Results: General Interaction Visualization}\dotfill\pageref{app:further-results-general-visualization}}}
}\vspace{2pt}
\clearpage

\subsection{Theoretical Evidence}\label{app:proofs}

\subsubsection{Proof of Proposition~\ref{prop:linearity}.}\label{app:proof-linearity}
  \emph{($\Rightarrow$)}
  If $g_{jl}(x_j, x_l) = (x_j - \expectation[\featureRV{j}]) \cdot \phi_l(x_l)$,
  then $\partial_j g_{jl}(x_j, x_l) = \phi_l(x_l)$, which is independent of $x_j$.

  \emph{($\Leftarrow$)}
  Suppose $\partial_j g_{jl}(x_j, x_l) = \phi_l(x_l)$ is independent of $x_j$.
  Integrating w.r.t.\ $x_j$ yields
  $$g_{jl}(x_j, x_l) = x_j \cdot \phi_l(x_l) + C(x_l)$$
  for some univariate function $C(x_l)$.
  Applying the FANOVA vanishing condition $\int g_{jl}(x_j, x_l) \, d\distribution{\featureRV{j}}(x_j) = 0$ gives
  $$\expectation[\featureRV{j}] \cdot \phi_l(x_l) + C(x_l) = 0 \iff
  C(x_l) = -\expectation[\featureRV{j}] \cdot \phi_l(x_l).$$
  Hence $g_{jl}(x_j, x_l) = (x_j - \expectation[\featureRV{j}]) \cdot \phi_l(x_l)$,
  which is the linear form.
  The condition $\expectation[\phi_l(\featureRV{l})] = 0$ follows from the vanishing condition in $X_l$.

\subsubsection{Proof of Proposition~\ref{prop:ratio}.}\label{app:proof-ratio}
  \emph{($\Rightarrow$)}
  If $g_{jl}(x_j, x_l) = \phi_j(x_j) \cdot \phi_l(x_l)$,
  then $\partial_j g_{jl}(x_j, x_l) = \phi_j'(x_j) \cdot \phi_l(x_l)$ and the ratio equals
  $$\frac{\phi_j'(x_j) \cdot \phi_l(x_l)}{\phi_j'(x_j) \cdot \phi_l(\tilde{x}_l^{\text{ref}})} = \frac{\phi_l(x_l)}{\phi_l(\tilde{x}_l^{\text{ref}})},$$
  which is independent of $x_j$.

  \emph{($\Leftarrow$)}
  Suppose the ratio $\partial_j g_{jl}(x_j, x_l) \,/\, \partial_j g_{jl}(x_j, \tilde{x}_l^{\text{ref}})$ is independent of $x_j$.
  Then $\partial_j g_{jl}$ must factor as $a(x_j) \cdot b(x_l)$ for some functions $a, b$.
  Integrating w.r.t.\ $x_j$, with $A$ such that $A' = a$:
  $$g_{jl}(x_j, x_l) = A(x_j) \cdot b(x_l) + C(x_l).$$
  Applying the FANOVA vanishing condition $\int g_{jl}(x_j, x_l) \, d\distribution{\featureRV{j}}(x_j) = 0$ gives
  $$\expectation_{\featureRV{j}}[A(\featureRV{j})] \cdot b(x_l) + C(x_l) = 0 \iff C(x_l) = -\expectation_{\featureRV{j}}[A(\featureRV{j})] \cdot b(x_l).$$
  Hence
  $g_{jl}(x_j, x_l) = \bigl(A(x_j) - \expectation_{\featureRV{j}}[A(\featureRV{j})]\bigr) \cdot b(x_l)$,
  which is product-separable with $\phi_j(x_j) = A(x_j) - \expectation_{\featureRV{j}}[A(\featureRV{j})]$ and $\phi_l(x_l) = b(x_l)$.

\subsubsection{Product-Separability Measure Under Correlations.}\label{prop:prodsep-corr}

For a product-sepa-rable interaction $g_{jl}(x_j, x_l) = \phi_j(x_j) \cdot \phi_l(x_l)$,
Eq.~\eqref{eq:smooth-terms} gives
\[
\smooth{l}{k}(x_l) \approx \phi_j'(\bar{z}_{k,j}) \cdot (\phi_l(x_l) - \expectation_{X_{-j}|X_j=\bar{z}_{k,j}}[\phi_l(\featureRV{l})]).
\]
Under independence, the conditional expectation is equal to the marginal expectation, for which we know $\expectation_{\featureRV{l}}[\phi_l(\featureRV{l})] = 0$ by the FANOVA vanishing conditions (see below). Hence, Proposition~\ref{prop:ratio} applies directly to the smooth terms, and $\smooth{l}{k}(x_l) / \smooth{l}{k}(\tilde{x}_l^{\text{ref}})$
equals $\phi_l(x_l) / \phi_l(\tilde{x}_l^{\text{ref}})$, which is independent of $k$.
Under correlations, $\expectation_{\featureRV{-j}| \featureRV{j} = \bar{z}_{k,j}}[\phi_l(\featureRV{l}) ]$ varies with $k$.
The smooth term ratio then depends on $k$ through the conditional expectations (which do not cancel out) and hence depends on $x_j$, causing the product-separability measure to fail.

\subsubsection{FANOVA Vanishing Conditions for Product-Separable Interactions.}\label{app:fanova-vanishing}

The claim $\expectation_{\featureRV{l}}[\phi_l(\featureRV{l})] = 0$ used above follows from the standard FANOVA vanishing conditions.
Under independence, it requires
$\int g_{jl}(x_j, x_l) \, d\distribution{X_l}(x_l) = 0$.
For $g_{jl}(x_j, x_l) = \phi_j(x_j) \cdot \phi_l(x_l)$:
\begin{align*}
  \int g_{jl}(x_j, x_l) \, d\distribution{X_l}(x_l) = 0
  &\Leftrightarrow \expectation_{\featureRV{l}}[\phi_j(x_j) \cdot \phi_l(\featureRV{l})] = 0 \\
  &\Leftrightarrow \phi_j(x_j) \cdot \expectation_{\featureRV{l}}[\phi_l(\featureRV{l})] = 0.
\end{align*}
Since $\phi_j(x_j) \neq 0$ in general, $\expectation_{\featureRV{l}}[\phi_l(\featureRV{l})] = 0$.
For linear interactions with $\phi_j(x_j) = x_j - \expectation[\featureRV{j}]$, the argument is analogous.

\subsection{Experiment Details}\label{app:experiment-details}

\subsubsection{Data Generation Details.}\label{app:data-generation}
The ground truth functions are as stated in Tab.~\ref{tab:settings};
the $x_1^2\log(|x_5|+1)$ term uses a small regularizer $+10^{-10}$ inside the logarithm in the implementation to avoid numerical issues at $x_5 = -1$, which has no effect on the functional form.
Correlations in Setting \textbf{II} are introduced via a Gaussian copula:
(1) independent standard normal features are drawn,
(2) correlations are applied via Cholesky decomposition using the target correlation matrix
$$
\begin{bmatrix}
1.00 & 0.30 & 0.55 & 0.85 & 0.00 & 0.00 & 0.85 & 0.00 & 0.00 \\
0.30 & 1.00 & 0.20 & 0.30 & 0.00 & 0.00 & 0.30 & 0.00 & 0.00 \\
0.55 & 0.20 & 1.00 & 0.50 & 0.00 & 0.00 & 0.50 & 0.00 & 0.00 \\
0.85 & 0.30 & 0.50 & 1.00 & 0.00 & 0.00 & 0.80 & 0.00 & 0.00 \\
0.00 & 0.00 & 0.00 & 0.00 & 1.00 & 0.00 & 0.00 & 0.00 & 0.00 \\
0.00 & 0.00 & 0.00 & 0.00 & 0.00 & 1.00 & 0.00 & 0.00 & 0.00 \\
0.85 & 0.30 & 0.50 & 0.80 & 0.00 & 0.00 & 1.00 & 0.00 & 0.00 \\
0.00 & 0.00 & 0.00 & 0.00 & 0.00 & 0.00 & 0.00 & 1.00 & 0.00 \\
0.00 & 0.00 & 0.00 & 0.00 & 0.00 & 0.00 & 0.00 & 0.00 & 1.00
\end{bmatrix},
$$
(3) each feature is transformed to its target $U(-1,1)$ distribution via the inverse CDF.
The CDF transformation may introduce minor deviations from the target correlations;
empirically these deviations are below $0.05$.
$X_7$ is included as a strongly correlated noise feature ($\rho_{17} \approx 0.85$) to test whether the method correctly distinguishes correlation from interaction.

\subsubsection{ML Models \& Tuning.}\label{app:ml-models}
As briefly mentioned in \S\ref{sec:eval}, we additionally study \ours{} in the simulation settings when using different ML models as the predictor instead of the ground truth directly (``oracle'') (see \S\ref{app:further-results-detection}-\ref{app:further-results-general-visualization}). We consider:
\begin{itemize}[topsep=0pt,leftmargin=*]
    \item \textbf{GAM}: correctly specified GAM, where tensor product splines are included for the true interactions.
    \item \textbf{XGB spec}: XGBoost with interaction constraints set to the true interactions.
    \item \textbf{XGB full}: XGBoost without interaction constraints.
    \item \textbf{SVM-RBF}: support vector machine (SVM) with RBF kernel.
\end{itemize}
This selection covers a range of model classes with different inductive biases and interaction modeling capabilities.
Hyperparameters are tuned beforehand on a separately drawn dataset from the same data-generating process, split into a training and validation set of 1000 observations each.
Tuning is performed by minimizing MSE on the validation set using 200 trials of a Tree-structured Parzen Estimator (TPE)~\cite{bergstra-neurips11a}.
Search spaces for SVM and XGBoost follow~\cite{probst-jmlr19a}.
For the GAM, the number of basis functions (5--50) and the penalty parameter $\lambda$ ($10^{-3}$--$10^{3}$, log scale) are tuned.
In addition, we consider a Random Forest with default hyperparameters as a baseline, which is only used for performance comparison (in \S\ref{app:supp-results}).
In each repetition and setting, the models are trained on the sample of 1000 observations that is also used for \ours{} (i.e., \ours{} is applied on the training sample), as a recent study~\cite{heiss-arxiv26a} suggests that feature effect estimation on training data is appropriate.

\subsubsection{\ours{} Parametrization.}\label{app:parametrization}
The FOI is partitioned into $K = 19$ intervals using 20 quantile-based grid points.
Surrogate GAMs use one centered B-spline smooth term per feature ($d=7$ basis functions, polynomial degree~3)
and a fixed small smoothing penalty of $10^{-5}$.
These are chosen empirically to provide high flexibility for modeling interactions, based on preliminary experiments.
For the linearity and product-separability measures,
a single penalized B-spline ($d=12$ basis functions, degree~3, penalty~$0.05$) is fitted to the pooled evaluations or ratios.
The product-separability reference value is $\tilde{x}_l^{\text{ref}} = -0.8 \ \forall l$.
Smooth terms with variance below 1\% of the local effect variance are excluded.
A practical challenge of the reference point choice is that $\smooth{l}{k}(\tilde{x}_l^{\text{ref}})$ may be near zero for some intervals $k$, rendering the ratio numerically unstable or undefined.
If no valid reference point exists across all smooth terms (e.g., because the smooth term supports do not overlap), $R^2_\text{prod}$ is not reported ($\tilde{x}_l^{\text{ref}} = -0.8$ is chosen as it is unlikely to be zero while still relatively likely to exist across all smooth terms).
% Goodness-of-fit of the surrogate GAMs is assessed via the generalized $R^2$.

\subsection{Supplementary Experiment Results}\label{app:supp-results}

\subsubsection{ML Model Performance.}\label{app:model-perf}
For each repetition, we draw 10{,}000 test samples in addition to the 1000 training samples.
While the ML models are trained on the training set (which is also used for \ours{}), the holdout test set is solely used for performance evaluation of the fitted models.
For the oracle and thus for the results reported in \S\ref{sec:eval}, this is merely a sanity check ($R^2$ is expected to be $\frac{5}{5+1} \approx 0.833$).
However, for the fitted ML models, this is crucial to confirm that the models are adequately fitted to the data and sensible to analyze with \ours{}.
Tab.~\ref{tab:model-perf} shows $R^2$ scores across 30 repetitions for all models.
All tuned models consistently exceed the untuned random forest baseline on the test set,
confirming that all models are adequately fitted for the subsequent interaction analysis.
\begin{table}[t]
    \centering
    \scriptsize
    \setlength{\tabcolsep}{1pt}
    \caption{$R^2$ scores (mean $\pm$ std across 30 repetitions) on training and test set per model and setting.}\label{tab:model-perf}
    \begin{tabular}{lllllllll}
        \toprule
        \textbf{Model}
          & \multicolumn{2}{l}{\textbf{I}}
          & \multicolumn{2}{l}{\textbf{II}}
          & \multicolumn{2}{l}{\textbf{III}}
          & \multicolumn{2}{l}{\textbf{IV}} \\
        & \textit{train} & \textit{test}
          & \textit{train} & \textit{test}
          & \textit{train} & \textit{test}
          & \textit{train} & \textit{test} \\
        \midrule
        Oracle
          & $0.832_{\pm .009}$ & $0.833_{\pm .002}$
          & $0.832_{\pm .009}$ & $0.833_{\pm .002}$
          & $0.832_{\pm .009}$ & $0.833_{\pm .002}$
          & $0.833_{\pm .008}$ & $0.833_{\pm .002}$ \\
        GAM
          & $0.841_{\pm .009}$ & $0.821_{\pm .003}$
          & $0.830_{\pm .009}$ & $0.824_{\pm .002}$
          & $0.826_{\pm .010}$ & $0.818_{\pm .003}$
          & $0.822_{\pm .010}$ & $0.812_{\pm .003}$ \\
        XGB spec
          & $0.888_{\pm .006}$ & $0.805_{\pm .003}$
          & $0.878_{\pm .007}$ & $0.811_{\pm .004}$
          & $0.917_{\pm .004}$ & $0.800_{\pm .003}$
          & $0.899_{\pm .006}$ & $0.799_{\pm .003}$ \\
        XGB full
          & $0.934_{\pm .004}$ & $0.804_{\pm .003}$
          & $0.913_{\pm .005}$ & $0.814_{\pm .003}$
          & $0.931_{\pm .004}$ & $0.802_{\pm .003}$
          & $0.934_{\pm .004}$ & $0.799_{\pm .003}$ \\
        SVM-RBF
          & $0.829_{\pm .009}$ & $0.810_{\pm .004}$
          & $0.831_{\pm .009}$ & $0.818_{\pm .003}$
          & $0.826_{\pm .010}$ & $0.807_{\pm .004}$
          & $0.824_{\pm .010}$ & $0.802_{\pm .004}$ \\
        RF
          & $0.969_{\pm .002}$ & $0.778_{\pm .005}$
          & $0.972_{\pm .002}$ & $0.801_{\pm .004}$
          & $0.968_{\pm .002}$ & $0.776_{\pm .005}$
          & $0.968_{\pm .002}$ & $0.773_{\pm .005}$ \\
        \bottomrule
    \end{tabular}
\end{table}

\subsubsection{Surrogate Goodness-of-Fit.}\label{app:surrogate-gof}
Tab.~\ref{tab:surrogate-gof} reports the generalized $R^2$ (based on deviance) of the surrogate GAMs,
averaged across all intervals and 30 repetitions.
In-sample $R^2$ values are near-perfect across all settings, confirming that the surrogate GAMs have sufficient flexibility to represent local effect variation.
Fits degrade very slightly under higher-order interactions (Settings \textbf{III} \& \textbf{IV}, reflected in the increased standard deviation), as the additive surrogates are misspecified here.
The lowest scores (still 0.999) correspond to XGBoost models, suggesting that smooth surrogate GAMs may struggle somewhat more with the step-function nature of tree-based predictions compared to the smoother SVM and GAM outputs.
Importantly, these scores are computed in-sample on the local effects used to fit the surrogates.
This reflects representational capacity rather than identification accuracy, since in-sample scores cannot detect surrogate overfitting.

\begin{table}[t]
    \centering
    \scriptsize
    \caption{Generalized $R^2$ of surrogate GAMs (mean $\pm$ std across intervals and 30 repetitions) per model and setting.}\label{tab:surrogate-gof}
    \begin{tabular}{lllll}
        \toprule
        \textbf{Model} & \textbf{I} & \textbf{II} & \textbf{III} & \textbf{IV} \\
        \midrule
        Oracle & $1.000_{\pm .000}$ & $1.000_{\pm .000}$ & $1.000_{\pm .000}$ & $1.000_{\pm .001}$ \\
        GAM & $1.000_{\pm .000}$ & $1.000_{\pm .000}$ & $1.000_{\pm .000}$ & $1.000_{\pm .000}$ \\
        XGB spec & $1.000_{\pm .000}$ & $1.000_{\pm .000}$ & $0.999_{\pm .005}$ & $0.999_{\pm .003}$ \\
        XGB full  & $0.999_{\pm .006}$ & $0.999_{\pm .005}$ & $0.999_{\pm .005}$ & $0.999_{\pm .006}$ \\
        SVM-RBF       & $1.000_{\pm .000}$ & $1.000_{\pm .000}$ & $1.000_{\pm .000}$ & $1.000_{\pm .000}$ \\
        \bottomrule
    \end{tabular}
\end{table}

\subsection{Ground Truth Two-Way Interaction Effects}\label{app:vis-ground-truth}

In Fig.~\ref{fig:gt-interaction}, we provide visualizations of the true underlying two-way interaction effects from our simulation experiments in \S\ref{sec:eval}. Note that these are the (isolated) two-way interactions over the entire feature space (i.e., under feature independence).
Thus, they are denoted by the function components:
\[
X_1X_2, \ X_1\exp(X_3), \ X_1^2X_4, \ X_1^2 \log(|X_5|+1), \ \sin(\pi X_1 X_6).
\]
We visualize them by evaluating on a 2D grid of the center of each interval for the FOI (same as in the experiments) and quantile-based grid points for the interacting feature (similar to our method). We ``doubly center'' each term.

\begin{figure}[h]
    \centering
    \includegraphics[width=1\linewidth]{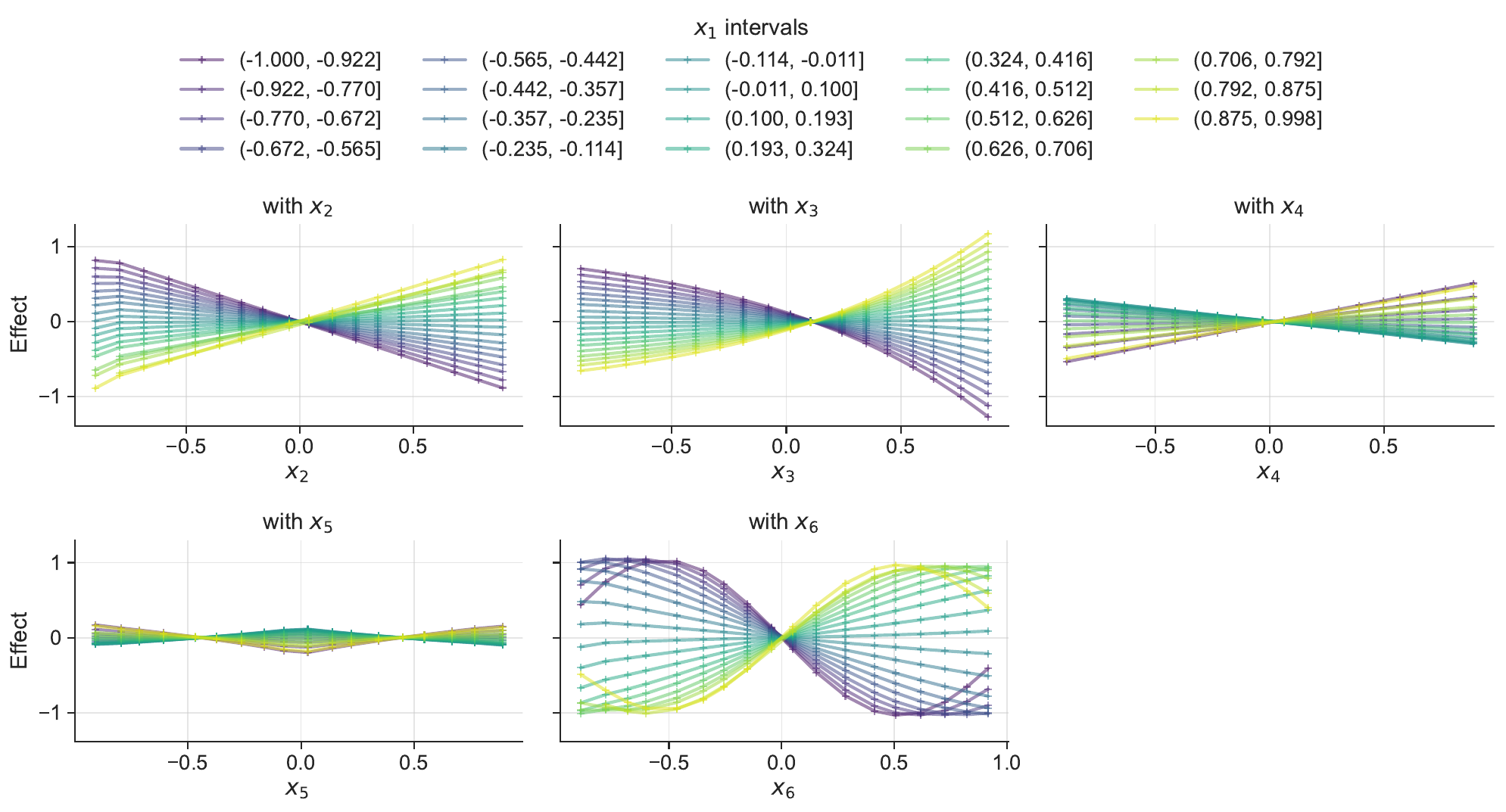}
    \caption{Visualization of the true underlying two-way interaction effects in our simulation experiments.}\label{fig:gt-interaction}
\end{figure}

\FloatBarrier
\subsection{Further Simulation Results: Interaction Detection}\label{app:further-results-detection}

The full results for our interaction detection experiments are provided in Fig.~\ref{fig:significance-full-results}, for the oracle and the fitted ML models.
For the GAM and XGBoost models, where we restricted possible interactions to the true interactions, the detection heuristics adequately reflect these constraints in most cases.
We observe larger variances in the values for the full XGBoost, potentially indicating that GAM surrogates struggle with tree-based models, or simply reflecting model variance.
For the SVM-RBF, we often detect interactions with all features.
When higher-order interactions are present, we observe larger variances and more values above $\alpha$. However, since we do not know the true interactions learned by the ML models, validating these patterns remains difficult.

\begin{figure*}[htbp]
    \centering
    \begin{subfigure}[t]{\textwidth}
        \centering
        \includegraphics[width=0.9\textwidth]{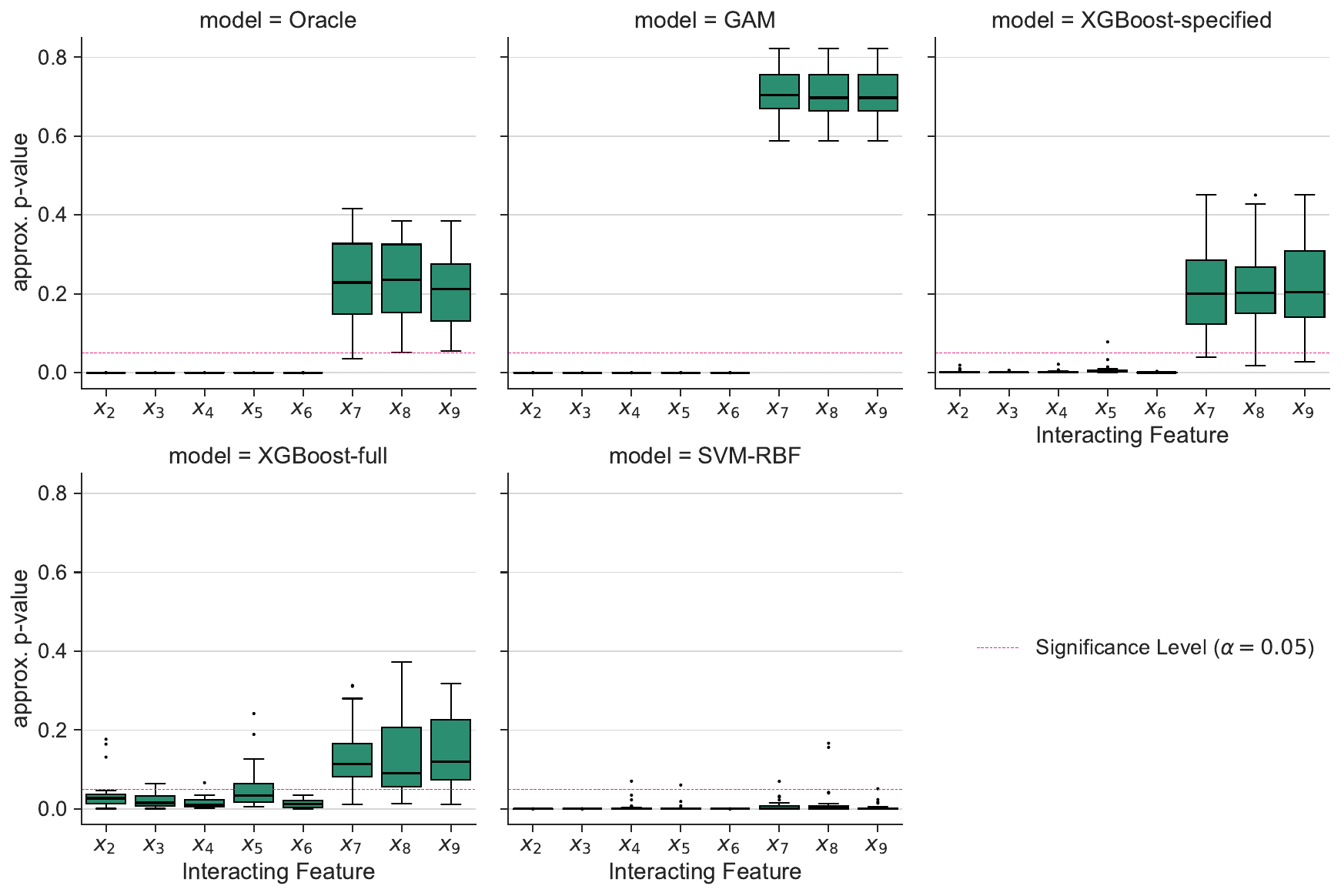}
        \caption{Interaction detection results for setting \textbf{I} (two-way, independent).}
    \end{subfigure}
\end{figure*}

\begin{figure*}[htbp]
    \centering
    \ContinuedFloat
    \begin{subfigure}[t]{\textwidth}
        \centering
        \includegraphics[width=0.9\textwidth]{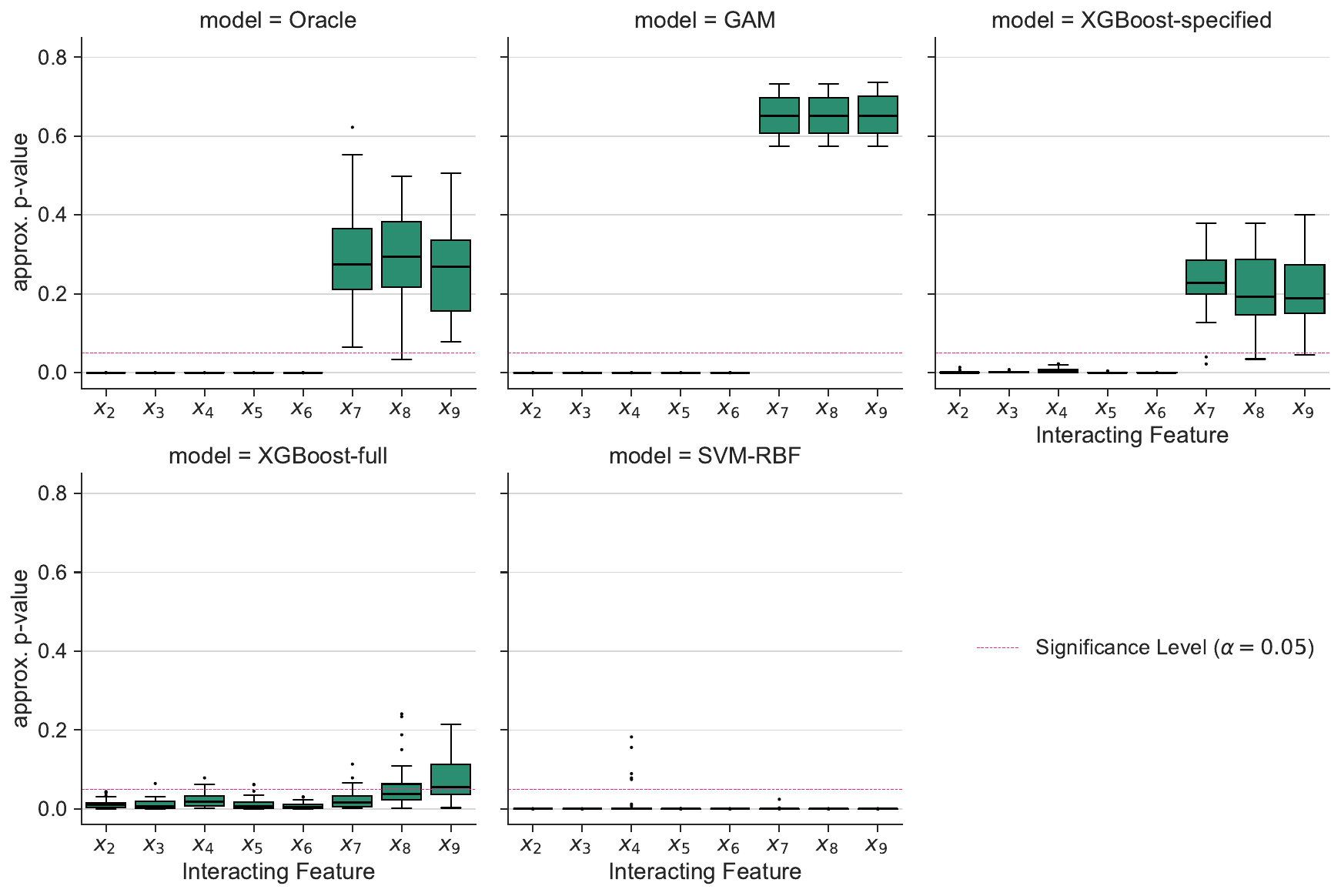}
        \caption{Interaction detection results for setting \textbf{II} (two-way, correlated).}
    \end{subfigure}
\end{figure*}

\begin{figure*}[htbp]
    \centering
    \ContinuedFloat
    \begin{subfigure}[t]{\textwidth}
        \centering
        \includegraphics[width=0.9\textwidth]{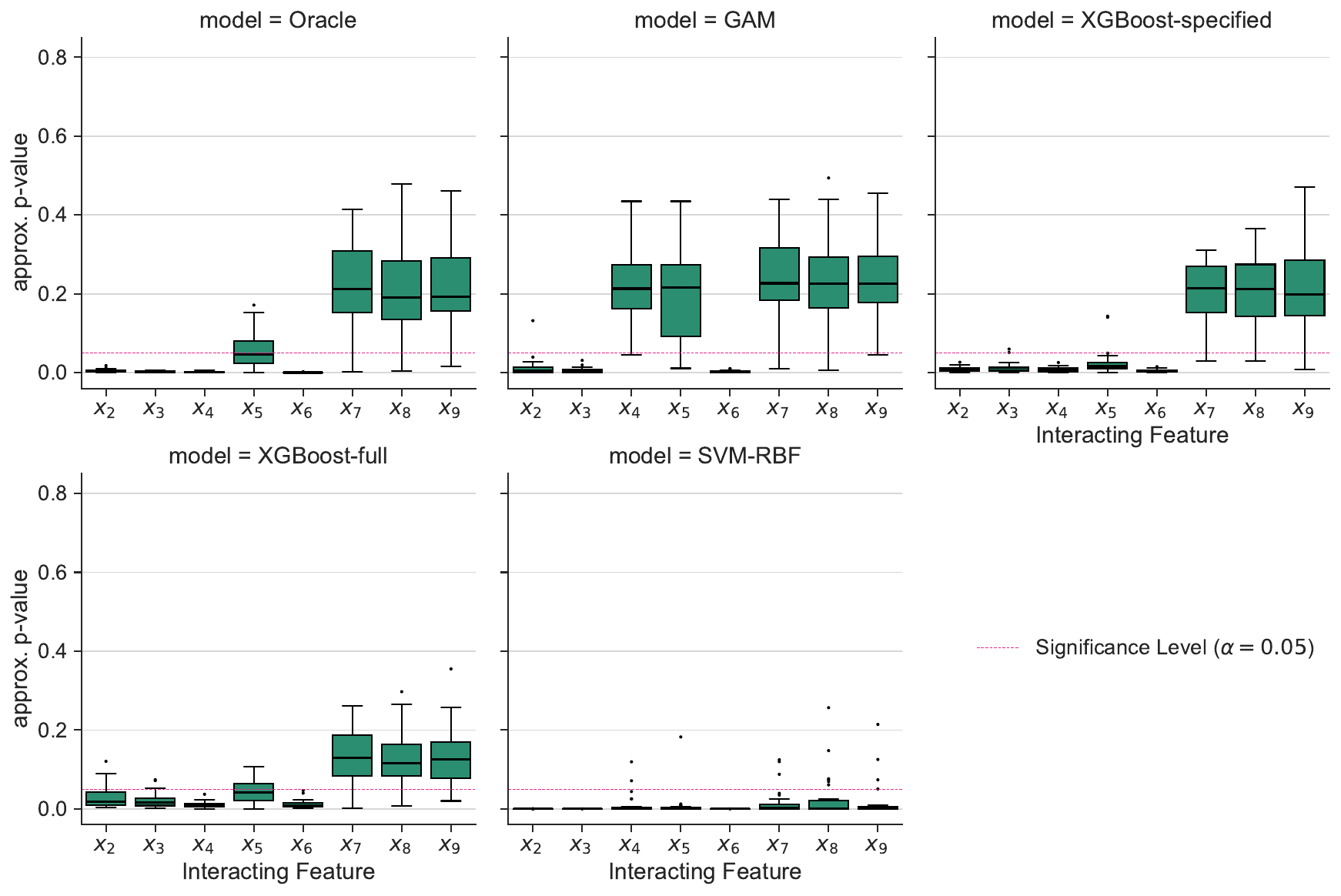}
        \caption{Interaction detection results for setting \textbf{III} (higher-order simple).}
    \end{subfigure}
\end{figure*}

\begin{figure*}[htbp]
    \centering
    \ContinuedFloat
    \begin{subfigure}[t]{\textwidth}
        \centering
        \includegraphics[width=0.9\textwidth]{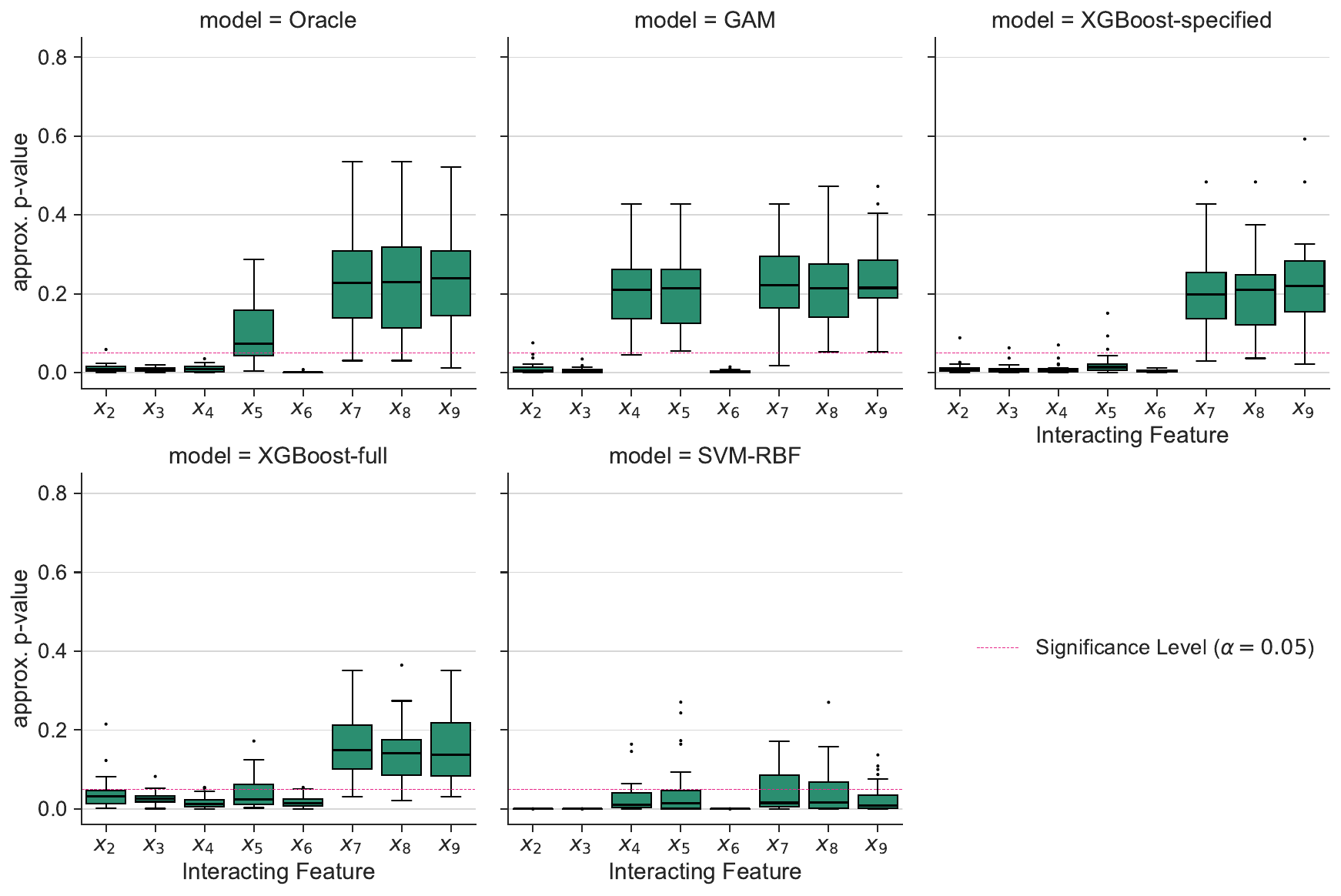}
        \caption{Interaction detection results for setting \textbf{IV} (higher-order complex).}
    \end{subfigure}
    \caption{Full simulation results for \ours{} interaction detection. Approximate p-values across 30 repetitions are visualized as boxplots. Dashed line: threshold of $\alpha=0.05$.}\label{fig:significance-full-results}
\end{figure*}

\FloatBarrier
\subsection{Further Simulation Results: Interaction Categorization}\label{app:further-results-categorization}

The full results for our interaction categorization measures are shown in Figs.~\ref{fig:cat-lin-full-results} \&~\ref{fig:cat-prod-full-results}, for the oracle and the fitted ML models.
The GAM often learns linear and product-separable interactions according to our measures, sometimes even when the true underlying effect is not product-separable.
For the SVM, our measures report that linear (and thus also product-separable) interactions are frequently learned despite the underlying effect being non-linear.
XGBoost models never learn linear interactions and only in single cases product-separable ones, which is not surprising considering the piecewise-constant nature of tree-based models.

\begin{figure*}[htbp]
    \centering
    \begin{subfigure}[t]{\textwidth}
        \centering
        \includegraphics[width=0.9\textwidth]{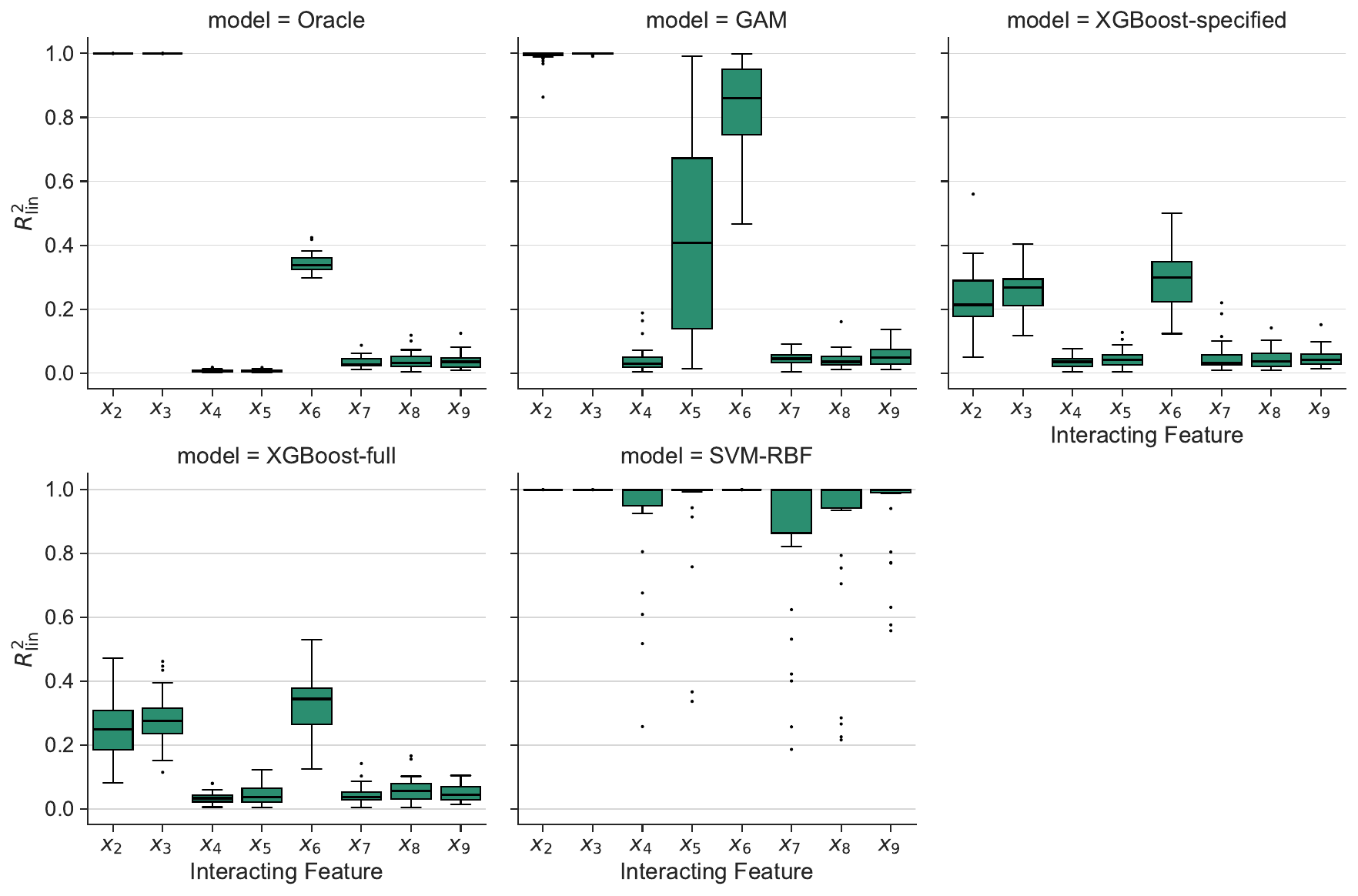}
        \caption{Linearity measure $R^2_\text{lin}$ for setting \textbf{I} (two-way, independent).}
    \end{subfigure}
\end{figure*}

\begin{figure*}[htbp]
    \centering
    \ContinuedFloat
    \begin{subfigure}[t]{\textwidth}
        \centering
        \includegraphics[width=0.9\textwidth]{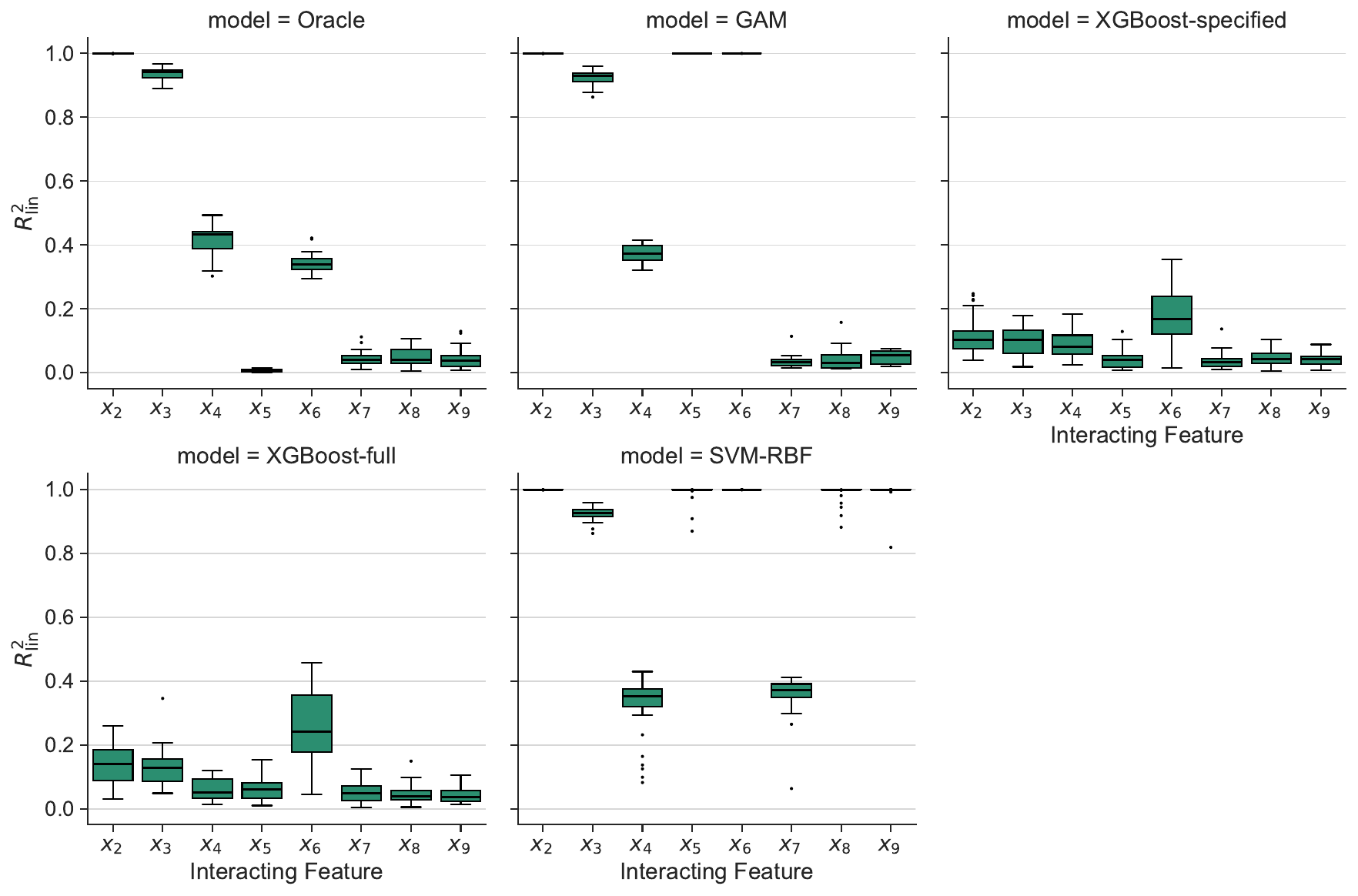}
        \caption{Linearity measure $R^2_\text{lin}$ for setting \textbf{II} (two-way, correlated).}
    \end{subfigure}
\end{figure*}

\begin{figure*}[htbp]
    \centering
    \ContinuedFloat
    \begin{subfigure}[t]{\textwidth}
        \centering
        \includegraphics[width=0.9\textwidth]{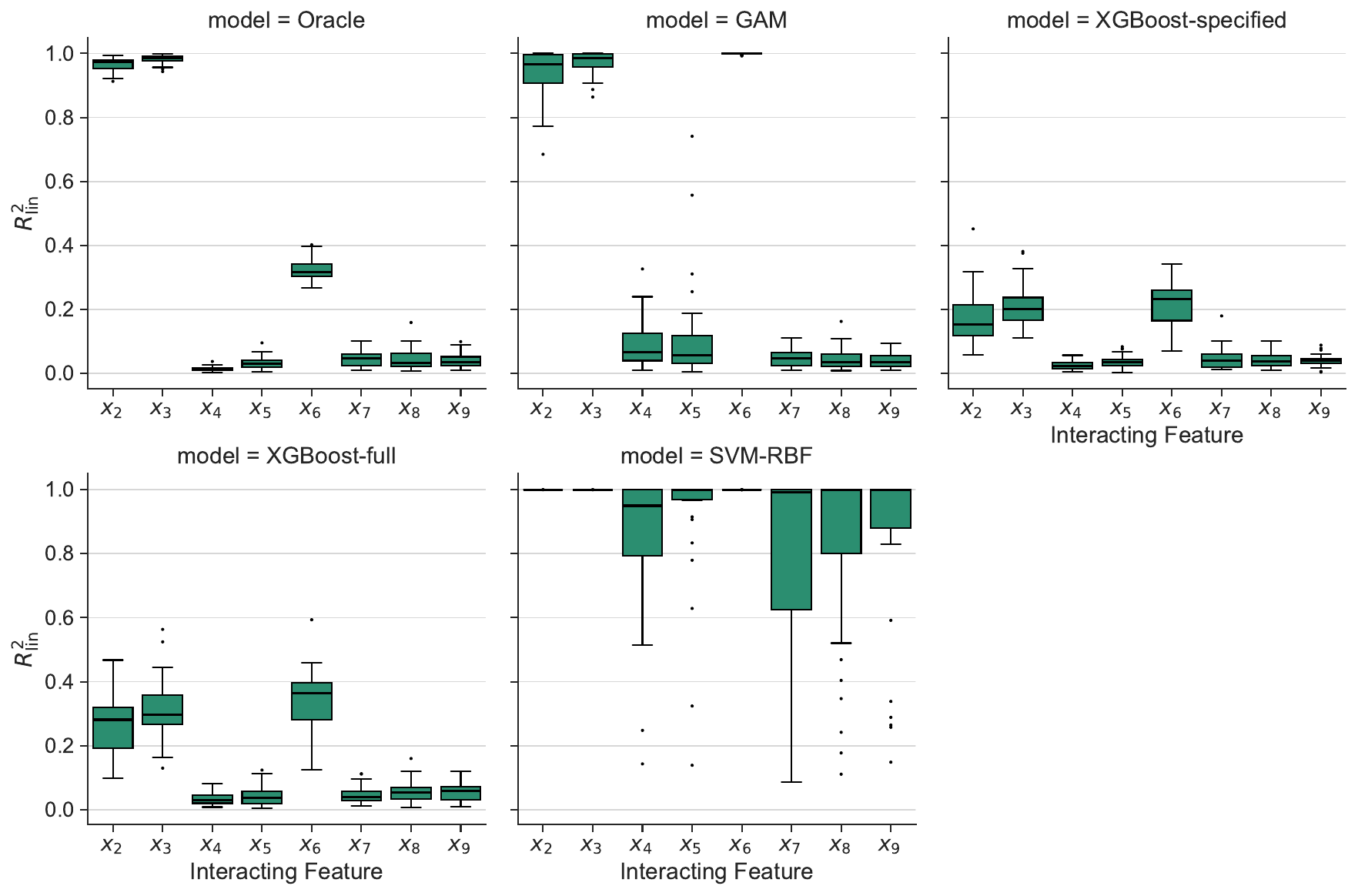}
        \caption{Linearity measure $R^2_\text{lin}$ for setting \textbf{III} (higher-order simple).}
    \end{subfigure}
\end{figure*}

\begin{figure*}[htbp]
    \centering
    \ContinuedFloat
    \begin{subfigure}[t]{\textwidth}
        \centering
        \includegraphics[width=0.9\textwidth]{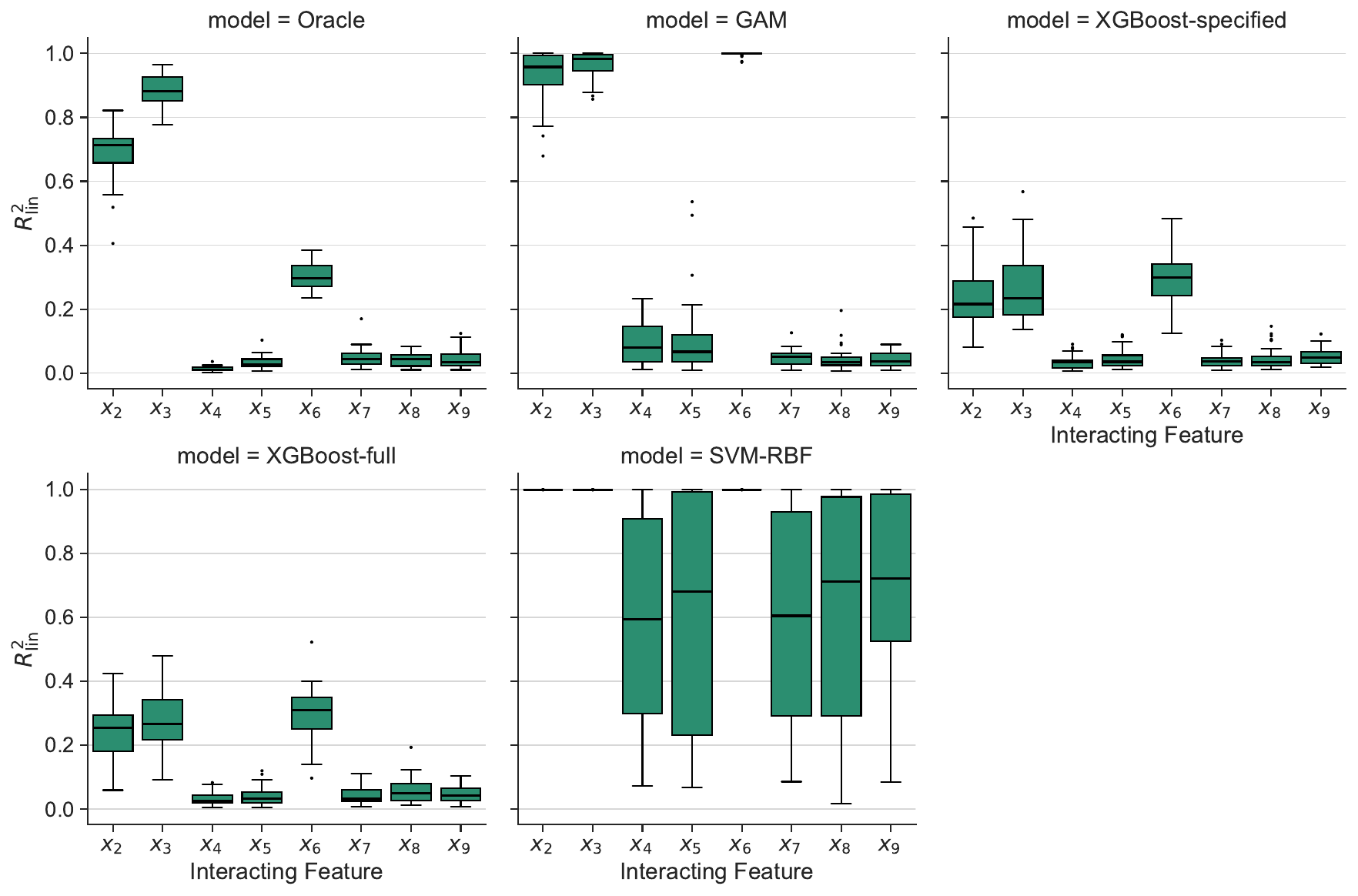}
        \caption{Linearity measure $R^2_\text{lin}$ for setting \textbf{IV} (higher-order complex).}
    \end{subfigure}
    \caption{Full simulation results for \ours{} linear interaction categorization measure. $R^2_\text{lin}$ across 30 repetitions visualized as boxplots.}
    \label{fig:cat-lin-full-results}
\end{figure*}

\begin{figure*}[htbp]
    \centering
    \begin{subfigure}[t]{\textwidth}
        \centering
        \includegraphics[width=0.9\textwidth]{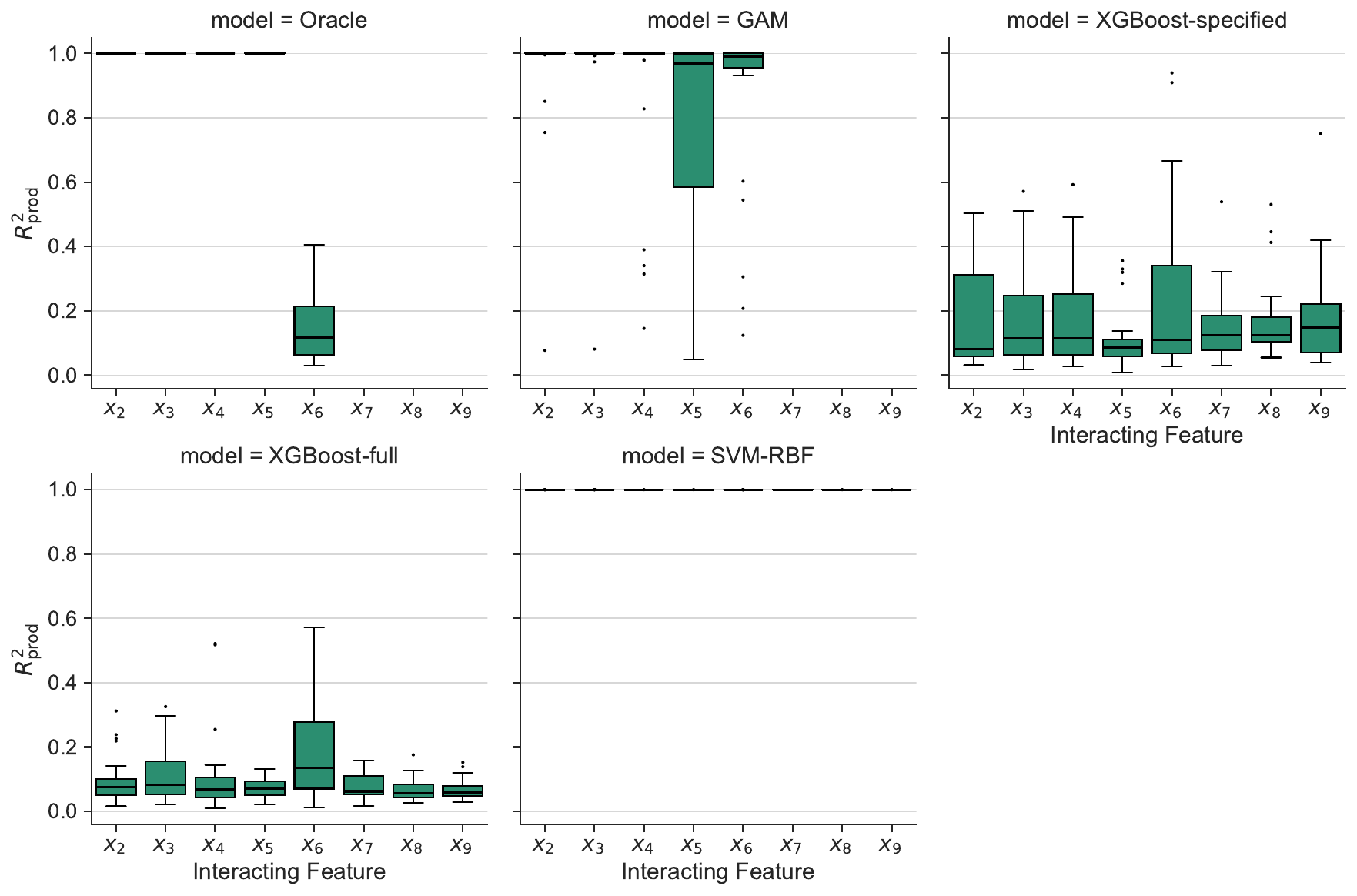}
        \caption{Product-separability measure $R^2_\text{prod}$ for setting \textbf{I} (two-way, independent).}
    \end{subfigure}
\end{figure*}

\begin{figure*}[htbp]
    \centering
    \ContinuedFloat
    \begin{subfigure}[t]{\textwidth}
        \centering
        \includegraphics[width=0.9\textwidth]{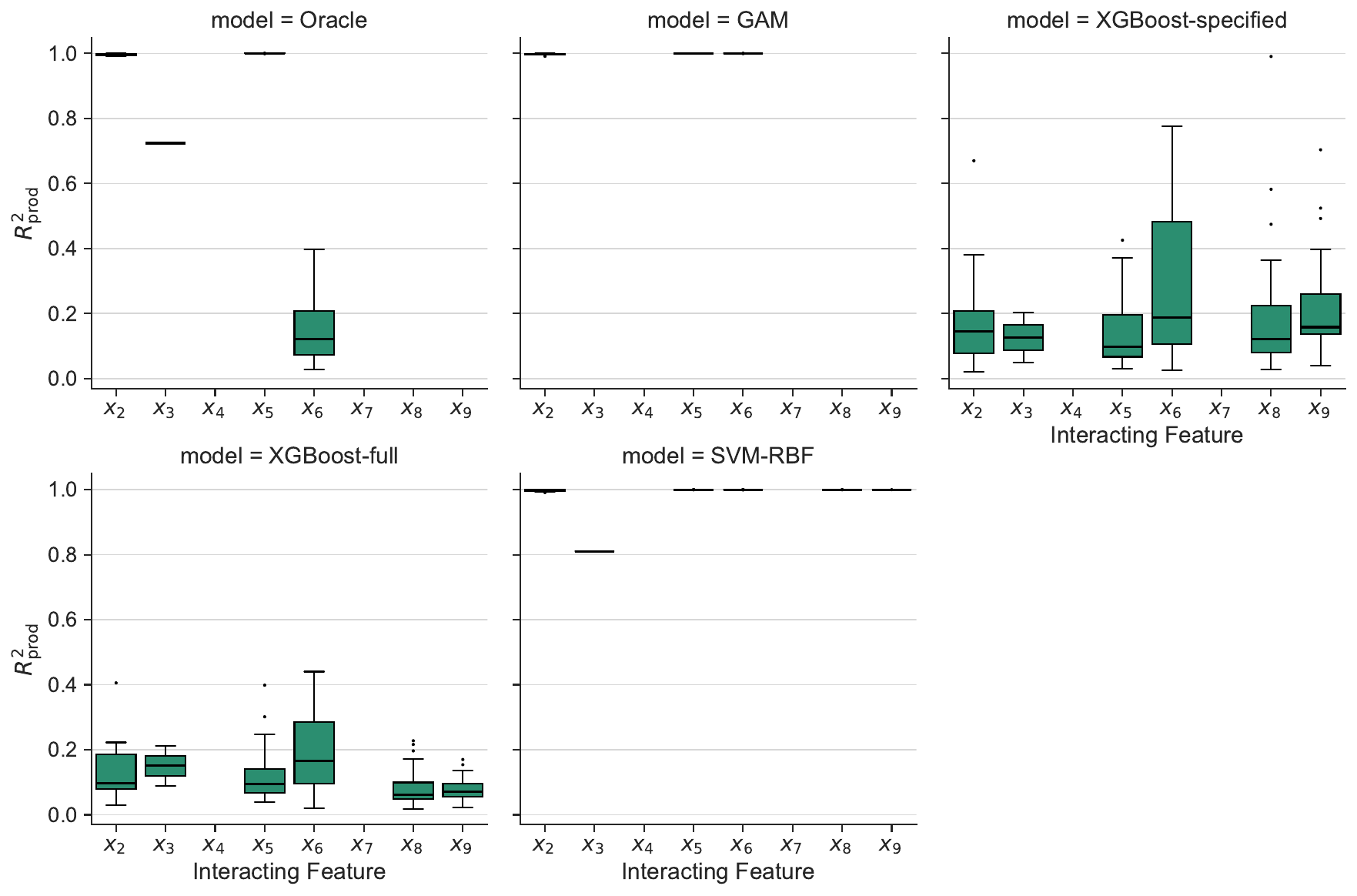}
        \caption{Product-separability measure $R^2_\text{prod}$ for setting \textbf{II} (two-way, correlated).}
    \end{subfigure}
\end{figure*}

\begin{figure*}[htbp]
    \centering
    \ContinuedFloat
    \begin{subfigure}[t]{\textwidth}
        \centering
        \includegraphics[width=0.9\textwidth]{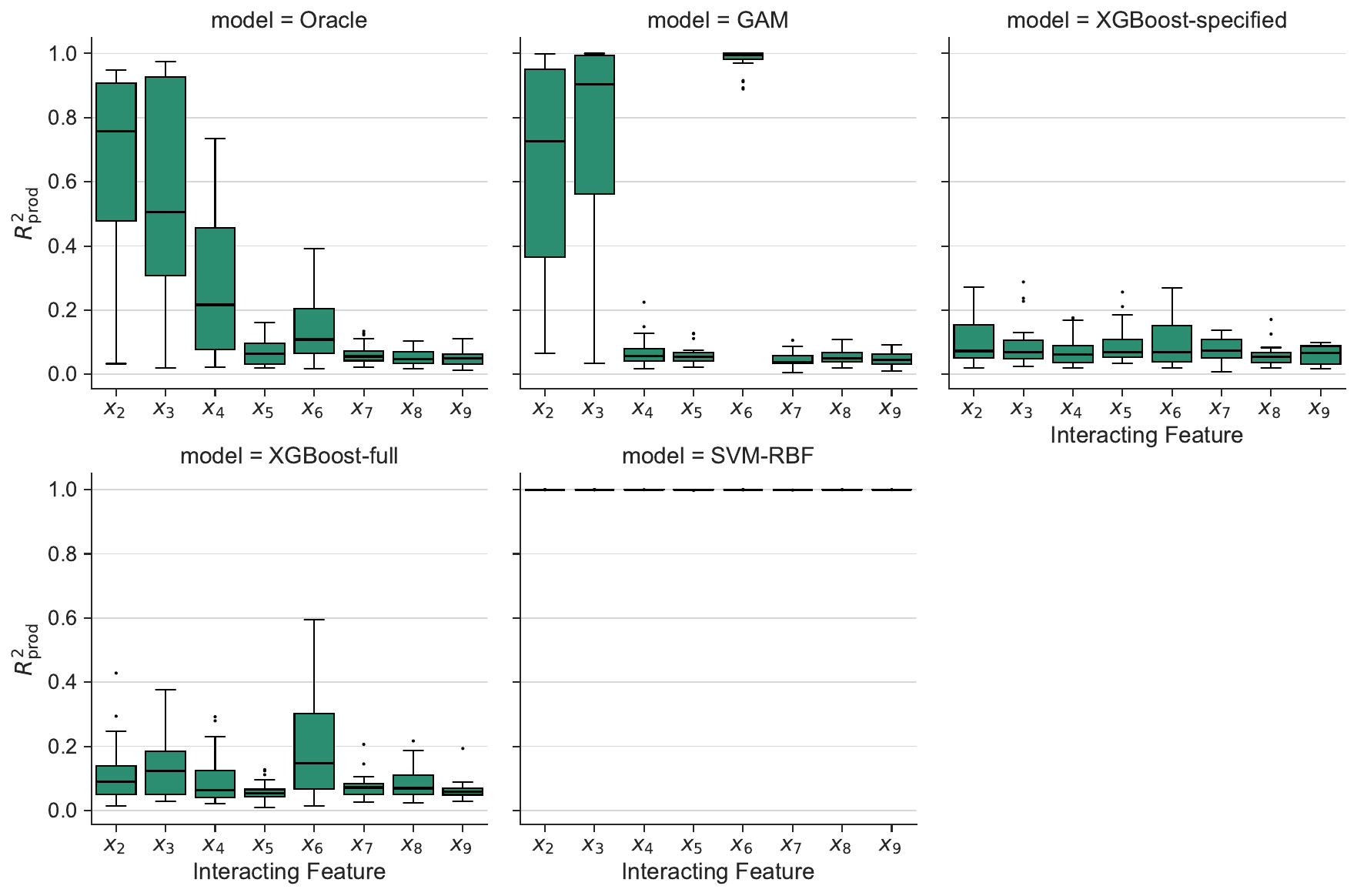}
        \caption{Product-separability measure $R^2_\text{prod}$ for setting \textbf{III} (higher-order simple).}
    \end{subfigure}
\end{figure*}

\begin{figure*}[htbp]
    \centering
    \ContinuedFloat
    \begin{subfigure}[t]{\textwidth}
        \centering
        \includegraphics[width=0.9\textwidth]{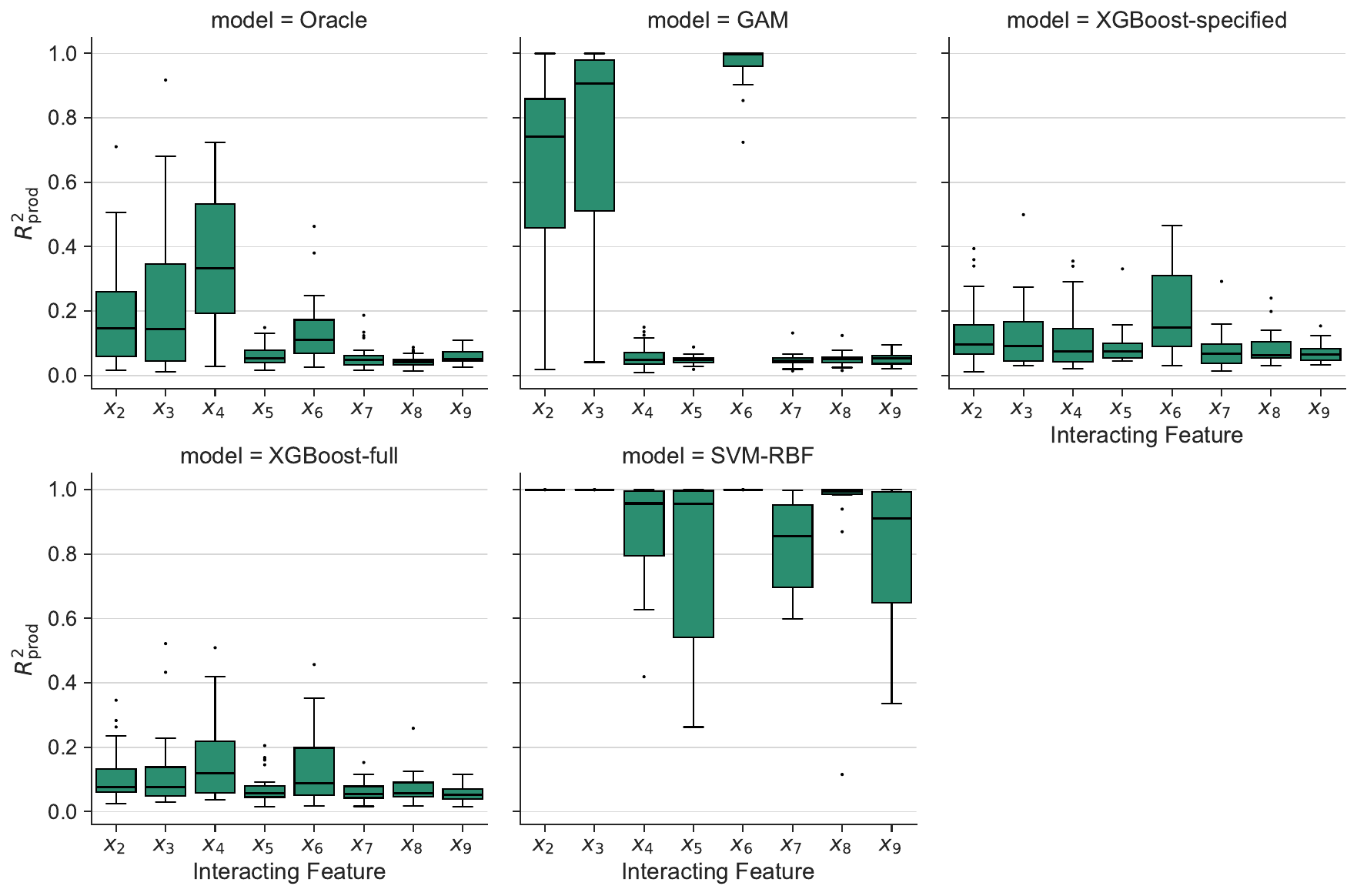}
        \caption{Product-separability measure $R^2_\text{prod}$ for setting \textbf{IV} (higher-order complex).}
    \end{subfigure}
    \caption{Full simulation results for \ours{} product-separable interaction categorization measure. $R^2_\text{prod}$ across 30 repetitions visualized as boxplots. For cases where all surrogate smooth terms were filtered out (below variance threshold) or the reference point did not exist for all smooth terms, no results are available.}
    \label{fig:cat-prod-full-results}
\end{figure*}

\FloatBarrier
\subsection{Further Simulation Results: Linear \& Product-Sep. Visualization}\label{app:further-results-lin-prod-visualization}

For the sake of completeness, we provide the remaining results for the \ours{} linear and product-separable interaction visualization in Fig.~\ref{fig:vis-lin-results} \&~\ref{fig:vis-ratio-results} for the oracle, as we restricted the analysis in \S\ref{sec:eval-vis} to a reduced set of interactions.
These additional results all confirm the patterns observed and discussed in \S\ref{sec:eval-vis}.
We solely focus on the oracle here, since for the fitted ML models, we do not know the true interactions learned by the models, and thus comparing the visualizations to ground truth functions remains difficult.

\begin{figure}[htbp]
    \centering
    \begin{subfigure}[t]{0.6\textwidth}
        \centering
        \includegraphics[width=\textwidth]{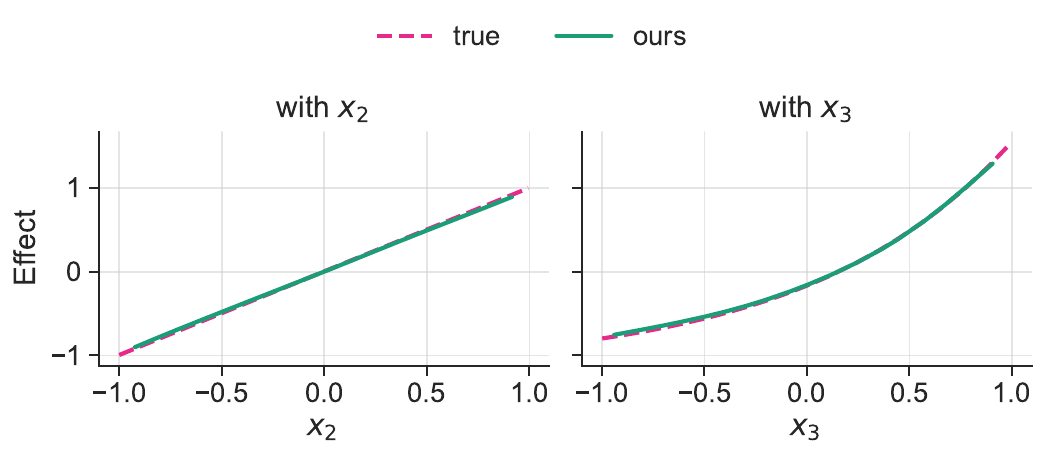}
        \caption{Setting \textbf{I} (two-way, indep.).}
    \end{subfigure}
    \begin{subfigure}[t]{0.6\textwidth}
        \centering
        \includegraphics[width=\textwidth]{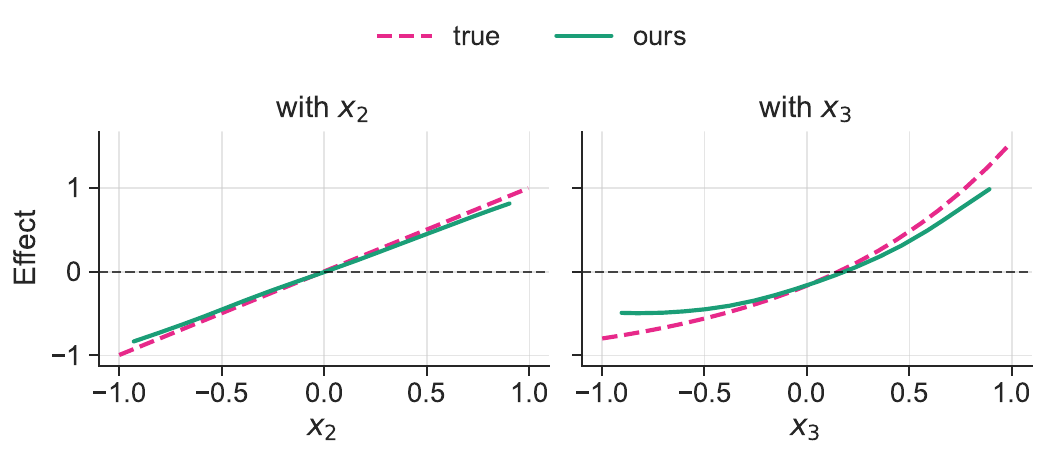}
        \caption{Setting \textbf{II} (two-way, corr.).}
    \end{subfigure}
    \begin{subfigure}[t]{0.6\textwidth}
        \centering
        \includegraphics[width=\textwidth]{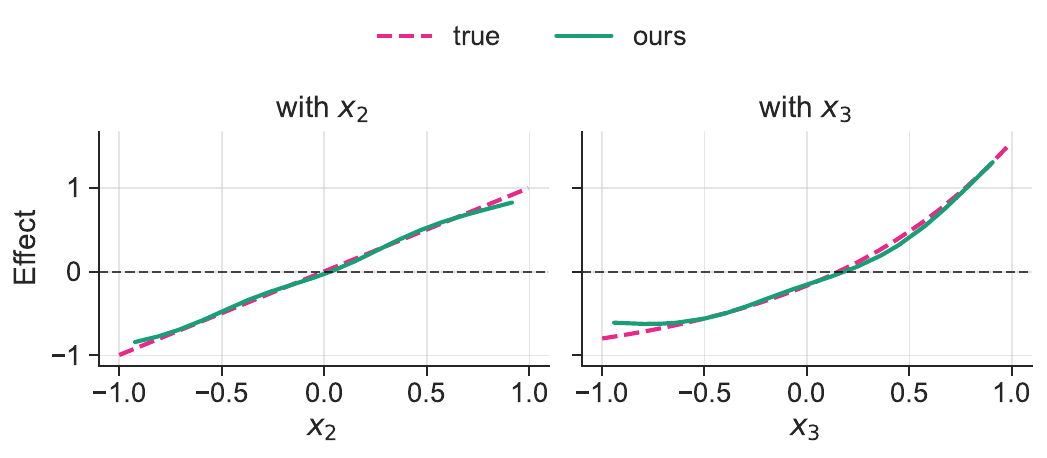}
        \caption{Setting \textbf{III} (higher-order simple).}
    \end{subfigure}
    \caption{Linear interaction visualizations for the oracle (last repetition). True is $\phi_l(X_l)$ (cf. Def.~\ref{def:linear}) as denoted in Tab.~\ref{tab:settings}.}\label{fig:vis-lin-results}
\end{figure}

\begin{figure}[h]
    \centering
    \includegraphics[width=0.6\textwidth]{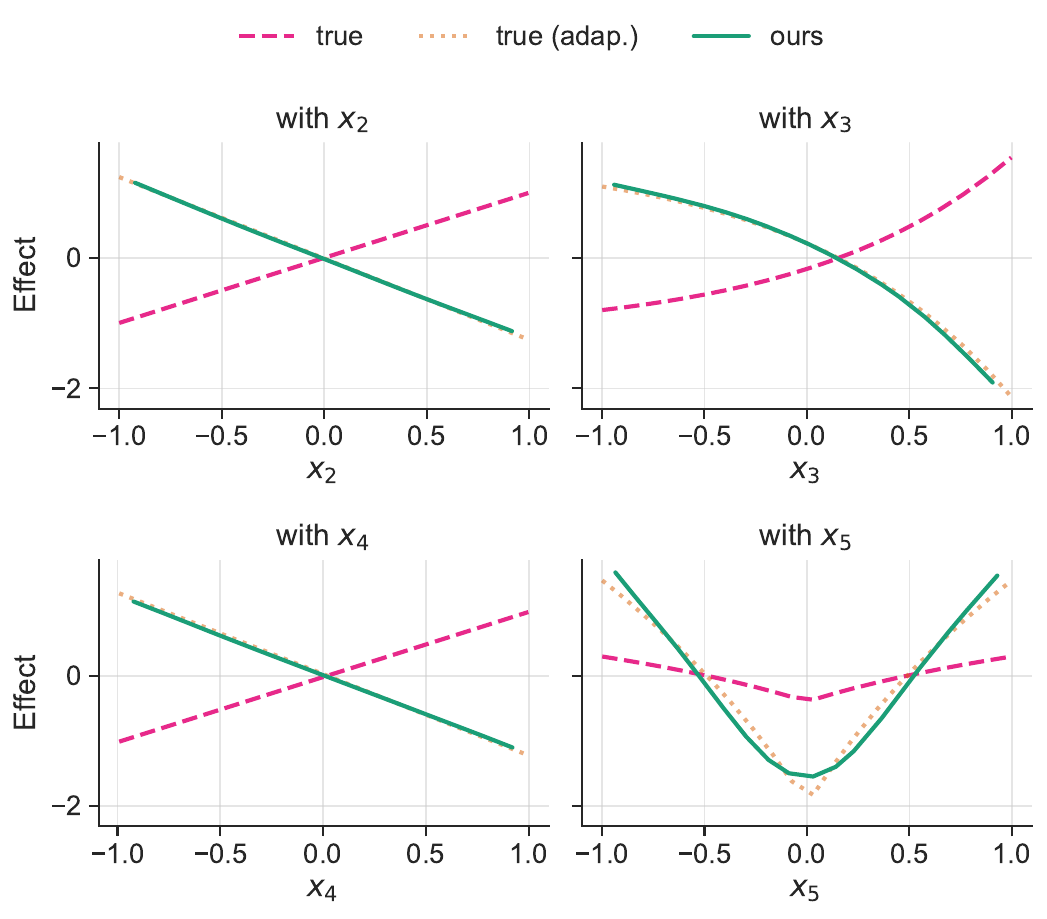}
    \caption{Product-separable interaction visualizations (ratio-based) for the oracle (last repetition) in Setting \textbf{I} (two-way, independent). In Setting \textbf{II}, only $X_5$ is detected as product-separable (cf.~\S\ref{sec:eval-vis}). For Setting \textbf{III} \& \textbf{IV}, no product-separable interactions are found. True is $\phi_l(X_l)$ (cf. Def.~\ref{def:prod-sep}) as denoted in Tab.~\ref{tab:settings}. Generally, $\phi_l$ is not uniquely defined. The result depends on $\tilde{x}_l^\text{ref}$; True (adap.) is rescaled accordingly as $\phi_l/\phi_l(\tilde{x}_l^\text{ref})$.}\label{fig:vis-ratio-results}
\end{figure}

\FloatBarrier
\subsection{Further Simulation Results: General Interaction Visualization}\label{app:further-results-general-visualization}

We provide full results of the \ours{} general interaction visualization based on integrated smooth terms for each model (oracle and ML models) from the last repetition of each experiment \textbf{I}, \textbf{II}, \textbf{III}, and \textbf{IV}.
We only visualize interactions for which the detection heuristic is below $\alpha=0.05$.
For the fitted ML models, we find that they all learn different interaction forms according to \ours{}.
The GAM learns smooth shapes close to the true interactions but more non-linear in the interacting feature under independence, and more linear under correlations. For both higher-order settings, only $X_2, X_3, X_6$ have detected interactions.
The SVM predominantly learns linear interactions, which is consistent with a very small $\gamma$ (inverse kernel width) chosen in the hyperparameter optimization, inducing strong regularization that makes model behavior near-linear.
XGBoost models generate highly non-linear shapes, particularly for weak interactions, with neighboring interval curves often grouped together, highlighting the trees' discrete partitioning.
These observations are consistent with the results for interaction categorization, and demonstrate that similarly performing models learn distinctly different interaction effects, exemplifying the \textit{Rashomon effect}~\cite{breiman-ss01a} in ML.
Although we cannot directly validate these visualizations as we do not know the true form of the learned interactions (only for the GAM), they indicate the potential for valuable insights into the models' internal workings.

% setting I
\begin{figure*}[htbp]
    \centering
    \begin{subfigure}[t]{\textwidth}
        \centering
        \includegraphics[width=0.98\textwidth]{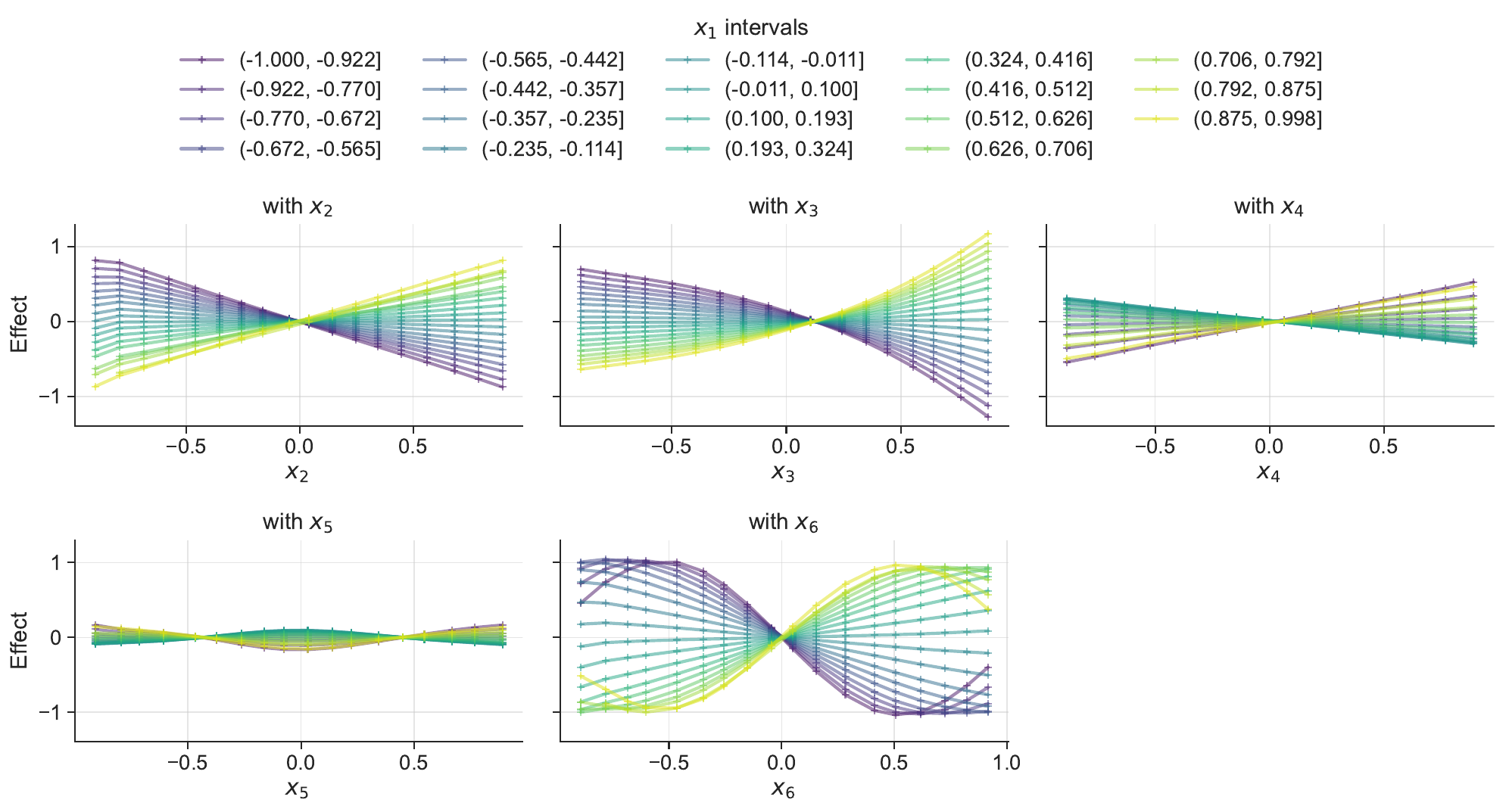}
        \caption{Oracle, setting \textbf{I} (two-way, independent).}
    \end{subfigure}
\end{figure*}

\begin{figure*}[htbp]
    \centering
    \ContinuedFloat
    \begin{subfigure}[t]{\textwidth}
        \centering
        \includegraphics[width=0.98\textwidth]{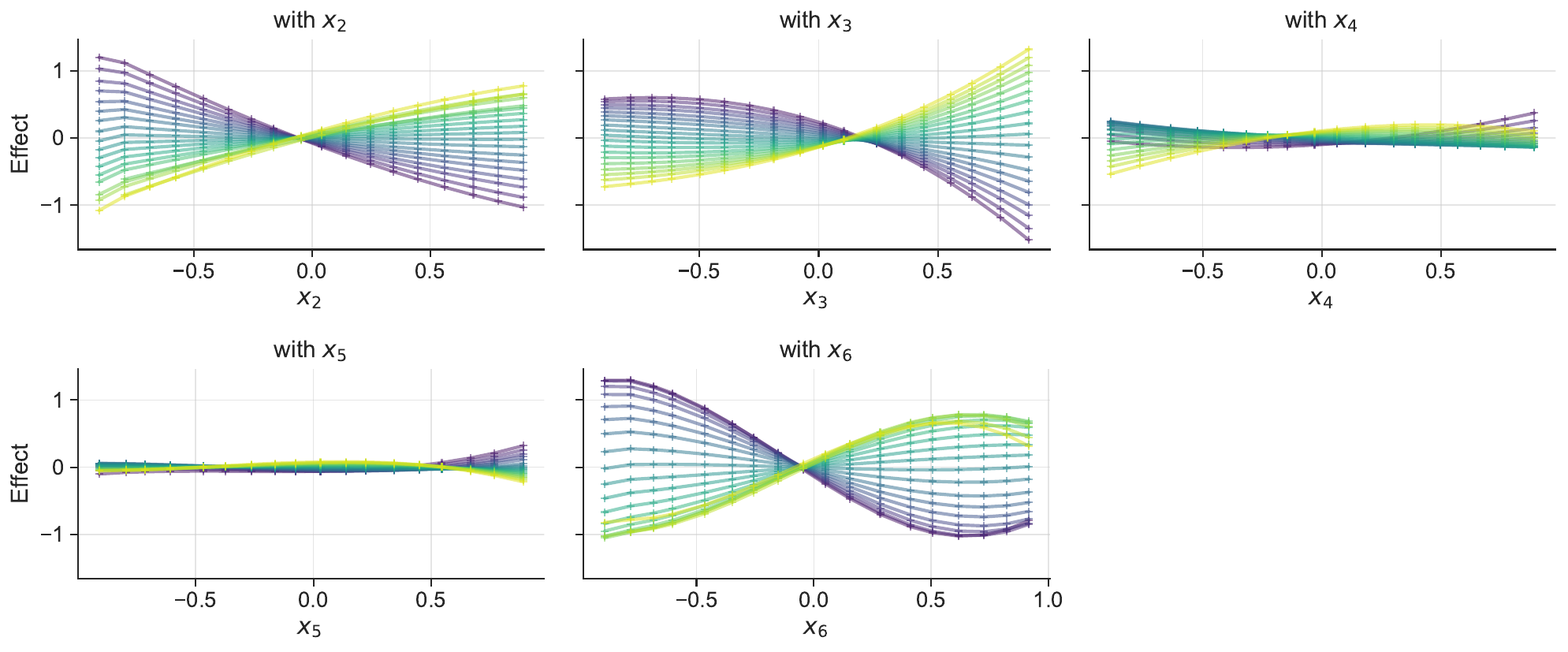}
        \caption{GAM, setting \textbf{I} (two-way, independent).}
    \end{subfigure}
\end{figure*}

\begin{figure*}[htbp]
    \centering
    \ContinuedFloat
    \begin{subfigure}[t]{\textwidth}
        \centering
        \includegraphics[width=0.98\textwidth]{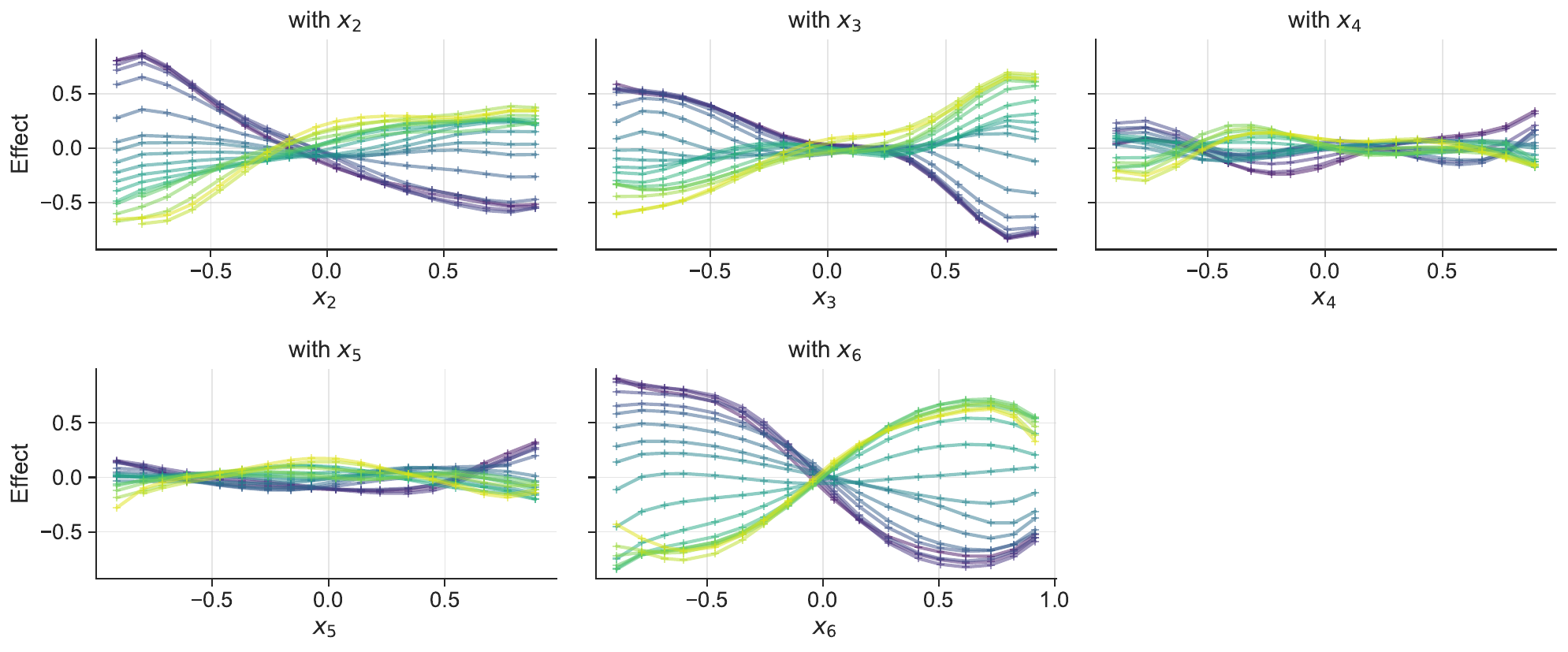}
        \caption{XGB spec, setting \textbf{I} (two-way, independent).}
    \end{subfigure}
\end{figure*}

\begin{figure*}[htbp]
    \centering
    \ContinuedFloat
    \begin{subfigure}[t]{\textwidth}
        \centering
        \includegraphics[width=0.98\textwidth]{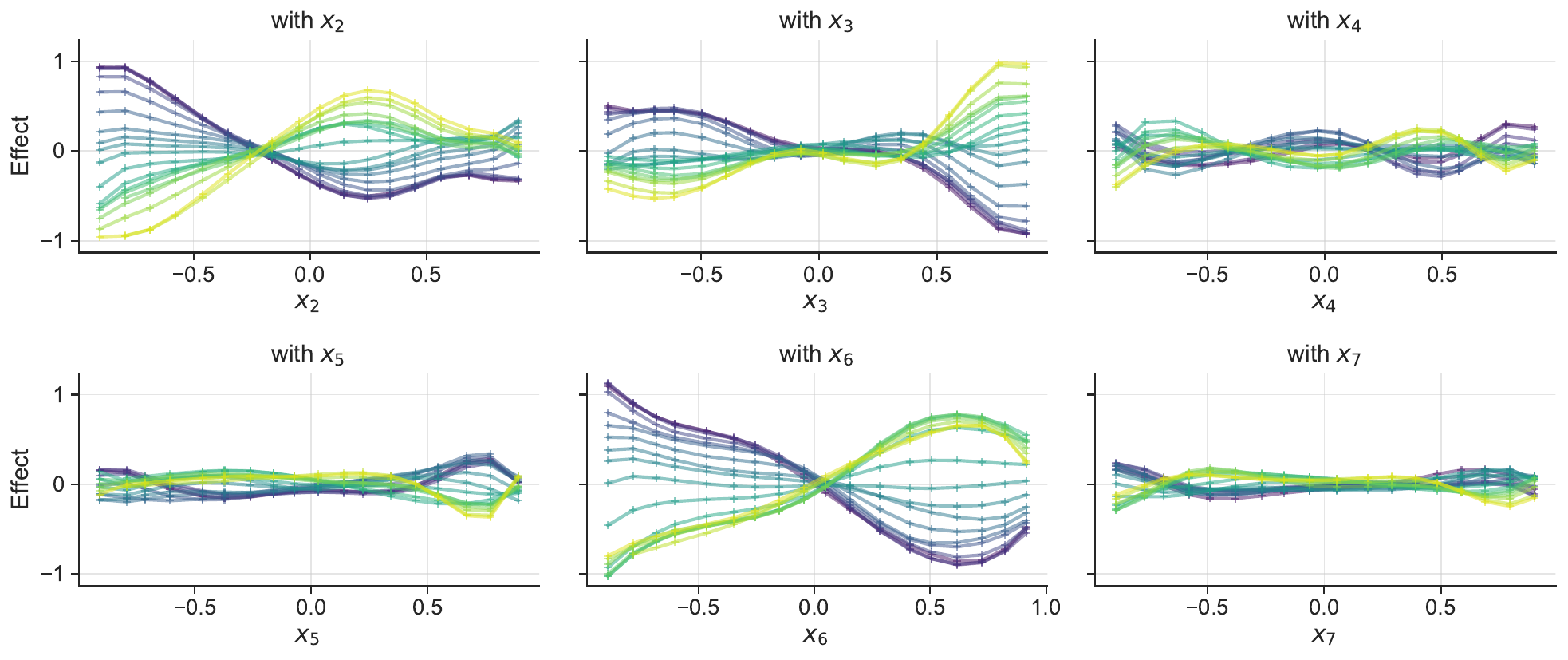}
        \caption{XGB full, setting \textbf{I} (two-way, independent).}
    \end{subfigure}
\end{figure*}

\begin{figure*}[htbp]
    \centering
    \ContinuedFloat
    \begin{subfigure}[t]{\textwidth}
        \centering
        \includegraphics[width=0.98\textwidth]{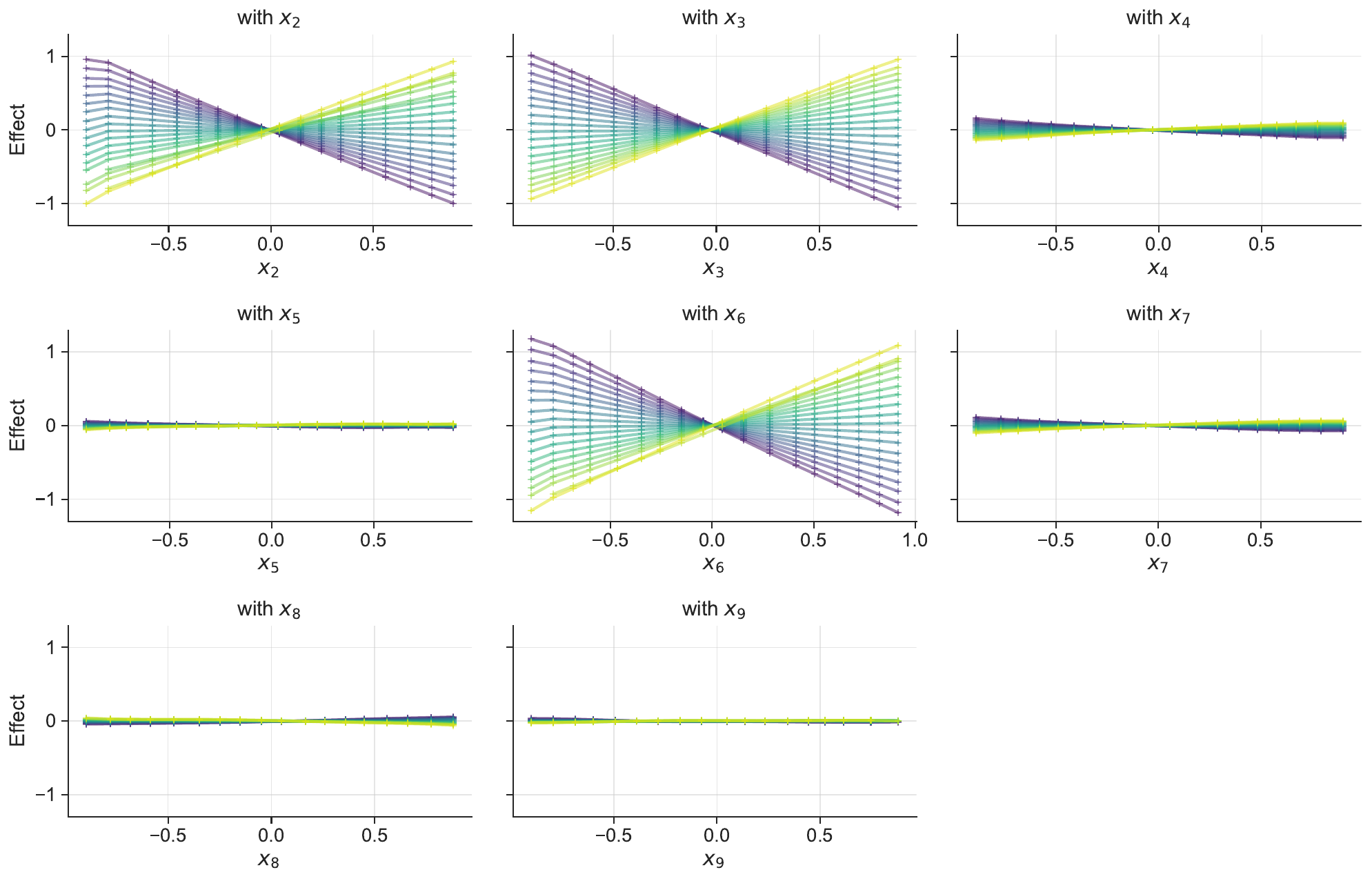}
        \caption{SVM-RBF, setting \textbf{I} (two-way, independent).}
    \end{subfigure}
    \caption{General interaction visualization (integrated smooths) for setting \textbf{I} (two-way, independent). Each curve corresponds to an interval of $X_1$.}
    \label{fig:vis-full-res-i}
\end{figure*}

% setting II
\begin{figure*}[htbp]
    \centering
    \begin{subfigure}[t]{\textwidth}
        \centering
        \includegraphics[width=0.98\textwidth]{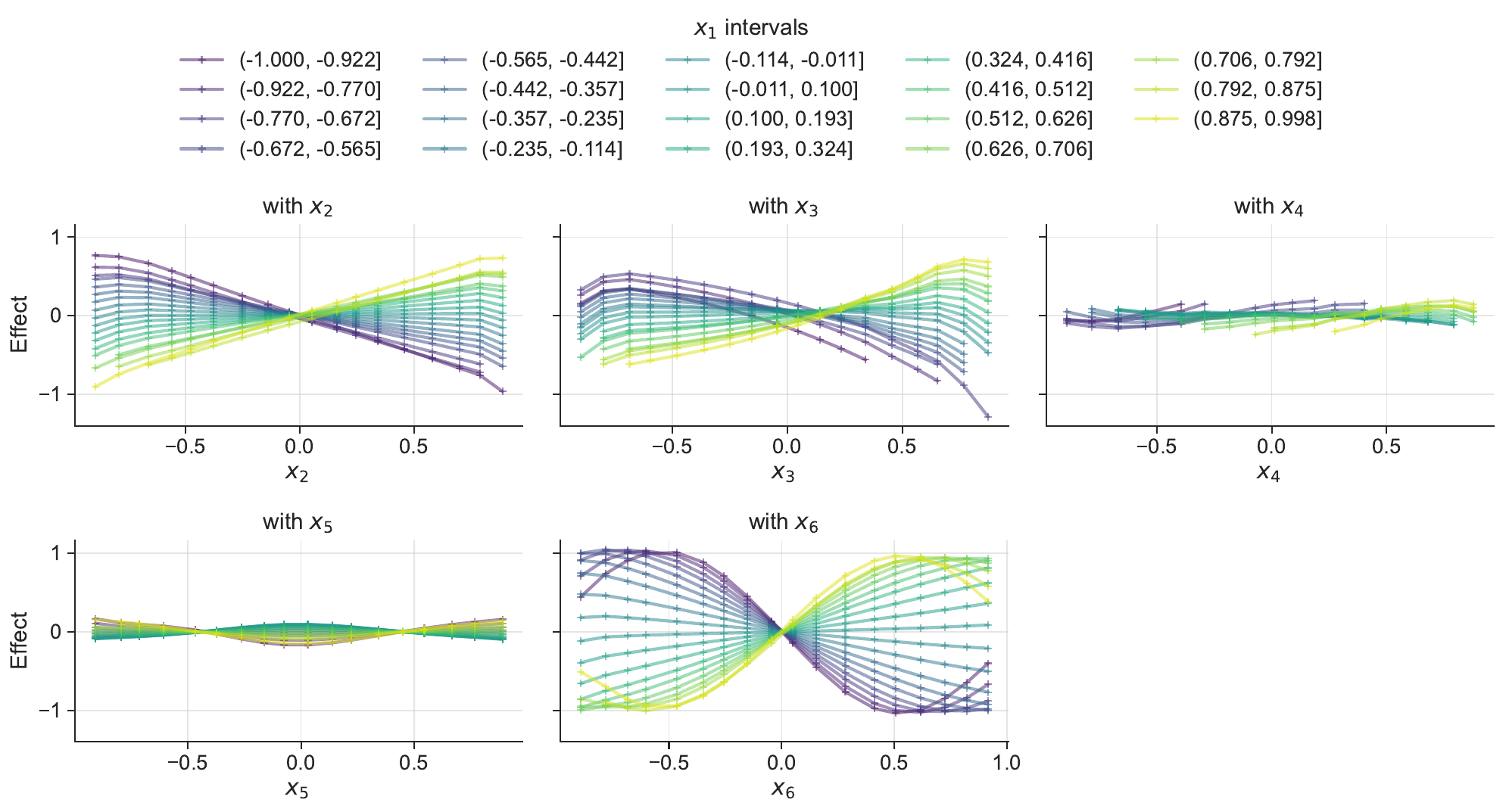}
        \caption{Oracle, setting \textbf{II} (two-way, correlated).}
    \end{subfigure}
\end{figure*}

\begin{figure*}[htbp]
    \centering
    \ContinuedFloat
    \begin{subfigure}[t]{\textwidth}
        \centering
        \includegraphics[width=0.98\textwidth]{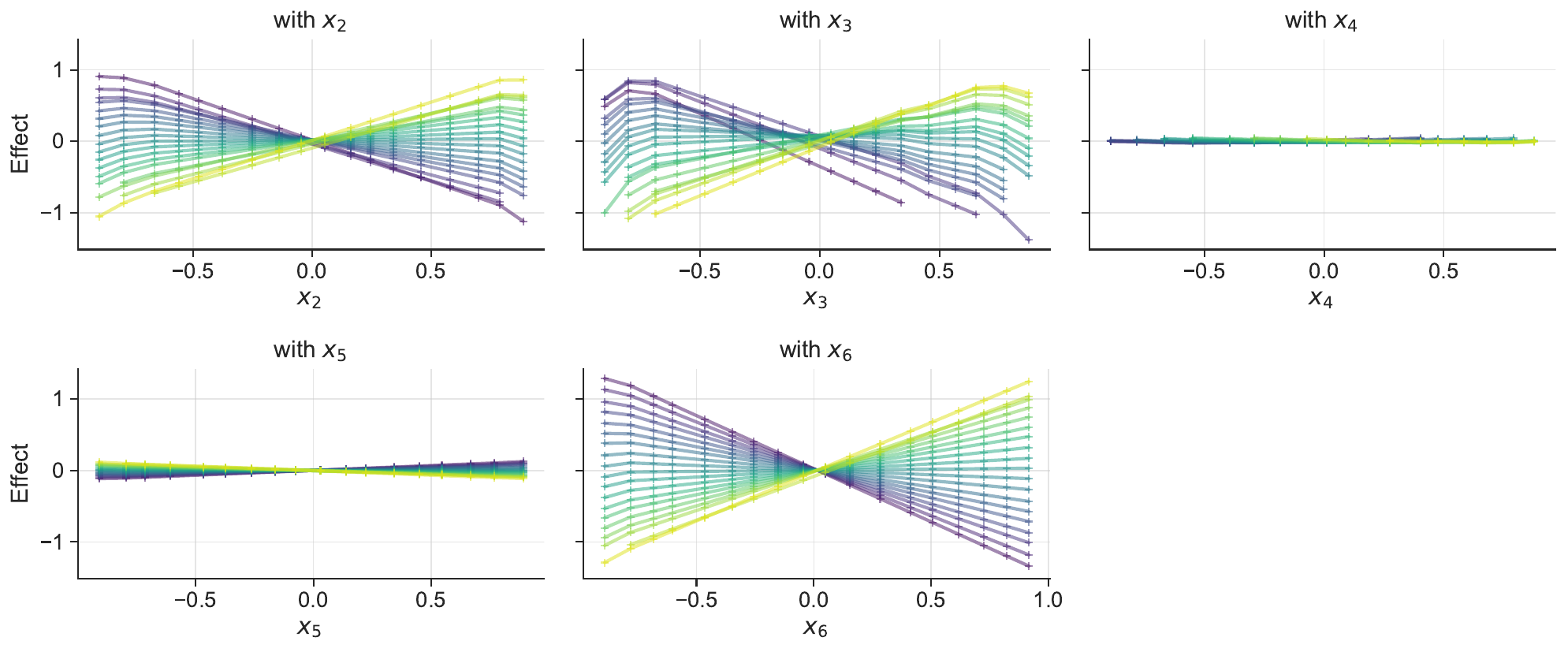}
        \caption{GAM, setting \textbf{II} (two-way, correlated).}
    \end{subfigure}
\end{figure*}

\begin{figure*}[htbp]
    \centering
    \ContinuedFloat
    \begin{subfigure}[t]{\textwidth}
        \centering
        \includegraphics[width=0.98\textwidth]{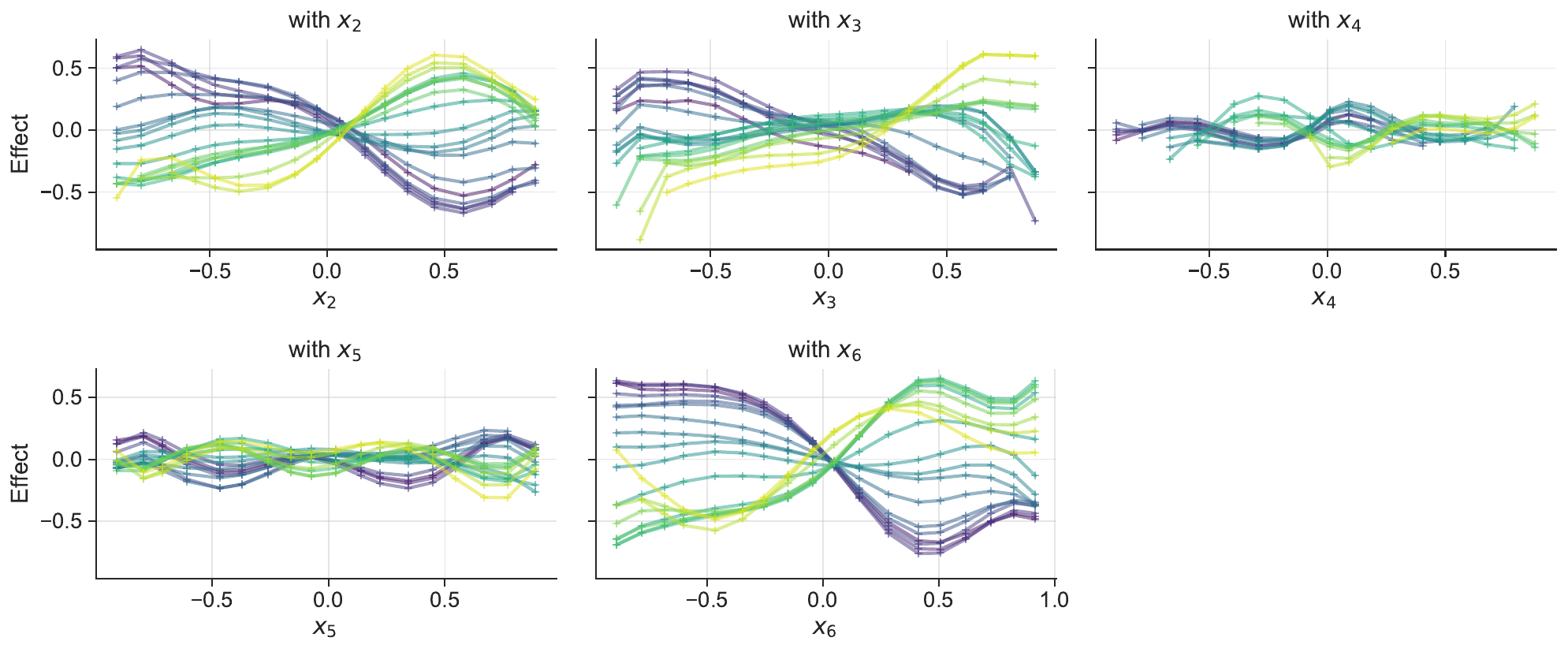}
        \caption{XGB spec, setting \textbf{II} (two-way, correlated).}
    \end{subfigure}
\end{figure*}

\begin{figure*}[htbp]
    \centering
    \ContinuedFloat
    \begin{subfigure}[t]{\textwidth}
        \centering
        \includegraphics[width=0.98\textwidth]{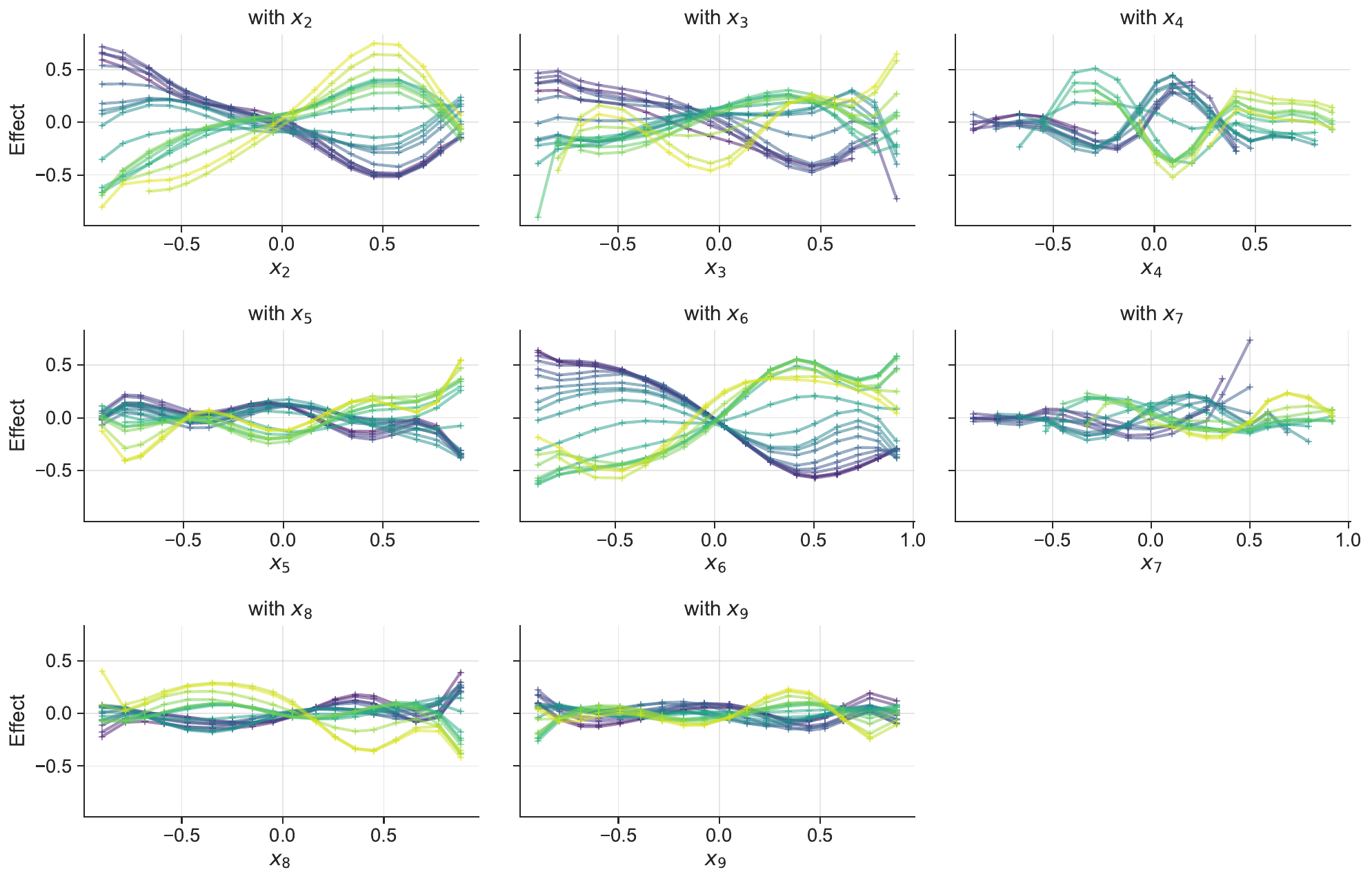}
        \caption{XGB full, setting \textbf{II} (two-way, correlated).}
    \end{subfigure}
\end{figure*}

\begin{figure*}[htbp]
    \centering
    \ContinuedFloat
    \begin{subfigure}[t]{\textwidth}
        \centering
        \includegraphics[width=0.98\textwidth]{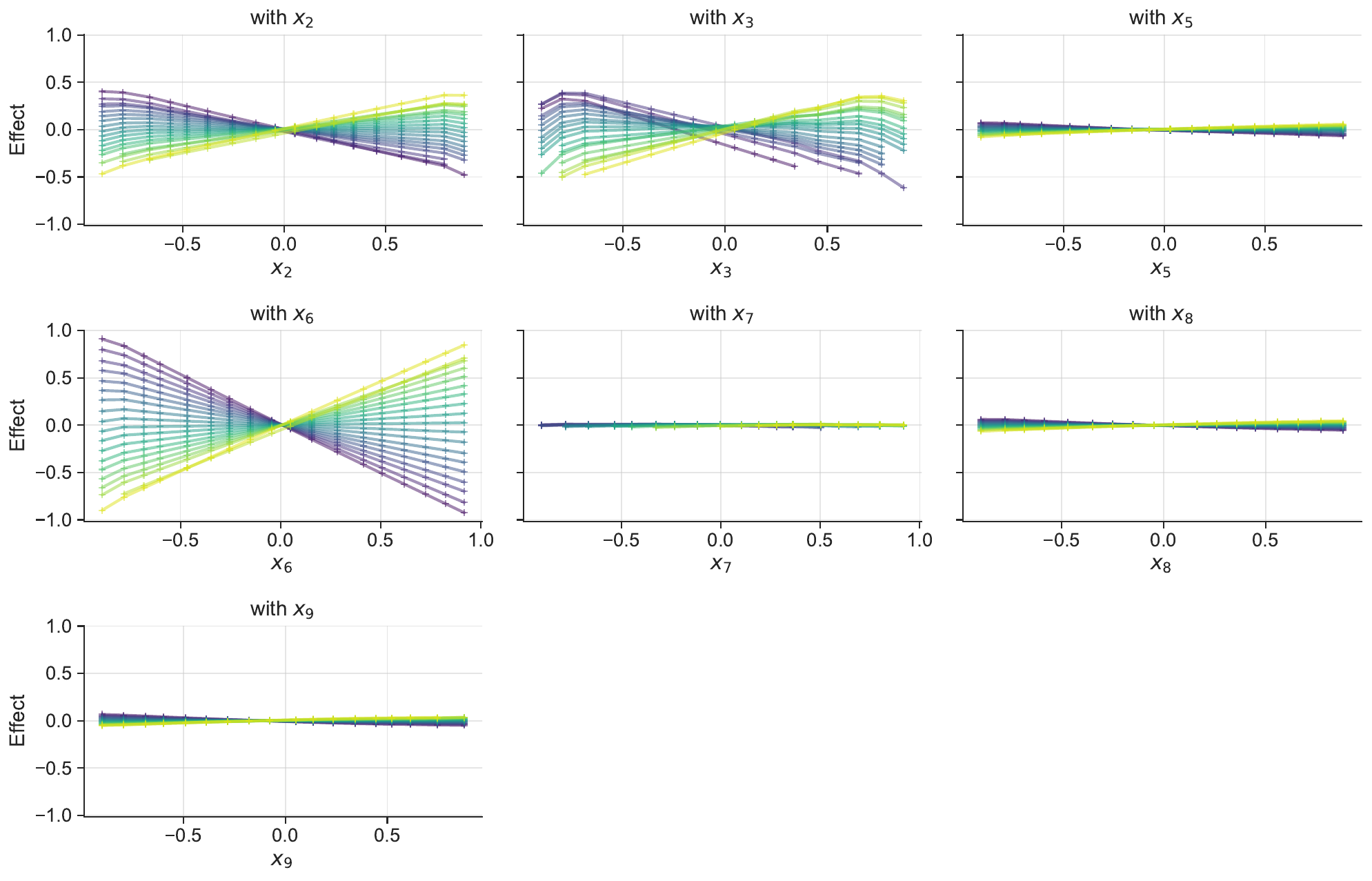}
        \caption{SVM-RBF, setting \textbf{II} (two-way, correlated).}
    \end{subfigure}
    \caption{General interaction visualization (integrated smooths) for setting \textbf{II} (two-way, correlated). Each curve corresponds to an interval of $X_1$.}
    \label{fig:vis-full-res-ii}
\end{figure*}

% setting III
\begin{figure*}[htbp]
    \centering
    \begin{subfigure}[t]{\textwidth}
        \centering
        \includegraphics[width=0.98\textwidth]{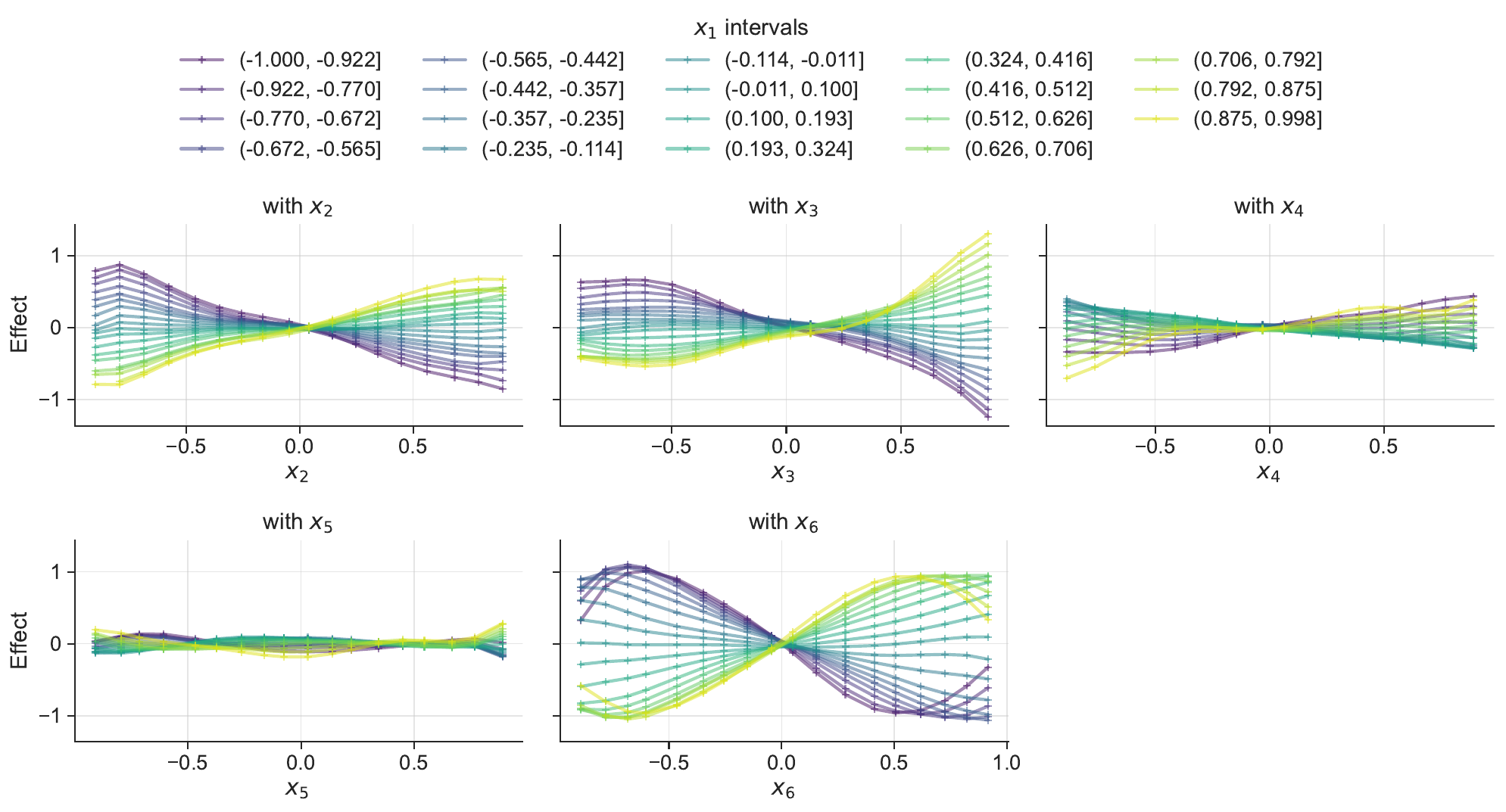}
        \caption{Oracle, setting \textbf{III} (higher-order simple).}
    \end{subfigure}
\end{figure*}

\begin{figure*}[htbp]
    \centering
    \ContinuedFloat
    \begin{subfigure}[t]{\textwidth}
        \centering
        \includegraphics[width=0.98\textwidth]{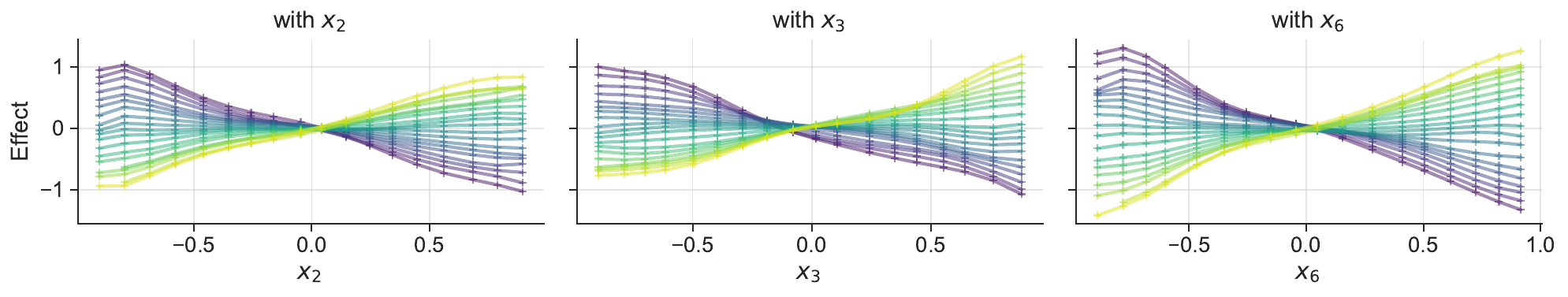}
        \caption{GAM, setting \textbf{III} (higher-order simple).}
    \end{subfigure}
\end{figure*}

\begin{figure*}[htbp]
    \centering
    \ContinuedFloat
    \begin{subfigure}[t]{\textwidth}
        \centering
        \includegraphics[width=0.98\textwidth]{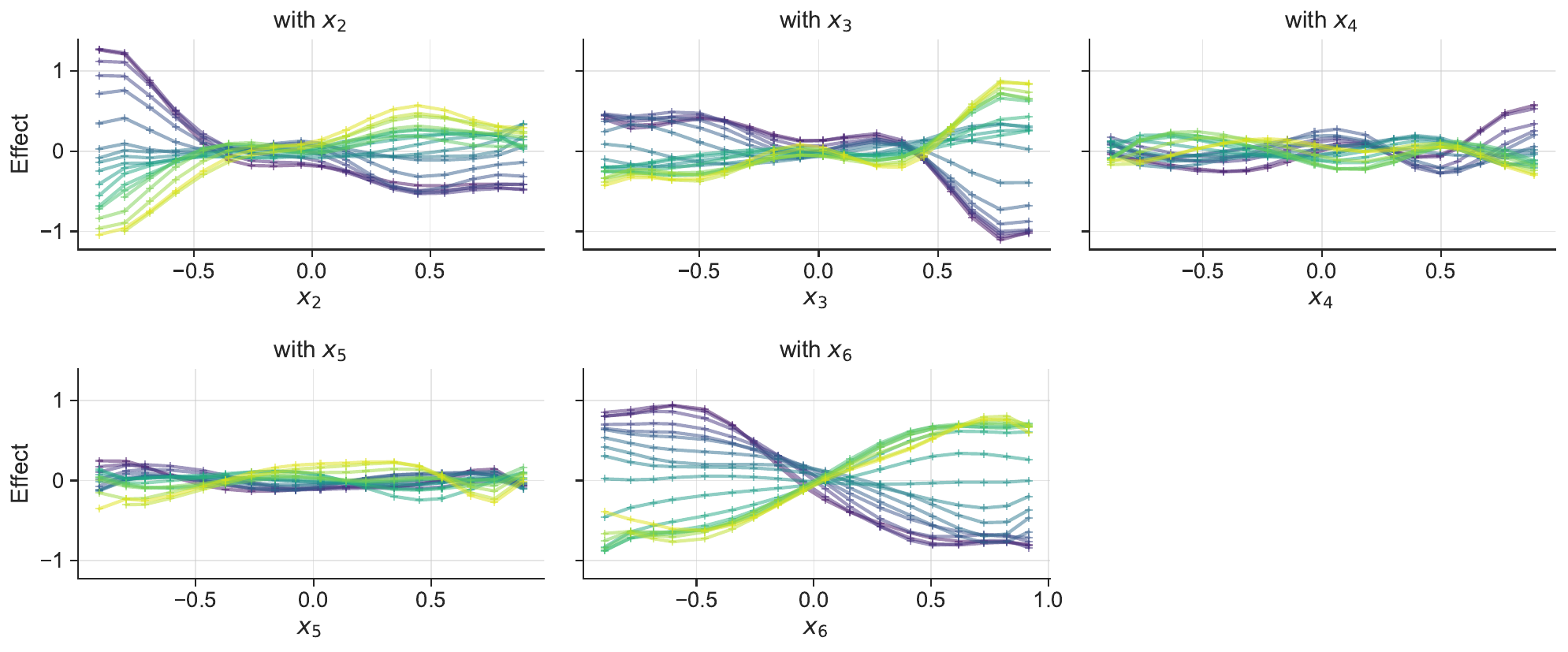}
        \caption{XGB spec, setting \textbf{III} (higher-order simple).}
    \end{subfigure}
\end{figure*}

\begin{figure*}[htbp]
    \centering
    \ContinuedFloat
    \begin{subfigure}[t]{\textwidth}
        \centering
        \includegraphics[width=0.98\textwidth]{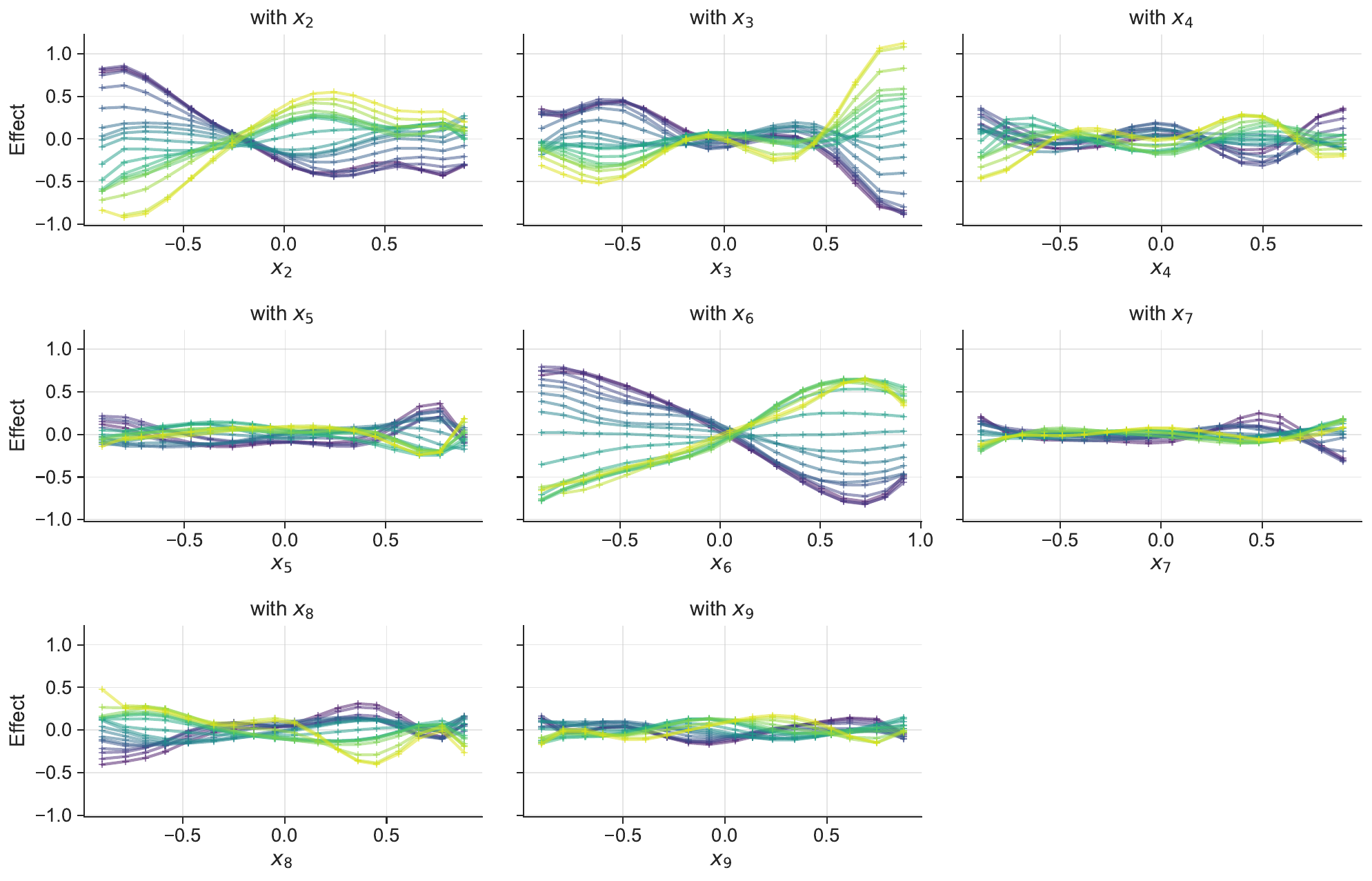}
        \caption{XGB full, setting \textbf{III} (higher-order simple).}
    \end{subfigure}
\end{figure*}

\begin{figure*}[htbp]
    \centering
    \ContinuedFloat
    \begin{subfigure}[t]{\textwidth}
        \centering
        \includegraphics[width=0.98\textwidth]{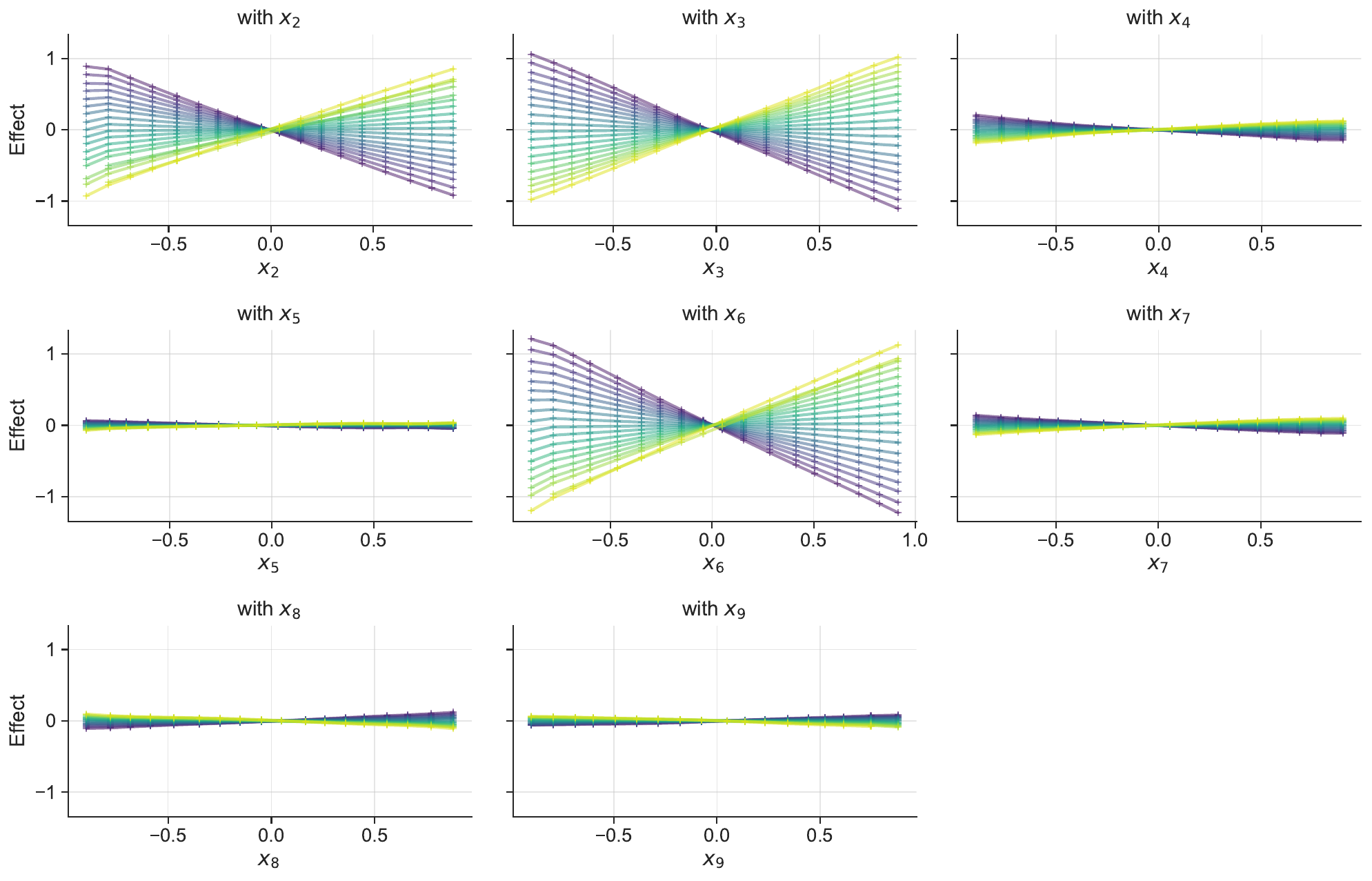}
        \caption{SVM-RBF, setting \textbf{III} (higher-order simple).}
    \end{subfigure}
    \caption{General interaction visualization (integrated smooths) for setting \textbf{III} (higher-order simple). Each curve corresponds to an interval of $X_1$.}
    \label{fig:vis-full-res-iii}
\end{figure*}

% setting IV
\begin{figure*}[htbp]
    \centering
    \begin{subfigure}[t]{\textwidth}
        \centering
        \includegraphics[width=0.98\textwidth]{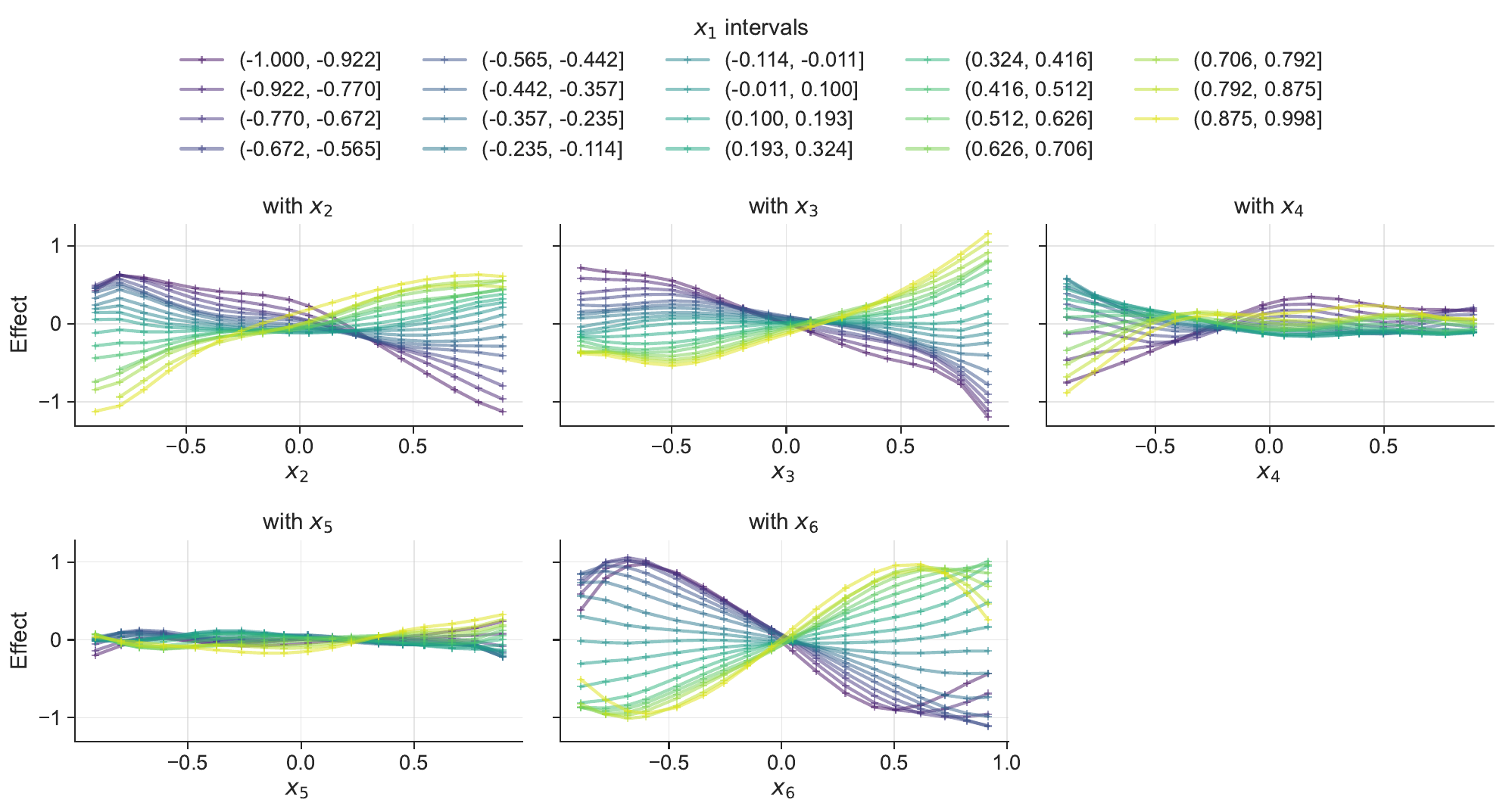}
        \caption{Oracle, setting \textbf{IV} (higher-order complex).}
    \end{subfigure}
\end{figure*}

\begin{figure*}[htbp]
    \centering
    \ContinuedFloat
    \begin{subfigure}[t]{\textwidth}
        \centering
        \includegraphics[width=0.98\textwidth]{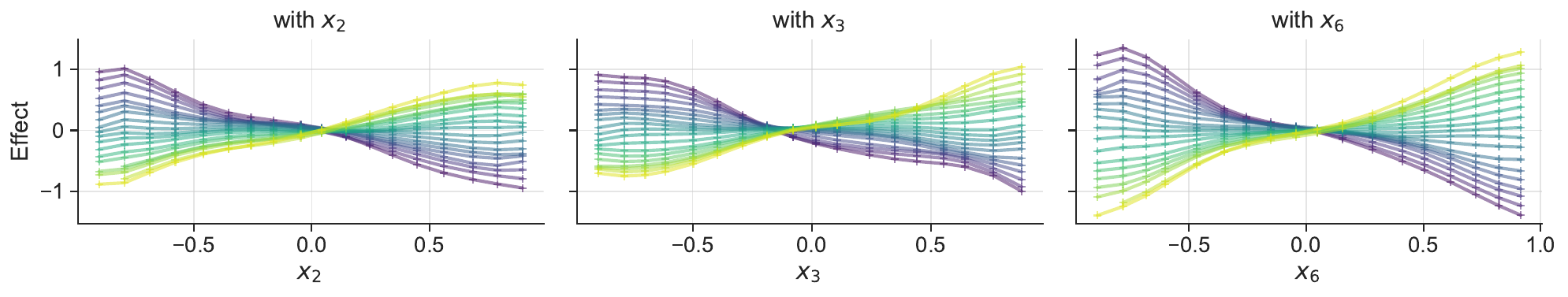}
        \caption{GAM, setting \textbf{IV} (higher-order complex).}
    \end{subfigure}
\end{figure*}

\begin{figure*}[htbp]
    \centering
    \ContinuedFloat
    \begin{subfigure}[t]{\textwidth}
        \centering
        \includegraphics[width=0.98\textwidth]{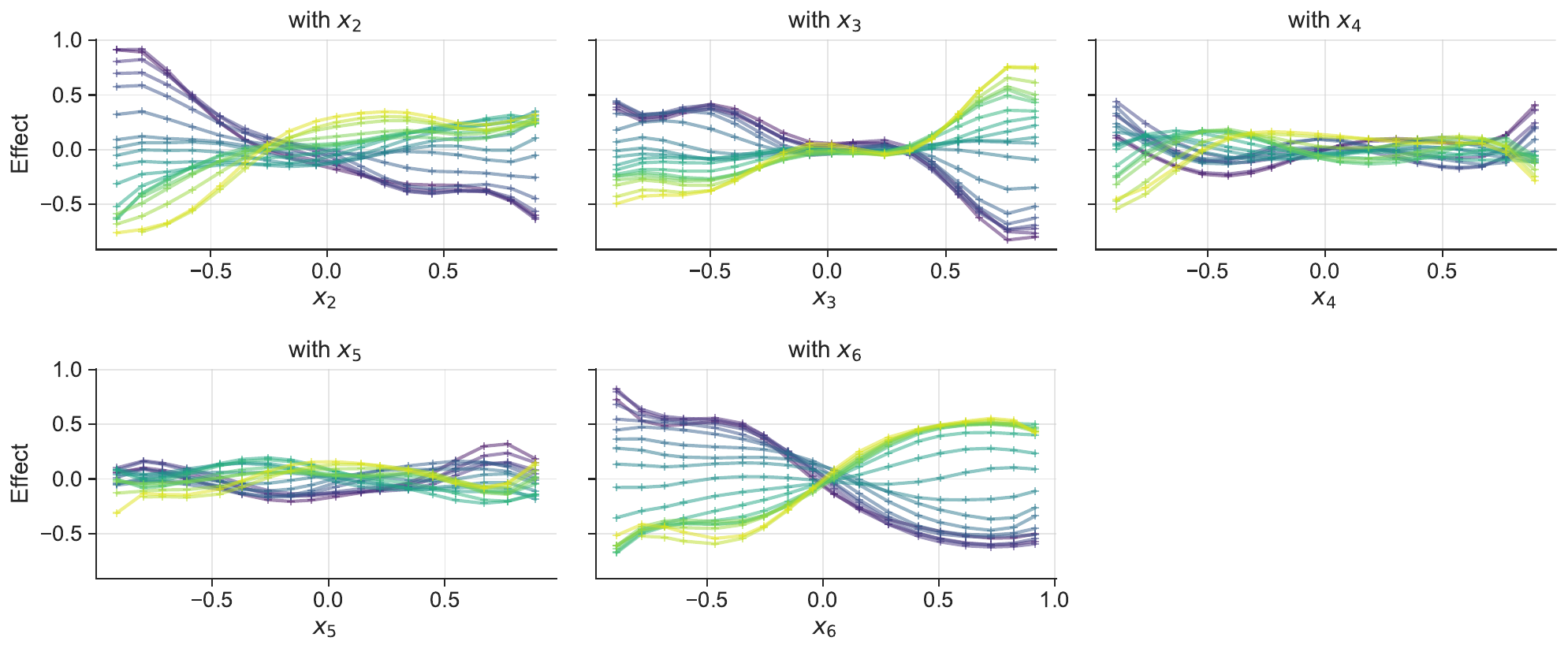}
        \caption{XGB spec, setting \textbf{IV} (higher-order complex).}
    \end{subfigure}
\end{figure*}

\begin{figure*}[htbp]
    \centering
    \ContinuedFloat
    \begin{subfigure}[t]{\textwidth}
        \centering
        \includegraphics[width=0.98\textwidth]{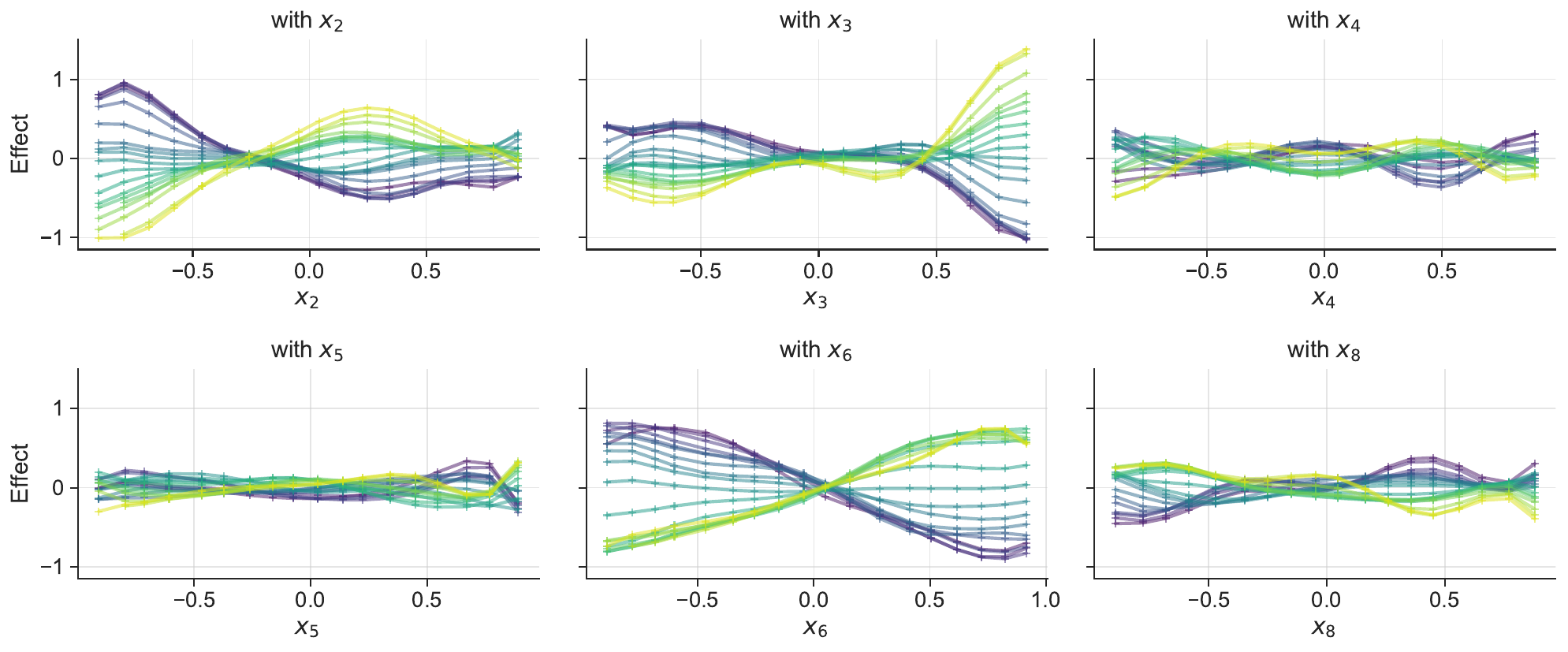}
        \caption{XGB full, setting \textbf{IV} (higher-order complex).}
    \end{subfigure}
\end{figure*}

\begin{figure*}[htbp]
    \centering
    \ContinuedFloat
    \begin{subfigure}[t]{\textwidth}
        \centering
        \includegraphics[width=0.98\textwidth]{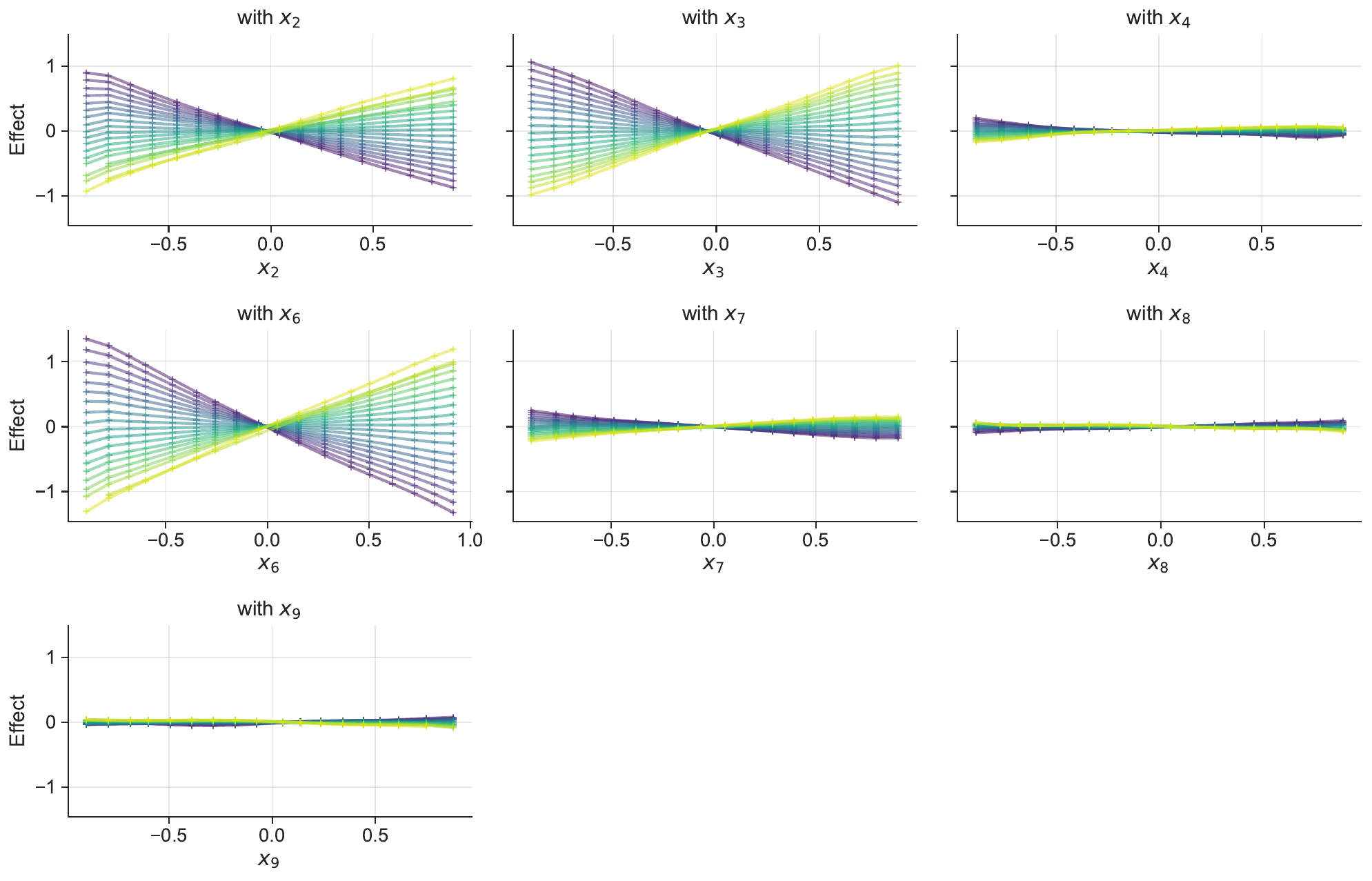}
        \caption{SVM-RBF, setting \textbf{IV} (higher-order complex).}
    \end{subfigure}
    \caption{General interaction visualization (integrated smooths) for setting \textbf{IV} (higher-order complex). Each curve corresponds to an interval of $X_1$.}
    \label{fig:vis-full-res-iv}
\end{figure*}

\fi

\end{document}